\documentclass[10pt,twocolumn,letterpaper]{article}

\usepackage{cvpr}              

\usepackage{amsmath,amsfonts,bm}

\def\eqref#1{equation~\ref{#1}}

\def\1{\bm{1}}

\DeclareMathAlphabet{\mathsfit}{\encodingdefault}{\sfdefault}{m}{sl}
\SetMathAlphabet{\mathsfit}{bold}{\encodingdefault}{\sfdefault}{bx}{n}

\DeclareMathOperator*{\argmax}{arg\,max}

\usepackage[accsupp]{axessibility}
\usepackage{url}

\newcommand{\norm}[1]{\lVert #1 \rVert}
\usepackage{booktabs}       
\usepackage{amsfonts}       
\usepackage{nicefrac}       
\usepackage{microtype}      
\usepackage{xcolor}         
\usepackage{amsmath}
\usepackage{graphicx}
 \usepackage{multirow}
\usepackage{subcaption}
\usepackage{bbm}

\usepackage[pagebackref,breaklinks,colorlinks]{hyperref}
\usepackage[capitalize]{cleveref}
\crefname{section}{Sec.}{Secs.}
\Crefname{section}{Section}{Sections}
\Crefname{table}{Table}{Tables}
\crefname{table}{Tab.}{Tabs.}

\def\cvprPaperID{8898} 
\def\confName{CVPR}
\def\confYear{2022}

\begin{document}

\title{Segment and Complete: Defending Object Detectors against Adversarial Patch Attacks with Robust Patch Detection}

\author{Jiang Liu \textsuperscript{1}, Alexander Levine\textsuperscript{2}, Chun Pong Lau\textsuperscript{1}, Rama Chellappa\textsuperscript{1}, Soheil Feizi\textsuperscript{2}\\
\textsuperscript{1}Johns Hopkins University, \textsuperscript{2}University of Maryland, College Park \\
{\tt\small \{jiangliu, clau13, rchella4\}@jhu.edu, \{alevine0,sfeizi\}@cs.umd.edu}}
\maketitle

\begin{abstract}
Object detection plays a key role in many security-critical systems. Adversarial patch attacks, which are easy to implement in the physical world, pose a serious threat to state-of-the-art object detectors. Developing reliable defenses for object detectors against patch attacks is critical but severely understudied. In this paper, we propose Segment and Complete defense (SAC), a general framework for defending object detectors against patch attacks through detection and removal of adversarial patches. We first train a patch segmenter that outputs patch masks which provide pixel-level localization of adversarial patches. We then propose a self adversarial training algorithm to robustify the patch segmenter. In addition, we design a robust shape completion algorithm, which is guaranteed to remove the entire patch from the images if the outputs of the patch segmenter are within a certain Hamming distance of the ground-truth patch masks. Our experiments on COCO and xView datasets demonstrate that SAC achieves superior robustness even under strong adaptive attacks with no reduction in performance on clean images, and generalizes well to unseen patch shapes, attack budgets, and unseen attack methods. Furthermore, we present the APRICOT-Mask dataset, which augments the APRICOT dataset with pixel-level annotations of adversarial patches. We show SAC can significantly reduce the targeted attack success rate of physical patch attacks. Our code is available at \url{https://github.com/joellliu/SegmentAndComplete}.
\end{abstract}

\section{Introduction}
\begin{figure}[h]
  \centering
  \includegraphics[width=0.4\textwidth]{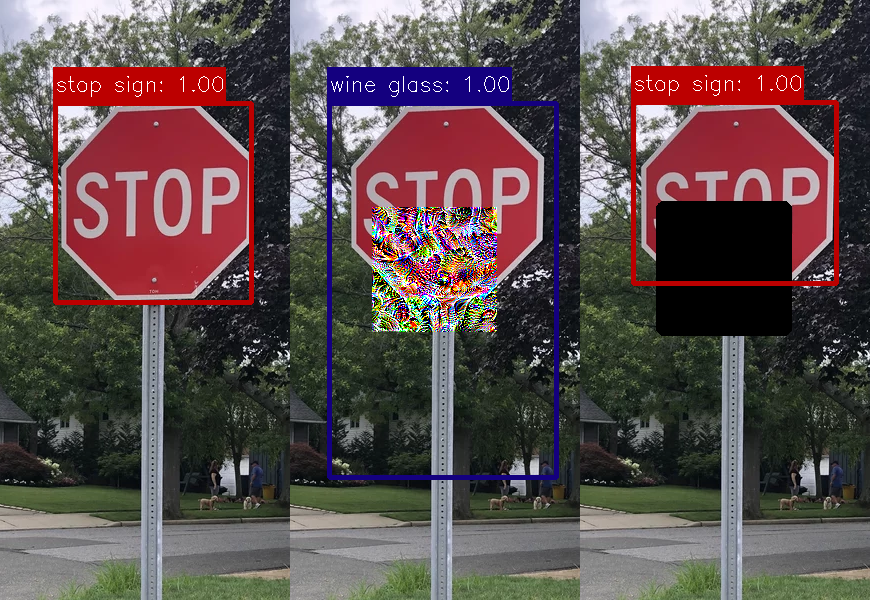}
  \caption{We adopt a ``detect and remove" strategy for defending object detectors against patch attacks. Left: Predictions on a clean image; middle: predictions on an adversarial image; right: predictions on SAC masked image.} 
  \label{fig:teaser}
\end{figure}
Object detection is an important computer vision task that plays a key role in many security-critical systems including autonomous driving, security surveillance, identity verification, and robot manufacturing~\cite{vahab2019applications}. Adversarial patch attacks, where the attacker distorts pixels within a region of bounded size, pose a serious threat to real-world object detection systems since they are easy to implement physically. For example, physical adversarial patches can make a stop sign~\cite{song2018physical} or a person~\cite{thys2019fooling} disappear from object detectors, which could cause serious consequences in security-critical settings such as autonomous driving. Despite the abundance ~\cite{song2018physical, thys2019fooling, wu2020making, liu2018dpatch, zhao2020object, li2018exploring, lang2021attention,chen2018shapeshifter, lee2019physical, wang2021towards,  10.1145/2976749.2978392} of adversarial patch attacks on object detectors, defenses against such attacks have not been extensively studied. Most existing defenses for patch attacks are restricted to image classification~\cite{10.1007/978-3-030-61638-0_31, wu2019defending, hayes2018visible, gittings2020vax, rao2020adversarial,  Chiang*2020Certified, xiang2021patchguard, levine2020randomized}. Securing object detectors is more challenging due to the complexity of the task. 

In this paper, we present Segment and Complete (SAC) defense that can robustify any object detector against patch attacks without re-training the object detectors. We adopt a ``detect and remove" strategy (\cref{fig:teaser}): we detect adversarial patches and remove the area from input images, and then feed the masked images into the object detector. This is based on the following observation: while adversarial patches are localized, they can affect predictions not only locally but also on objects that are farther away in the image because object detection algorithms utilize spatial context for reasoning~\cite{saha2020role}. This effect is especially significant for deep learning models, as a small localized adversarial patch can significantly disturb feature maps on a large scale due to large receptive fields of neurons. By removing them from the images, we minimize the adverse effects of adversarial patches both locally and globally.

\begin{figure*}[h]
  \centering
  \includegraphics[width=1\textwidth]{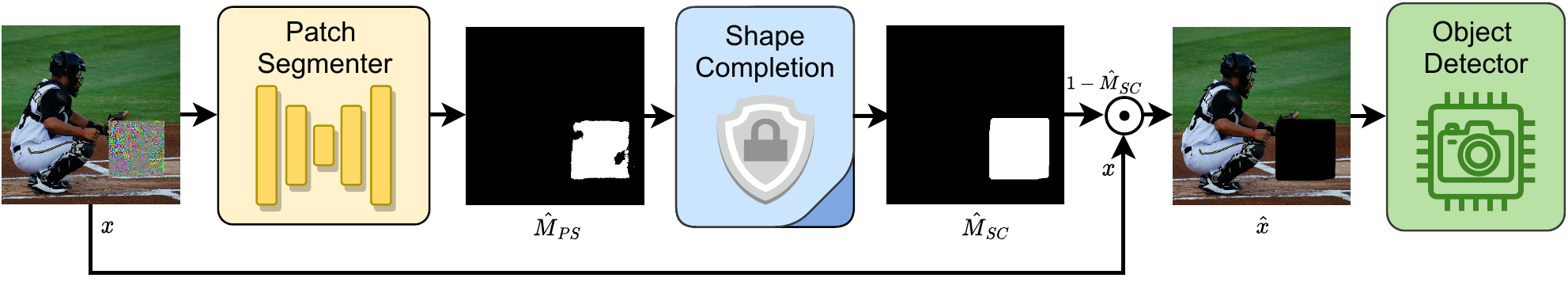}
  \caption{Pipeline of the SAC approach. SAC detects and removes adversarial patches on pixel-level through patch segmentation and shape completion, and feeds the masked images into the base object detector for prediction.} 
  \label{fig:pipeline}
\end{figure*}

The key of SAC is to robustly detect adversarial patches. We first train a patch segmenter to segment adversarial patches from the inputs and produce an initial patch mask. We propose a self adversarial training algorithm to enhance the robustness of the patch segmenter, which is efficient and object-detector agnostic. Since the attackers can potentially attack the segmenter and disturb its outputs under adaptive attacks, we further propose a robust shape completion algorithm that exploits the patch shape prior to ensure robust detection of adversarial patches. Shape completion takes the initial patch mask and generates a ``completed patch mask" that is {\it guaranteed} to cover the entire adversarial patch, given that the initial patch mask is within a certain Hamming distance from the ground-truth patch mask. The overall pipeline of SAC is shown in~\cref{fig:pipeline}. SAC achieves 45.0\% mAP under $100\times 100$ patch attacks, providing 30.6\% mAP gain upon the undefended model while maintaining the same 49.0\% clean mAP on the COCO dataset.

Besides {\it digital} domains, patch attacks have become a serious threat for object detectors in the {\it physical} world~\cite{ song2018physical, thys2019fooling, wu2020making, chen2018shapeshifter, lee2019physical, wang2021towards, 10.1145/2976749.2978392}. Developing and evaluating defenses against physical patch attacks require physical-patch datasets which are costly to create. To the best of our knowledge, APRICOT~\cite{braunegg2020APRICOT} is the only publicly available dataset of physical adversarial attacks on object detectors. However, APRICOT only provides bounding box annotations for each patch without pixel-level annotations. This hinders the development and evaluation of patch detection and removal techniques like SAC. To facilitate research in this direction, we create the {\it APRICOT-Mask} dataset, which provides segmentation masks and more accurate bounding boxes for adversarial patches in APRICOT. We train our patch segmenter with segmentation masks from APRICOT-Mask and show that SAC can effectively reduce the patch attack success rate from 7.97\% to 2.17\%. 

In summary, our contributions are as follows:
\begin{itemize}
    \item We propose Segment and Complete, a general method for defending object detectors against patch attacks via patch segmentation and a robust shape completion algorithm.
    \item We evaluate SAC on both digital and physical attacks. SAC achieves superior robustness under both non-adaptive and adaptive attacks with no reduction in performance on clean images, and generalizes well to unseen shapes, attack budgets, and unseen attack methods.
    \item We present the APRICOT-Mask dataset, which is the first publicly available dataset that provides pixel-level annotations of physical adversarial patches. 
\end{itemize}
\section{Related Work}
\subsection{Adversarial Patch Attacks}
Adversarial patch attacks are localized attacks that allow the attacker to distort a bounded region. Adversarial patch attacks were first proposed for image classifiers~\cite{brown2017adversarial, karmon2018lavan, 8578273}. 
Since then, numerous adversarial patch attacks have been proposed to fool state-of-the-art object detectors including both digital~\cite{liu2018dpatch, zhao2020object, li2018exploring, lang2021attention, saha2020role} and physical attacks~\cite{ song2018physical, thys2019fooling, wu2020making, chen2018shapeshifter, lee2019physical, wang2021towards, 10.1145/2976749.2978392}. Patch attacks for object detection are more complicated than image classification due to the complexity of the task. The attacker can use different objective functions to achieve different attack effects such as object hiding, misclassification, and spurious detection. 
\subsection{Defenses against Patch Attacks}
Many defenses have been proposed for image classifiers against patch attacks, including both empirical~\cite{10.1007/978-3-030-61638-0_31, wu2019defending, naseer2019local, hayes2018visible, gittings2020vax, rao2020adversarial} and certified defenses~\cite{Chiang*2020Certified,xiang2021patchguard, levine2020randomized}. Local gradient smoothing (LGS)~\cite{naseer2019local} is based on the observation that patch attacks introduce concentrated high-frequency noises and therefore proposes to perform gradient smoothing on regions with high gradients magnitude. Digital watermarking (DW)~\cite{hayes2018visible} finds unnaturally dense regions in the saliency map of the classifier and covers these regions to avoid their influence on classification. LGS and DW both use a similar detect and remove strategy as SAC. However, they detect patch regions based on predefined criteria, whereas SAC uses a learnable patch segmenter which is more powerful and can be combined with adversarial training to provide stronger robustness. In addition, we make use of the patch shape prior through shape completion.

In the domain of object detection, most existing defenses focus on global perturbations with the $l_p$ norm constraint~\cite{chiang2020detection, zhang2019towards, chen2021class} and only a few defenses~\cite{saha2020role, ji2021adversarial, xiang2021detectorguard} for patch attacks have been proposed. These methods are designed for a specific type of patch attack or object detector, while SAC provides a more general defense. Saha~\cite{saha2020role} proposed Grad-defense and OOC defense for defending blindness attacks where the detector is blind to a specific object category chosen by the adversary. Ji~\etal~\cite{ji2021adversarial} proposed Ad-YOLO to defend human detection patch attacks by adding a patch class on YOLOv2~\cite{redmon2017yolo9000} detector such that it detects both the objects of interest and adversarial patches. DetectorGuard (DG)~\cite{xiang2021detectorguard} is a provable defense against localized patch hiding attacks. Unlike SAC, DG does not localize or remove adversarial patches. It is an alerting algorithm that uses an objectness explainer that detects unexplained objectness for issuing alerts when attacked, while SAC solves the problem of ``detection under adversarial patch attacks" that aims to improve model performance under attacks and is not limited to hiding attacks, although the patch mask detected by SAC can also be used as a signal for issuing attack alerts.

\section{Preliminary}
\subsection{Faster R-CNN Object Detector}
In this paper, we use Faster R-CNN~\cite{10.5555/2969239.2969250} as our base object detector, though SAC is compatible with any object detector and we show the results for SSD~\cite{liu2016ssd} in the supplementary material. Faster R-CNN is a proposal-based two-stage object detector. In the first stage, a region proposal network (RPN) is used to generate class-agnostic candidate object bounding boxes called region proposals; in the second stage, a Fast R-CNN network~\cite{girshick2015fast} is used to output an object class and refine the bounding box coordinates for each region proposal. The total loss of Faster RCNN is the sum of bounding-box regression and classification losses of RPN and Fast R-CNN:
\begin{equation}
    \mathcal{L}^{\text{Faster R-CNN}}=\mathcal{L}_{\text{reg}}^{\text{RPN}} +\mathcal{L}_{\text{cls}}^{\text{RPN}} + \mathcal{L}_{\text{reg}}^{\text{Fast R-CNN}} + \mathcal{L}_{\text{cls}}^{\text{Fast R-CNN}}
\end{equation}
\subsection{Attack Formulation}
In this paper, we consider image- and location-specific untargeted patch attack for object detectors, which is strictly stronger than universal, location invariant attacks. Let $x\in [0,1]^{H\times W \times 3}$ be a clean image, where $H$ and $W$ are the height and width of $x$. We solve the following optimization problem to find an adversarial patch:
\begin{equation}
   \label{eq:attack}
    \hat{P}(x,l) = \argmax_{P\in\{P^{'}:\norm{P^{'}}_{\infty} \leq \epsilon\}}\mathcal{L}(h(A(x,l,P));y),
\end{equation}
where $h$ denotes an object detector, $A(x,l,P)$ is a ``patch applying function" that adds patch $P$ to $x$ at location $l$,  $\norm{\cdot}_{\infty}$ is $l_\infty$ norm, $\epsilon$ is the attack budget, $y$ is the ground-truth class and bounding box labels for objects in $x$, and $\mathcal{L}$ is the loss function of the object detector. We use $\mathcal{L}=\mathcal{L}^{\text{Faster R-CNN}}$ for a general attack against Faster R-CNN. We solve \cref{eq:attack} using the projected gradient descent (PGD) algorithm~\cite{madry2017towards}:
\begin{equation}
    \label{eq:maskedPGD}
    P^{t+1}=\prod_{\mathbb{P}}(P^{t} + \alpha\operatorname{sign}(\nabla_{P^{t}}\mathcal{L}(h(A(x,l,P^{t}));y))),
\end{equation}
where $\alpha$ is the step size, $t$ is the iteration number, and $\prod$ is the projection function that projects $P$ to the feasible set $\mathbb{P} = \{P:\norm{P}_{\infty} \leq \epsilon \text{ and } A(x,l,P)\in [0,1]^{H\times W \times 3} \}$. The adversarial image $x_{\text{adv}}$ is given by: $x_{\text{adv}}=A(x,l,\hat{P}(x,l))$.

We consider square patches $P\in \mathbb{R}^{s\times s\times 3}$, where $s$ is the patch size, and apply one patch per image following previous works~\cite{liu2018dpatch, lee2019physical, Chiang*2020Certified, karmon2018lavan}. We use an attack budget $\epsilon=1$ that allows the attacker arbitrarily distort pixels within a patch without a constraint, which is the case for physical patch attacks and most digital patch attacks~\cite{brown2017adversarial, liu2018dpatch, zhao2020object, li2018exploring, lang2021attention, saha2020role}. 
\section{Method}
\looseness -1 SAC defends object detectors against adversarial patch attacks through detection and removal of adversarial patches in the input image $x$. The pipeline of SAC is shown in~\cref{fig:pipeline}. It consists of two steps: first, a patch segmenter (\cref{sec:ps}) generates initial patch masks $\hat{M}_{PS}$ and then a robust shape completion algorithm (\cref{sec:sc}) is used to produce the final patch masks $\hat{M}_{SC}$. The masked image $\hat{x} = x \odot (1 - \hat{M}_{SC})$  is fed into the base object detector for prediction, where $\odot$ is the Hadamard product.

\subsection{Patch Segmentation}
\label{sec:ps}
\paragraph{Training with pre-generated adversarial images} We formulate patch detection as a segmentation problem and train a U-Net~\cite{ronneberger2015u} as the patch segmenter to provide initial patch masks. Let $PS_\theta$ be a patch segmenter parameterized by $\theta$. We first generate a set of adversarial images $\mathcal{X}_{\text{adv}}$ by attacking the base object detector with~\cref{eq:attack}, and then use the pre-generated adversarial images $\mathcal{X}_{\text{adv}}$ to train $PS_\theta$:
\begin{equation}
 \min_{\theta}\sum_{x_{\text{adv}}\in \mathcal{X}_{\text{adv}}} \mathcal{L}_{\text{BCE}}(PS_\theta(x_{\text{adv}}), M),
\end{equation}
where $M$ is the ground-truth patch mask, $PS_\theta(x_{\text{adv}})\in [0,1]^{H\times W}$ is the output probability map, and $\mathcal{L}_{\text{BCE}}$ is the binary cross entropy loss:

\begin{equation} 
\begin{split}
    \mathcal{L}_{\text{BCE}}(\hat{M}, M)=-&\sum_i^H\sum_j^W[M_{ij}\cdot \log\hat{M}_{ij} \\
    &+(1-M_{ij})\cdot \log(1-\hat{M}_{ij})].
\end{split}
\end{equation}

\paragraph{Self adversarial training} \looseness -1 Training with $\mathcal{X}_{\text{adv}}$ provides prior knowledge for $PS_\theta$ about ``how adversarial patches look like". We further propose a  self adversarial training algorithm to robustify $PS_\theta$. Specifically, we attack $PS_\theta$ to generate adversarial patch $\hat{P}_{\text{s-AT}}\in \mathbb{R}^{s\times s\times 3}$:
\begin{equation}
    \label{eq:attack_sAT}
    \hat{P}_{\text{s-AT}}(x,l) = \argmax_{P\in\{P^{'}:\norm{P^{'}}_{\infty} \leq \epsilon\}}\mathcal{L}_{\text{BCE}}(PS_\theta(A(x,l,P)), M),
\end{equation}
which is solved by PGD similar to~\cref{eq:maskedPGD}. We train $PS_\theta$ in self adversarial training by solving:  
\begin{equation}
    \label{eq:at}
    \begin{split}
      &\min_{\theta}	\left[\lambda\mathbb{E}_{x\sim\mathcal{D}}\mathcal{L}_{\text{BCE}}(PS_\theta(x), M)\right.   \\
     + & \left. (1-\lambda) \mathbb{E}_{x\sim\mathcal{D}, l\sim \mathcal{T}}\mathcal{L}_{\text{BCE}}(PS_\theta(A(x,l,\hat{P}_{\text{s-AT}}(x,l))), M)\right], 
    \end{split}
\end{equation}
where $\mathcal{T}$ is the set of allowable patch locations, $\mathcal{D}$ is the image distribution, $M$ is the ground-truth mask, and $\lambda$ controls the weights between clean and adversarial images.

One alternative is to train $PS_\theta$ with patches generated by~\cref{eq:attack}. Compared to~\cref{eq:attack}, \cref{eq:attack_sAT} does not require external labels since the ground-truth mask $M$ is determined by $l$ and known. Indeed,~\cref{eq:at} trains the patch segmenter in a manner that no external label is needed for both crafting the adversarial samples and training the model; it strengthens $PS_\theta$ to detect any ``patch-like" area in the images. Moreover, \cref{eq:attack_sAT} does not involve the object detector $h$, which makes $PS_\theta$ object-detector agnostic and speeds up the optimization as the model size of $PS_\theta$ is much smaller than $h$. 

The patch segmentation mask $\hat{M}_{PS}$ is obtained by thresholding the output of $PS_\theta$: $\hat{M}_{PS} = PS_\theta(x) > 0.5.$

\subsection{Shape Completion}
\label{sec:sc}
\subsubsection{Desired Properties}
\looseness -1 If we know that the adversary is restricted to attacking a patch of a specific shape, such as a square, we can use this information to ``fill in" the patch-segmentation output $\hat{M}_{PS}$ to cover the ground truth patch mask $M$. We adopt a conservative approach: given  $\hat{M}_{PS}$, we would like to produce an output $\hat{M}_{SC}$ which \textit{entirely} covers the true patch mask $M$. In fact, we want to guarantee this property -- however, if $\hat{M}_{PS}$ and $M$ differ arbitrarily, then we clearly cannot provide any such guarantee. Because both the ground-truth patch mask $M$ and the patch-segmentation output $\hat{M}_{PS}$ are binary vectors, it is natural to measure their difference as a Hamming distance $d_H(\hat{M}_{PS}, M)$. To provide appropriate scale, we compare this quantity to the total magnitude of the ground-truth mask $\|M\|_H := d_H(\textbf{0}, M)$. We therefore would like a patch completion algorithm with the following property:
\begin{equation}
    \label{eq:desiderata}
    \text{If } \frac{d_H(\hat{M}_{PS}, M)}{\|M\|_H} \leq \gamma \text{ then } \forall i,j: \hat{M}_{SC\,(i,j)} \geq M_{(i,j)}
\end{equation} 
\subsubsection{Proposed Method}
\begin{figure}
    \centering
    \includegraphics[width=3.25in]{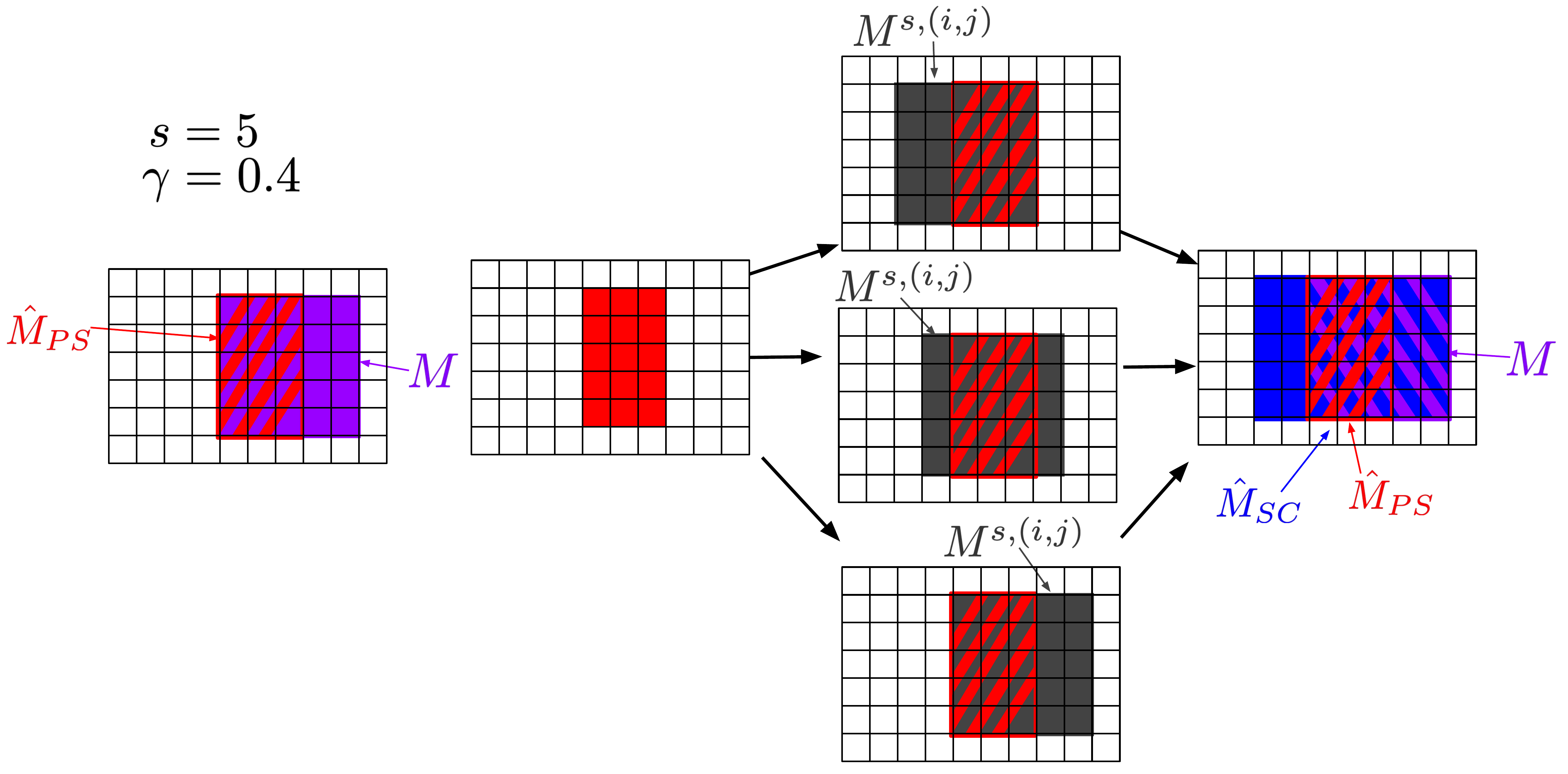}
    \caption{Construction of $\hat{M}_{SC}$ in \cref{eq:completion-simple_main}:  $\hat{M}_{SC}$ is the \textit{union} of all candidate masks $M^{s,(i,j)}$ which are $\gamma$-close to $\hat{M}_{PS}$. If $M$ is $\gamma$-close to $\hat{M}_{PS}$, this guarantees that $M$ is covered by  $\hat{M}_{SC}$.}
    \label{fig:SAC_completion_diag}
\end{figure}

If the size of the ground-truth patch is known, then we can satisfy~\cref{eq:desiderata}, \textit{minimally}, by construction. In particular, suppose that $M$ is known to be an $s \times s$ patch, and let $M^{s,(i,j)}$ refer to the mask of an $s \times s$ patch with upper-left corner at $(i,j)$. Then~\cref{eq:desiderata} is minimally satisfied by the following mask:
\begin{equation} 
    \label{eq:completion-simple_main}
    \hat{M}_{SC\,(i,j)}  := \begin{cases}
    1&\text{    if}\,\,\,\,\exists\,i',\,j':\,\,\, M^{s,(i',j')}_{(i,j)} = 1 \text{ and }   \\
    & \frac{d_H(\hat{M}_{PS}, M^{s,(i',j')})}{s^2} \leq \gamma  \\
    0&\text{otherwise.}
    \end{cases}
\end{equation}
where we have used that $\|M\|_H = s^2$. In other words, we must cover \textit{every} pixel within \textit{any} $s \times s$ patch $ M^{s,(i',j')}$ that is $\gamma$-close to the observed mask $\hat{M}_{PS}$, because any such patch may in fact be $M$: a mask consisting of only these pixels is therefore the minimal mask necessary to satisfy~\cref{eq:desiderata}. See~\cref{fig:SAC_completion_diag} for an example. While~\cref{eq:completion-simple_main} may appear daunting, there is a simple dynamic programming algorithm that allows the entire mask $\hat{M}_{SC}$ to be computed in $O(H\times W)$ time: this is presented in the supplementary material.
\subsubsection{Unknown Patch Sizes}
In~\cref{eq:completion-simple_main}, we assume that the ground-truth patch size $s$ is known; and is further parameterized by the distortion threshold $\gamma$. Let $\hat{M}_{SC}(s,\gamma)$ represent this parameterized mask, as defined in~\cref{eq:completion-simple_main}. If we do not know $s$, but instead have a set of possible patch sizes $S$ such that the true patch size $s \in S$, then we can satisfy~\cref{eq:desiderata} by simply combining all of the masks generated for each possible value of $s$:
\begin{equation} \label{eq:completion-2}
    \hat{M}_{SC}(S,\gamma)_{(i,j)} := \bigvee_{s\in S} \hat{M}_{SC}(s,\gamma)_{(i,j)} 
\end{equation}
\looseness -1 ~\cref{eq:completion-2} is indeed again the minimal mask required to satisfy the constraint: a pixel $(i,j)$ is included in  $\hat{M}_{SC}(\gamma)$ if and only if there exists some $ M^{s,(i',j')}$, for some $s \in S$, such that $(i,j)$ is part of $ M^{s,(i',j')}$ and $ M^{s,(i',j')}$ is $\gamma-$close to $\hat{M}_{PS}$. In practice, this method can be highly sensitive to the hyperparameter $\gamma$. To deal with this issue, we initially apply~\cref{eq:completion-2} with low values of $\gamma$, and then gradually increase $\gamma$ if no mask is returned -- stopping when either some mask is returned or a maximum value is reached, at which point we assume that there is no ground-truth adversarial patch. The details can be found in the supplementary material.
\subsubsection{Unknown Patch Shapes}
\label{sec:unknown_shape}
\looseness -1 In some cases, we may not know the shape of the patch. Since the patch segmenter is agnostic to patch shape, we use the union of $\hat{M}_{PS}$ and $\hat{M}_{SC}$ as the final mask output: $\hat{M}=\hat{M}_{PS}\bigcup\hat{M}_{SC}$. We empirically evaluate the effectiveness of this approach in~\cref{sec:generalization}.

\section{Evaluation on Digital Attacks}
\looseness -1 In this section, we evaluate the robustness of SAC on digital patch attacks. We consider both non-adaptive and adaptive attacks, and demonstrate the generalizability of SAC.
\subsection{Evaluation Settings}
\label{Sec:dataset}
We use COCO~\cite{lin2014microsoft} and xView~\cite{lam2018xview} datasets in our experiments. COCO is a common object detection dataset while xView is a large public dataset of overhead imagery. For each dataset, we evaluate model robustness on 1000 test images and report mean Average Precision (mAP) at Intersection over Union (IoU) 0.5. For attacking, we iterate 200 steps with a set step size $\alpha=0.01$. The patch location $l$ is randomly selected within each image. We evaluate three rounds with different random patch locations and report the mean and standard deviation of mAP. 

\subsection{Implementation Details}
\label{Sec:detail}
All experiments are conducted on a server with ten GeForce RTX 2080 Ti GPUs. For base object detectors, we use Faster-RCNN~\cite{10.5555/2969239.2969250} with feature pyramid network (RPN)~\cite{lin2017feature} and ResNet-50~\cite{he2016deep} backbone. We use the pre-trained model provided in \texttt{torchvision}~\cite{10.1145/1873951.1874254} for COCO and the model provided in \texttt{armory}~\cite{armory} for xView. For patch segmenter, we use U-Net~\cite{ronneberger2015u} with sixteen initial filters. To train the patch segmenters, for each dataset we generate 55k fixed adversarial images from the training set with patch size $100\times 100$. Training on pre-generated adversarial images took around three hours on a single GPU. For self adversarial training, we train each model for one epoch by~\cref{eq:at} using PGD attacks with 200 iterations and step size $\alpha=0.01$ with $\lambda=0.3$, which takes around eight hours on COCO and four hours on xView using ten GPUs. For patch completion, we use a square shape prior and the possible patch sizes $S = \{25,50,75,100\}$ for xView and $S = \{25,50,75,100, 125\}$ for COCO. More details can be found in the supplementary material.
\begin{table*}[h]
\centering
\caption{mAP (\%) under non-adaptive and adaptive attacks with different patch sizes. The best performance of each column is in \textbf{bold}.} 
\label{tab:non_adapt}
\begin{tabular}{c|c|c|c|c|c|c|c|c} 
\toprule
\multirow{2}{*}{Dataset} & \multirow{2}{*}{Method} & \multirow{2}{*}{Clean} & \multicolumn{3}{c|}{Non-adaptive Attack} &\multicolumn{3}{c}{Adaptive Attack}\\ \cline{4-9}

& & & 75$\times$75                       & 100$\times$100   & 125$\times$125      & 75$\times$75                       & 100$\times$100   & 125$\times$125                 \\
\midrule
\multirow{6}{*}{COCO}           & Undefended &\textbf{ 49.0} & 19.8$\pm$1.0  & 14.4$\pm$0.6  & 9.9$\pm$0.5& 19.8$\pm$1.0  & 14.4$\pm$0.6  & 9.9$\pm$0.5\\
                           
                            & AT~\cite{madry2017towards} & 40.2 &  23.5$\pm$0.7     &  18.6$\pm$0.8                           &   13.9$\pm$0.3 & 23.5$\pm$0.7     &  18.6$\pm$0.8                           &   13.9$\pm$0.3 \\ 
                            & JPEG~\cite{dziugaite2016study} & 45.6 &  39.7$\pm$0.3     &  37.2$\pm$0.3                           &   33.3$\pm$0.4 &  22.8$\pm$0.9     &  18.0$\pm$0.8                           &   13.4$\pm$0.7\\
                            & Spatial Smoothing~\cite{xu2017feature} & 46.0 &  40.4$\pm$0.6     &  38.1$\pm$0.6                           &   34.3$\pm$0.1 & 23.2$\pm$0.7     &  17.5$\pm$1.0                           &   13.5$\pm$0.6\\
                            &  LGS~\cite{naseer2019local}  &  42.7     &  36.8$\pm$0.1     &  35.2$\pm$0.6                           &   32.8$\pm$0.9   &  20.8$\pm$0.7     &  15.9$\pm$0.5                           &   12.2$\pm$0.9\\
                            &SAC (\textbf{Ours})        & \textbf{49.0} & \textbf{45.7$\pm$0.3}& \textbf{45.0$\pm$0.6} & {40.7$\pm$1.0} & \textbf{43.6$\pm$0.9} & \textbf{44.0$\pm$0.3} & \textbf{39.2$\pm$0.7}\\
\midrule
\multirow{2}{*}{Dataset} & \multirow{2}{*}{Method} & \multirow{2}{*}{Clean} & \multicolumn{3}{c|}{Non-adaptive Attack} &\multicolumn{3}{c}{Adaptive Attack}\\ \cline{4-9}

& &  & 50$\times$50 & 75$\times$75                       & 100$\times$100     & 50$\times$50     & 75$\times$75                       & 100$\times$100                  \\ 
\midrule
\multirow{6}{*}{xView}           &    Undefended & \textbf{27.2} & 8.4$\pm$1.6  & 7.1$\pm$0.4  & 5.3$\pm$ 1.1 & 8.4$\pm$1.6  & 7.1$\pm$0.4  & 5.3$\pm$ 1.1    \\
                             & AT~\cite{madry2017towards} & 22.2 &  12.1$\pm$0.4     &  8.6$\pm$0.1                           &   7.2$\pm$0.7 &  12.1$\pm$0.4     &  8.6$\pm$0.1                           &   7.15$\pm$0.7 \\
                            & JPEG~\cite{dziugaite2016study} & 23.3 &  19.3$\pm$0.4     &  17.8$\pm$1.0                           &   15.9$\pm$0.4 &  11.2$\pm$0.3     &  9.5$\pm$1.0                           &   8.3$\pm$0.3 \\
                            & Spatial Smoothing~\cite{xu2017feature} & 21.8 &  16.2$\pm$0.7     &  14.2$\pm$1.1                           &   12.4$\pm$0.8 & 11.0$\pm$0.7     &  7.9$\pm$0.6                           &   6.5$\pm$0.2  \\
                            &LGS~\cite{naseer2019local}&   19.1     &  11.9$\pm$0.5     &  10.9$\pm$0.3                           &   9.8$\pm$0.5  &  8.2$\pm$0.8    &  6.5$\pm$0.4                           &   5.4$\pm$0.5  \\
                            &SAC (\textbf{Ours})     & \textbf{27.2}& \textbf{25.3$\pm$0.3}& \textbf{23.6$\pm$1.2}& \textbf{23.2$\pm$0.3}& \textbf{24.4$\pm$0.8}& \textbf{23.0$\pm$0.9}& \textbf{22.1$\pm$0.6}\\
\bottomrule
\end{tabular}
\end{table*}
\subsection{Robustness Analysis}
\label{sec:robustness}

\subsubsection{Baselines}
\label{sec:baseline}
We compare the proposed method with vanilla adversarial training (AT),  JPEG compression~\cite{dziugaite2016study}, spatial smoothing~\cite{xu2017feature}, and LGS~\cite{naseer2019local}. For AT, we use PGD attacks with thirty iterations and step size 0.067, which takes around twelve hours per epoch on the xView training set and thirty-two hours on COCO using ten GPUs. Due to the huge computational cost, we adversarially train Faster-RCNN models for ten epochs with pre-training on clean images. More details can be found in the supplementary material.
\subsubsection{Non-adaptive Attack}
\begin{figure*}
    \centering
    \begin{subfigure}[t]{0.23\textwidth}
        \centering
        \includegraphics[width=\textwidth]{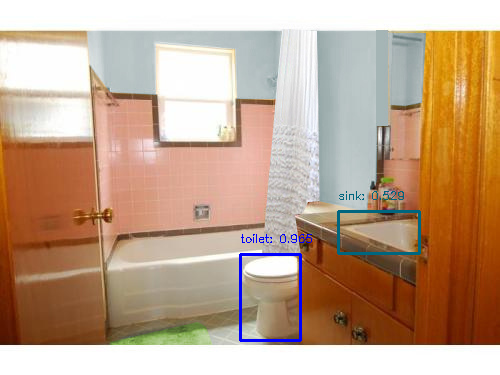}
    \end{subfigure}
    \hfill
    \begin{subfigure}[t]{0.23\textwidth}  
        \centering 
        \includegraphics[width=\textwidth]{fig/9_pred_clean.png}
    \end{subfigure}
    \hfill
    \begin{subfigure}[t]{0.23\textwidth}   
        \centering 
        \includegraphics[width=\textwidth]{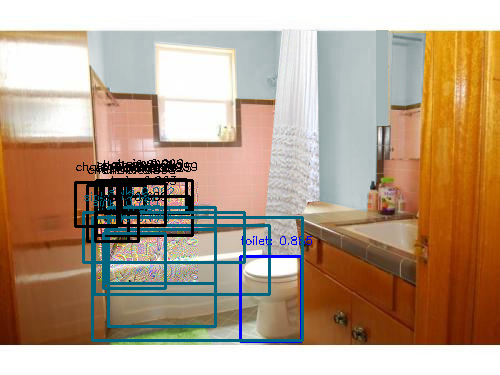}
    \end{subfigure}
    \hfill
    \begin{subfigure}[t]{0.23\textwidth}   
        \centering 
        \includegraphics[width=\textwidth]{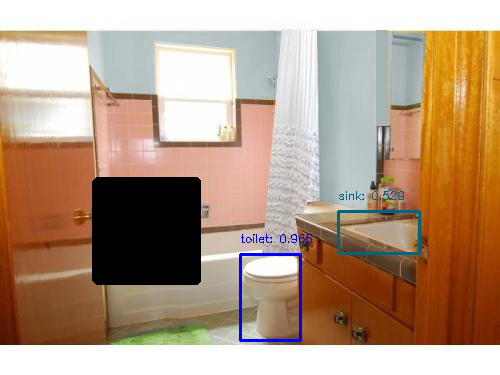}
    \end{subfigure}
    \hfill
    \begin{subfigure}[t]{0.23\textwidth}
        \centering
        \includegraphics[width=\textwidth]{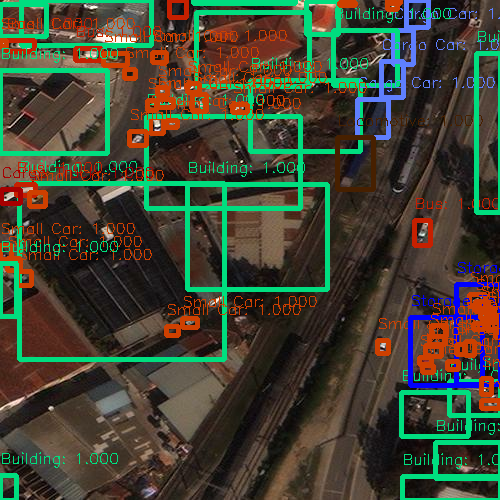}
        \caption{{\small Ground-truth on clean image.}}    
    \end{subfigure}
    \hfill
    \begin{subfigure}[t]{0.23\textwidth}  
        \centering 
        \includegraphics[width=\textwidth]{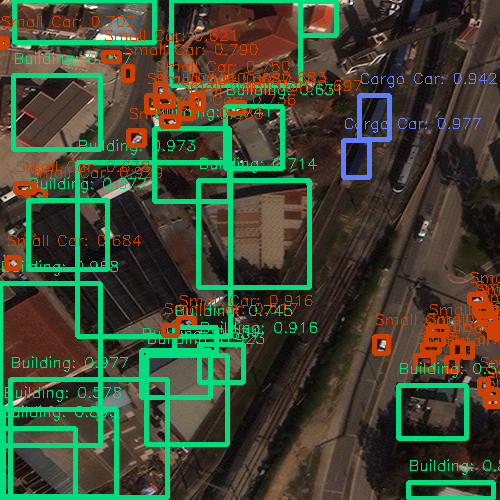}
        \caption{{\small Predictions on clean image.}}
    \end{subfigure}
    \hfill
    \begin{subfigure}[t]{0.23\textwidth}   
        \centering 
        \includegraphics[width=\textwidth]{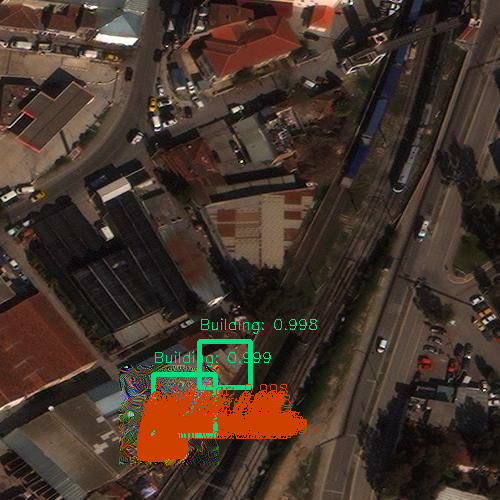}
        \caption{{\small Predictions on adversarial image with a $100 \times 100$ patch.}}
    \end{subfigure}
    \hfill
    \begin{subfigure}[t]{0.23\textwidth}   
        \centering 
        \includegraphics[width=\textwidth]{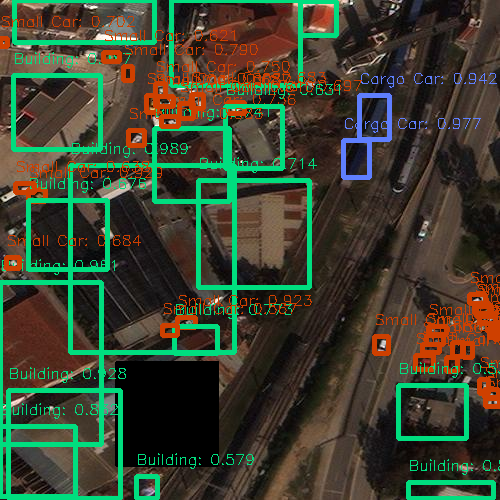}
       \caption{{\small Predictions on SAC masked image.}}
    \end{subfigure}
    \vskip\baselineskip

    \caption{Visualization of  object detection results with examples from COCO dataset (top) and xView dataset (bottom). Adversarial patches create spurious detections, and make the detector ignore the ground-truth objects. SAC masks out the patch and restores model predictions.} 
    \label{fig:coco}
\end{figure*}

\looseness -1 The defense performance under non-adaptive attacks is shown in~\cref{tab:non_adapt}, where the attacker only attacks the object detectors. SAC is very robust across different patch sizes on both datasets and has the highest mAP compared to baselines. In addition, SAC maintains high clean performance as the undefended model.~\cref{fig:coco} shows two examples of object detection results before and after SAC defense. Adversarial patches create spurious detections and hide foreground objects. SAC masks out adversarial patches and restores model predictions. We provide more examples as well as some failure cases of SAC in the supplementary material.

\subsubsection{Adaptive Attack}
\label{sec:adaptive_attack}
We further evaluate the defense performance under adaptive attacks where the adversary attacks the whole object detection pipeline. To adaptively attack preprocessing-based baselines (JPEG compression, spatial smoothing, and LGS), we use BPDA~\cite{athalye2018obfuscated} assuming the output of each defense approximately equals to the original input. To adaptively attack SAC, we use straight-through estimators (STE)~\cite{bengio2013estimating} when backpropagating through the thresholding operations, which is the strongest adaptive attack we have found for SAC (see the supplementary material for details). The results are shown in ~\cref{tab:non_adapt}. The performances of preprocessing-based baselines drop a lot under adaptive attacks. AT achieves the strongest robustness among the baselines while sacrificing clean performance. The robustness of SAC has little drop under adaptive attacks and significantly outperforms the baselines. Since adaptive attacks are stronger than non-adaptive attacks, we only use adaptive attacks for the rest of the experiments.

\subsubsection{Generalizability of SAC}
\label{sec:generalization}
\begin{figure}[h]
    \centering
    \begin{subfigure}[b]{0.23\textwidth}   
    \centering 
    \includegraphics[width=\textwidth]{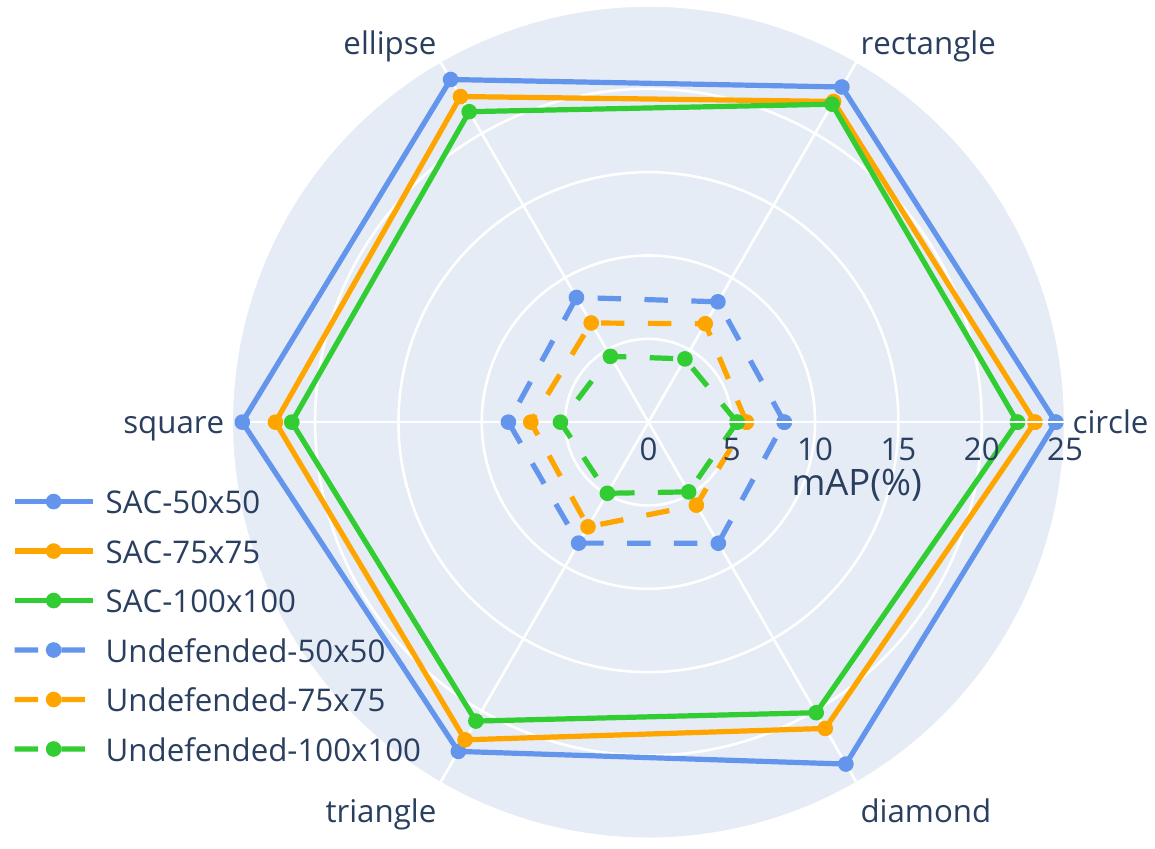}
    \caption{xView dataset.}
    \end{subfigure}
    \hfill
    \begin{subfigure}[b]{0.23\textwidth}
    \centering
    \includegraphics[width=\textwidth]{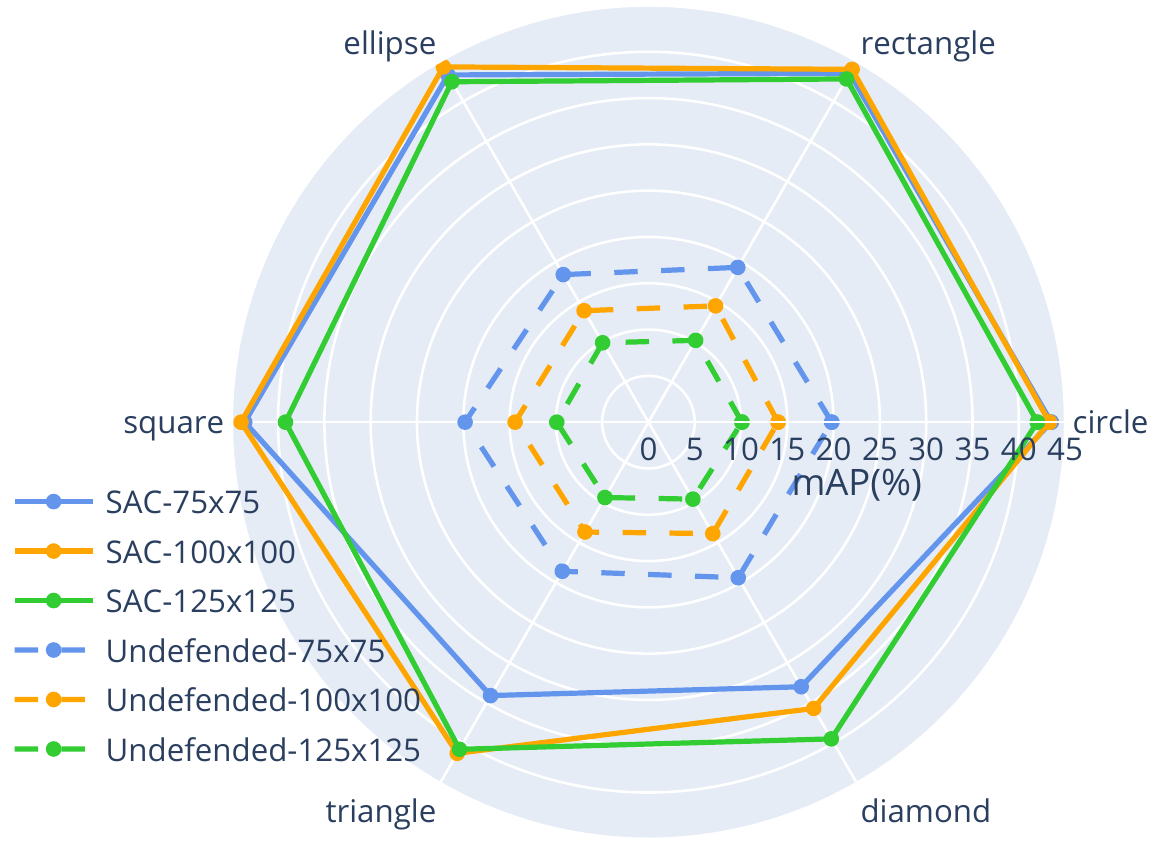}
    \caption{COCO dataset. }
    \end{subfigure}

    \caption{Performance of SAC under adaptive attacks with different patch shapes and sizes. SAC demonstrates strong robustness under rectangle, circle and ellipse patch attacks, even though these shapes mismatch with the square shape prior used in SAC.}
    \label{fig:unseen_shapes}
\end{figure}

\paragraph{Generalization to unseen shapes} We train the patch segmenter with square patches and use the square shape prior in shape completion. Since adversarial patches may not always be square in the real world, we further evaluate square-trained SAC with adversarial patches of different shapes while fixing the number of pixels in the patch. The details of the shapes used can be found in the supplementary material. We use the union of $\hat{M}_{PS}$ and $\hat{M}_{SC}$ as described in~\cref{sec:unknown_shape}. The results are shown in~\cref{fig:unseen_shapes}. SAC demonstrates strong robustness under rectangle, circle, diamond, triangle, and ellipse patch attacks, even though these shapes mismatch with the square shape prior used in SAC. 

\begin{figure}
    \centering
    \centering
    \includegraphics[width=0.45\textwidth]{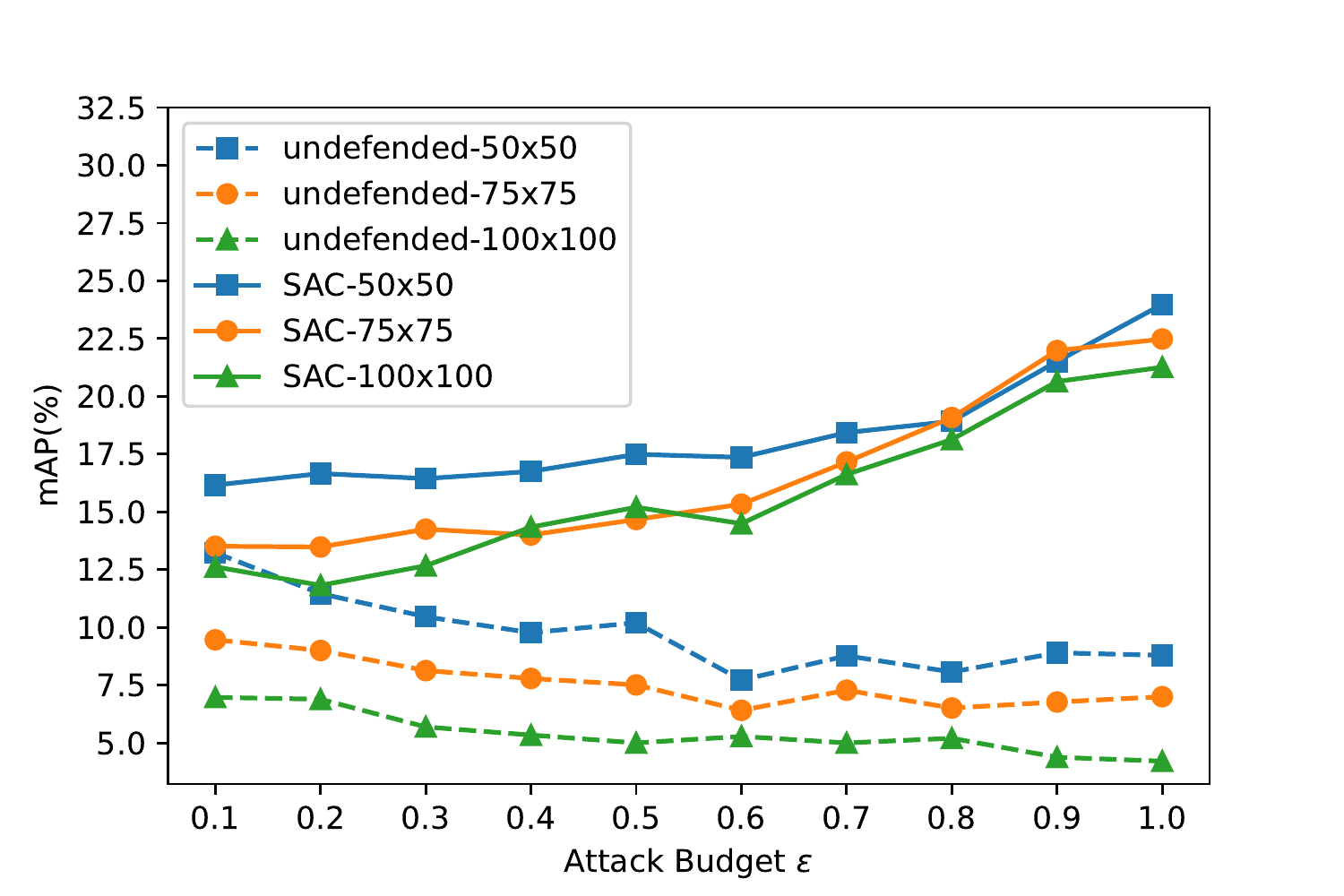}
    \caption{SAC performance under different attack budgets on xView dataset. SAC is trained with $\epsilon=1$.}   
    \label{fig:budget}
\end{figure}

\paragraph{Generalization to attack budgets} In~\cref{eq:attack}, we set $\epsilon=1$ that allows the attacker to arbitrarily modify the pixel values within the patch region. In practice, the attacker may lower the attack budget to generate less visible adversarial patches to evade patch detection in SAC. To test how SAC generalizes to lower attack budgets, we evaluate SAC trained on $\epsilon=1$ under lower $\epsilon$ values on the xView dataset. We set iteration steps to 200 and step size to $\epsilon/200$.~\cref{fig:budget} shows that SAC remains robust under a wide range of $\epsilon$. Although the performance of SAC degrades when $\epsilon$ becomes smaller because patches become imperceptible, SAC still provides significant robustness gain upon undefended models. In addition, SAC is flexible and we can use smaller $\epsilon$ in training to provide better protection against imperceptible patches.
\paragraph{Generalization to unseen attack methods} In the previous sections, we use PGD (\cref{eq:maskedPGD}) to create adversarial patches. We further evaluate SAC under unseen attack methods, including DPatch~\cite{liu2018dpatch} and MIM~\cite{dong2018boosting} attack. We use 200 iterations for both attacks, and set the learning rate to 0.01 for DPatch and decay factor $\mu=1.0$ for MIM. The performance is shown in~\cref{tab:unseen_attack}. SAC achieves more than 40.0\% mAP on COCO and 21\% mAP on xView under both attacks, providing strong robustness upon undefended models. 
\begin{table}[h]
\centering
\setlength{\tabcolsep}{2pt}
\caption{mAP (\%) under adaptive unseen attack methods with different patch sizes.} 
\label{tab:unseen_attack}
\begin{tabular}{c|ccccc} 
\toprule
& Attack & Method & 75$\times$75                       & 100$\times$100   & 125$\times$125                       \\ 
\midrule
\multirow{4}{*}{\rotatebox{90}{COCO}}           & \multirow{2}{*}{DPatch~\cite{liu2018dpatch}}  & Undefended & 33.6$\pm$0.8  & 29.1$\pm$0.6  & 25.0$\pm$1.7\\
                           
                            & &SAC (\textbf{Ours})        & 45.3$\pm$0.3 & 44.1$\pm$0.6 & 42.1$\pm$0.8 \\ 
                            & \multirow{2}{*}{MIM~\cite{dong2018boosting}}  & Undefended & 20.1$\pm$1.2  & 14.2$\pm$0.8  & 10.5$\pm$0.2\\
                           
                            & &SAC (\textbf{Ours})        & 42.2$\pm$0.9 & 43.5$\pm$1.0 & 40.0$\pm$0.2 \\
\midrule
& Attack & Method  & 50$\times$50                       & 75$\times$75                       & 100$\times$100   \\ 
\midrule
\multirow{4}{*}{\rotatebox{90}{xView}}          & \multirow{2}{*}{DPatch~\cite{liu2018dpatch}}  & Undefended & 16.0$\pm$0.5  & 13.4$\pm$0.9  & 11.1$\pm$0.9\\
                            & &SAC (\textbf{Ours})        & 25.3$\pm$0.5 & 22.7$\pm$1.1 & 21.8$\pm$0.5 \\
                            & \multirow{2}{*}{MIM~\cite{dong2018boosting}}  & Undefended & 8.3$\pm$0.4  & 7.3$\pm$0.8  & 6.5$\pm$1.5\\
                           
                            & &SAC (\textbf{Ours})        & 24.7$\pm$0.7 & 23.0$\pm$0.9 & 22.1$\pm$0.6 \\
\bottomrule
\end{tabular}
\end{table}
\subsection{Ablation Study}
\label{sec:ablation}
In this section, we investigate the effect of each component of SAC. We consider three models: 1) patch segmenter trained with pre-generated adversarial images (PS); 2) PS further trained with self adversarial training (self AT); 3) Self AT trained PS combining with shape completion (SC), which is the whole SAC defense. The performance of these models under adaptive attacks is shown in~\cref{tab:ablation}. PS alone achieves good robustness under adaptive attacks (comparable or even better performance than the baselines in~\cref{tab:non_adapt}) thanks to the inherent robustness of segmentation models~\cite{10.5555/3295222.3295441, arnab2018robustness}. Self AT significantly boosts the robustness, especially on the COCO dataset. SC further improves the robustness. Interestingly, we find that adaptive attacks on models with SC would force the attacker to generate patches that have more structured noises trying to fool SC (see supplementary material).

\begin{table}
\centering
\caption{mAP (\%) under adaptive attacks of ablated models. }
\label{tab:ablation}
\begin{tabular}{c|lccc} 
\toprule
& Method & 75$\times$75                       & 100$\times$100   & 125$\times$125 \\
\midrule
\multirow{4}{*}{\rotatebox{90}{COCO}} &Undefended & 19.8$\pm$1.0  & 14.4$\pm$0.6  & 9.9$\pm$0.5\\
&PS       & 23.3$\pm$0.7 & 18.7$\pm$0.3 & 13.1$\pm$0.3  \\
&~~+ self AT  & 41.5$\pm$0.2 & 40.5$\pm$0.6 & 36.6$\pm$0.1  \\
&~~~~+ SC  & \textbf{43.6$\pm$0.9} & \textbf{44.0$\pm$0.3} & \textbf{39.2$\pm$0.7} \\
\midrule
 & Method & 50$\times$50 & 75$\times$75                       & 100$\times$100    \\
\midrule
\multirow{4}{*}{\rotatebox{90}{xView}} & Undefended & 8.4$\pm$1.6  & 7.1$\pm$0.4  & 5.3$\pm$ 1.1    \\ 
&PS        & 16.8$\pm$0.6 &13.6$\pm$0.4 &11.1$\pm$0.3\\
&~~+ self AT  & 20.6$\pm$0.4 &17.6$\pm$0.5 &15.4$\pm$0.6  \\
&~~~~+ SC  & \textbf{24.4$\pm$0.8}& \textbf{23.0$\pm$0.9} & \textbf{22.1$\pm$0.6}  \\
\bottomrule
\end{tabular}
\end{table}
\section{Evaluation on Physical Attack}
In this section, we evaluate the robustness of SAC on physical patch attacks. We first introduce the APRICOT-Mask dataset and further demonstrate the effectiveness of SAC on the APRICOT dataset.
\subsection{APRICOT-Mask Dataset}
\label{sec:APRICOT-mask}
\begin{figure}[htbp]
    \centering
    \begin{subfigure}[t]{0.18\textwidth}
        \centering
        \includegraphics[width=\textwidth]{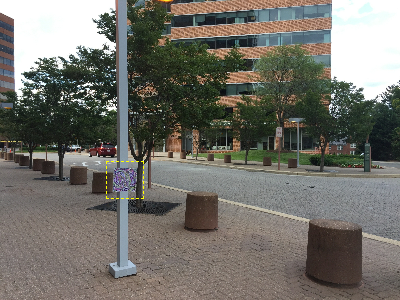}
    \end{subfigure}
    \hfill
    \begin{subfigure}[t]{0.135\textwidth}  
        \centering 
        \includegraphics[width=\textwidth]{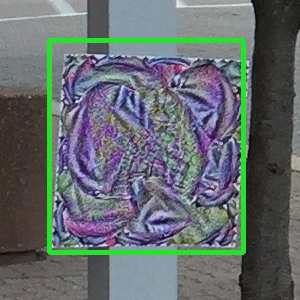}
    \end{subfigure}
    \hfill
    \begin{subfigure}[t]{0.135\textwidth}  
        \centering 
        \includegraphics[width=\textwidth]{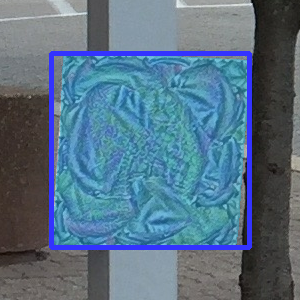}
    \end{subfigure}
    \vskip-0.01\baselineskip 
        \begin{subfigure}[t]{0.18\textwidth}
        \centering
        \includegraphics[width=\textwidth]{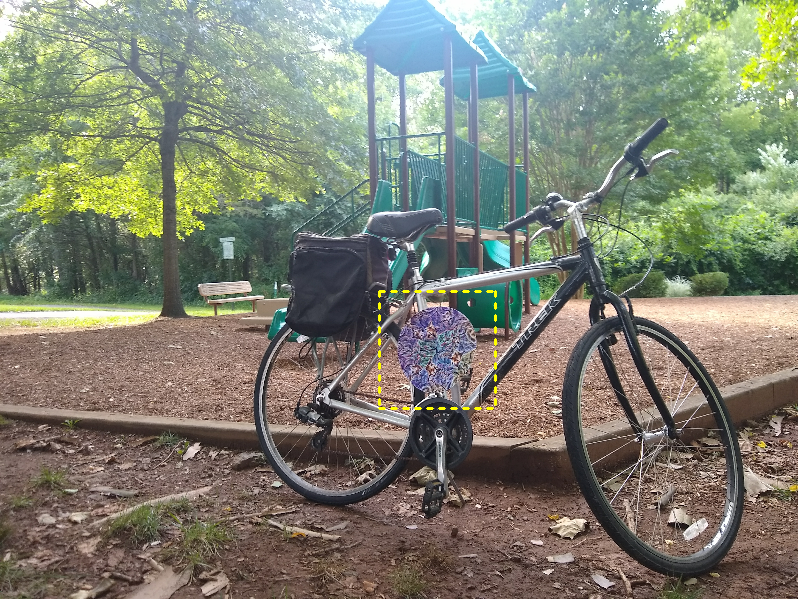}
    \end{subfigure}
    \hfill
    \begin{subfigure}[t]{0.135\textwidth}  
        \centering 
        \includegraphics[width=\textwidth]{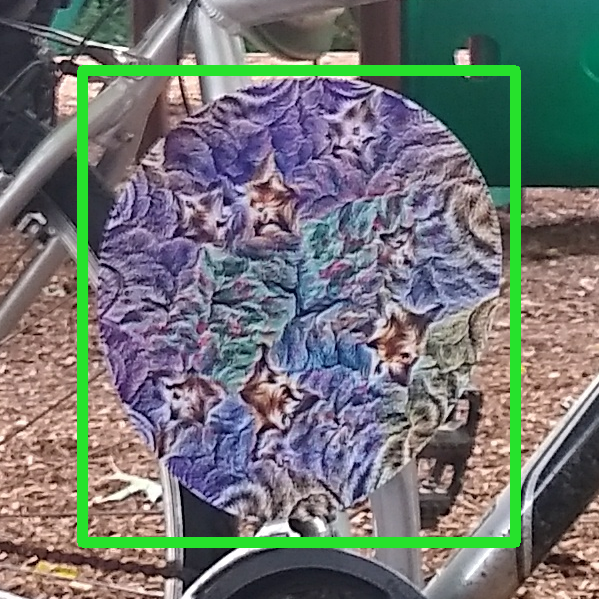}
    \end{subfigure}
    \hfill
    \begin{subfigure}[t]{0.135\textwidth}  
        \centering 
        \includegraphics[width=\textwidth]{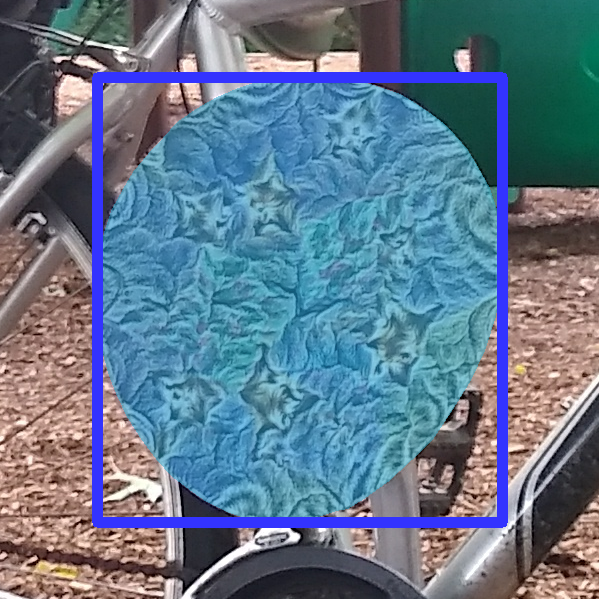}
    \end{subfigure}
    \vskip-0.01\baselineskip 
    \caption{Images and patch annotations from the APRICOT and APRICOT-Mask datasets. Left: adversarial images from the APRICOT dataset; middle: patch bounding boxes provided by the APRICOT dataset; right: patch bounding boxes and segmentation masks provided by the APRICOT-Mask dataset.} 
    \label{fig:apricot}
\end{figure}

\looseness -1 APRICOT~\cite{braunegg2020APRICOT} contains 1,011 images of sixty unique physical adversarial patches photographed in the real world, of which six patches (138 photos) are in the development set, and the other fifty-four patches (873 photos) are in the test set. APRICOT provides bounding box annotations for each patch. However, there is no pixel-level annotation of the patches. We present the APRICOT-Mask dataset \footnote{\url{https://aiem.jhu.edu/datasets/apricot-mask}}, which provides segmentation masks and more accurate bounding boxes for adversarial patches in the APRICOT dataset (see two examples in~\cref{fig:apricot}). The segmentation masks are annotated by three annotators using Labelbox~\cite{labelbox} and manually reviewed to ensure the annotation quality. The bounding boxes are then generated automatically from the segmentation masks. We hope APRICOT-Mask along with the APRICOT dataset can facilitate the research in building defenses against physical patch attacks, especially patch detection and removal techniques.

\subsection{Robustness Evaluation}
\label{sec:APRICOT-rob}
\begin{figure}
    \centering
    \centering
    \includegraphics[width=0.35\textwidth]{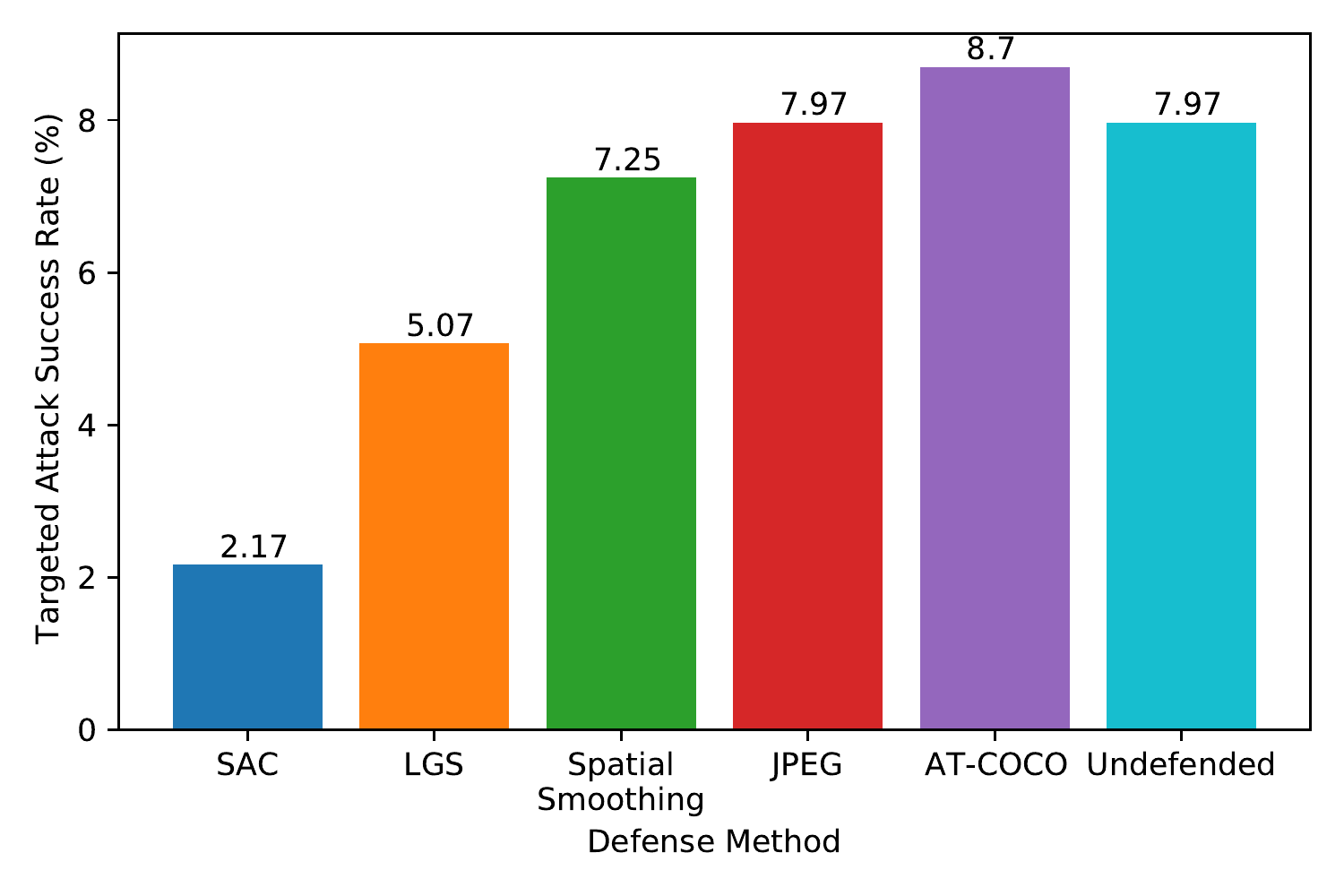}
    \caption{Targeted attack success rates on the APRICOT dataset.}   
    \label{fig:apricot_success_rate}
\end{figure}

\paragraph{Evaluation Metrics} We evaluate the defense effectiveness by the targeted attack success rate. A patch attack is ``successful" if the object detector generates a detection that overlaps a ground truth adversarial patch bounding box with an IoU of at least 0.10, has a confidence score greater than 0.30, and is classified as the same object class as the patch's target~\cite{braunegg2020APRICOT}. 
\paragraph{Evaluation Results} We train the patch segmenter on the APRICOT test set using the segmentation masks from the APRICOT-Mask dataset. The training details can be found in the supplementary material. Since APRICOT patches are generated from three detection models trained on the COCO dataset targeting ten COCO object categories, we use a Faster-RCNN model pretrained on COCO~\cite{10.1145/1873951.1874254} as our base object detector, which is a black-box attack setting with target and substitute models trained on the same dataset. We evaluate the targeted attack success rate on the development set and compare SAC with the baselines as in~\cref{sec:baseline}. For AT, we use the Faster-RCNN model adversarially trained on the COCO dataset (AT-COCO) as the size of the APRICOT dataset is not enough to retrain an object detector. The results are shown in~\cref{fig:apricot_success_rate}. SAC significantly brings down the targeted attack success rate of the undefended model from 7.97\% to 2.17\%, which is the lowest among all defense methods. AT has a slightly higher targeted attack success rate than the undefended model, which may be due to the domain gap between COCO and APRICOT datasets.

\section{Discussion and Conclusion}
In this paper, we propose the Segment and Complete defense that can secure any object detector against patch attacks by robustly detecting and removing adversarial patches from input images. We train a robust patch segmenter and exploit patch shape priors through a shape completion algorithm. Our evaluation on digital and physical attacks demonstrates the effectiveness of 
SAC. In addition, we present the APRICOT-Mask dataset to facilitate the research in building defenses against physical patch attacks. 

SAC can be improved in several ways. First, although SAC does not require re-training of base object detectors, fine-tuning them on images with randomly-placed black blocks can further improve their performance on SAC masked images. Second, in this paper, we adopt a conservative approach that masks out the entire patch region after we detect the patch. This would not cause information loss when the attacker is allowed to arbitrarily distort the pixels and destroy all the information within the patch such as in physical patch attacks. However, in the case where the patches are less visible, some information may be preserved in the patched area. Instead of masking out the patches, one can potentially impaint or reconstruct the content within the patches, which can be the future direction of this work. 
\paragraph{\textbf{Acknowledgment}} This work was supported by the DARPA GARD Program HR001119S0026-GARD-FP-052.

{\small
\bibliographystyle{ieee_fullname}
\bibliography{main}
}

\newpage
\appendix
\section*{Supplementary Material}
\section{Baselines Details}
For JPEG compression~\cite{dziugaite2016study}, we set the quality parameter to 50. For spatial smoothing, we use window size 3. For LGS~\cite{naseer2019local}, we set the block size to 30, overlap to 5, threshold to 0.1, and smoothing factor to 2.3. For AT, we use PGD attacks with 30 iterations and step size 0.067, which takes around twelve hours per epoch on the xView training set and thirty-two hours on COCO using ten GPUs. We use SGD optimizers with an initial learning rate of 0.01, momentum 0.9, weight decay $5\times 10^{-4}$, and batch size 10. We train each model with ten epochs. There is a possibility that the AT models would perform better if we train them longer or tune the training hyper-parameters. However, we were unable to do so due to the extremely expensive computation needed.
\section{SAC Details}
\subsection{Training the Patch Segmenter}
\paragraph{COCO and xView datasets} We use U-Net~\cite{ronneberger2015u} with 16 initial filters as the patch segmenter on the COCO and xView datasets. To train the patch segmenters, for each dataset we generate 55k fixed adversarial images from the training set with a patch size $100\times 100$ by attacking base object detectors, among which 50k are used for training and 5k for validation. We randomly replace each adversarial image with its clean counterpart with a probability of 30\% during training to ensure good performance on clean data. All images are cropped to squares and resized to $500\times500$ during training. We use RMSprop~\cite{hinton2012neural} optimizer with an initial learning rate of $10^{-4}$, momentum 0.9, weight decay $10^{-8}$, and batch size 16. We train patch segmenters for five epochs and evaluate them on the validation set five times in each epoch. We reduce the learning rate by a factor of ten if there is no improvement after two evaluations. For self adversarial training, we train each model for one epoch with $\lambda=0.3$ using PGD attacks with 200 iterations and step size $\alpha=0.01$, which takes around eight hours on COCO and four hours on xView using ten GPUs. 
\paragraph{APRICOT dataset} Detecting adversarial patches in the physical world can be more challenging, as the shape and appearance of patches can vary a lot under different viewing angles and lighting conditions. For the APRICOT dataset, We use U-Net~\cite{ronneberger2015u} with 64 initial filters as the patch segmenter. We downscale each image by a factor of two during training and evaluation to save memory as each image is approximately 12 megapixels (\eg, $4000\times 3000$ pixels). We use 85\% of the APRICOT test set (742 images) as the training set, and the rest (131 images) as the validation set. During training, we randomly crop $500\times 500$ image patches from the downscale images, with a probability of $60\%$ that an image patch contain an adversarial patch and $40\%$ that it contain no patch. We use RMSprop~\cite{hinton2012neural} optimizer with an initial learning rate of $10^{-3}$, momentum 0.9, weight decay $10^{-8}$, and batch size 24. We train patch segmenters for 100 epochs, and reduce the learning rate by a factor of ten if the dice score on the validation set has no improvement after 10 epochs. After training, we pick the checkpoint that has the highest dice score on the validation set as our final model. The training takes around four hours on six gpus.

\subsection{Different Patch Shapes for Evaluating SAC}
\begin{figure}
\begin{subfigure}{.11\textwidth}
  \centering
  \includegraphics[width=\linewidth]{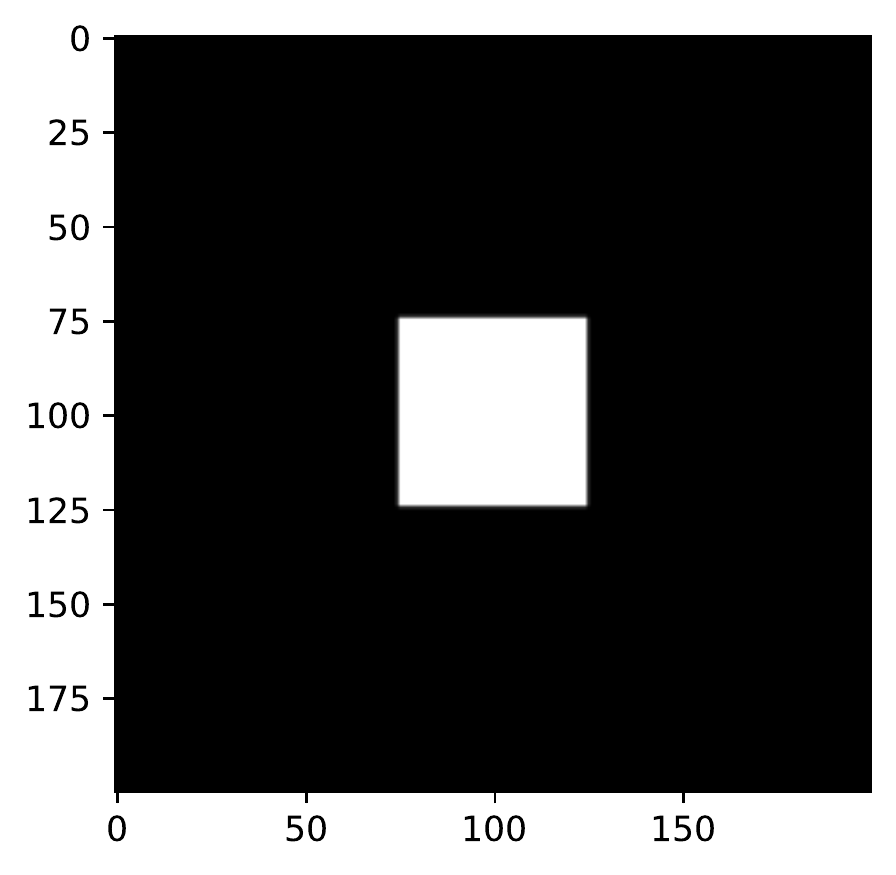}  
\end{subfigure}
\hfill
\begin{subfigure}{.11\textwidth}
  \centering
  \includegraphics[width=\linewidth]{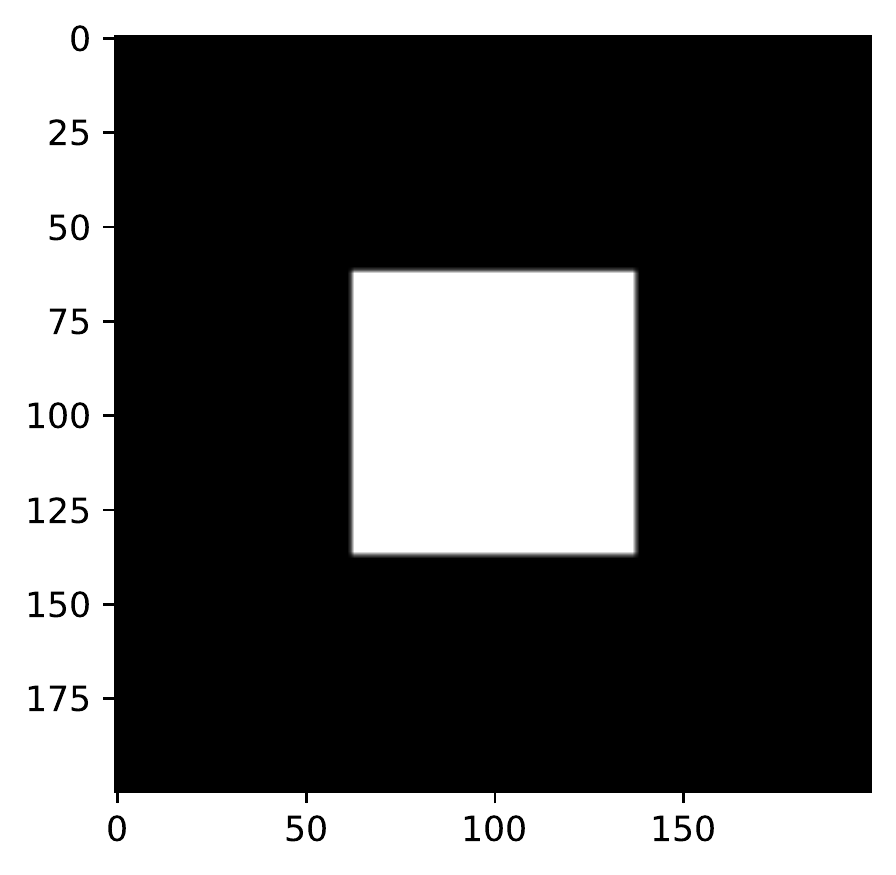}  
\end{subfigure}
\hfill
\begin{subfigure}{.11\textwidth}
  \centering
  \includegraphics[width=\linewidth]{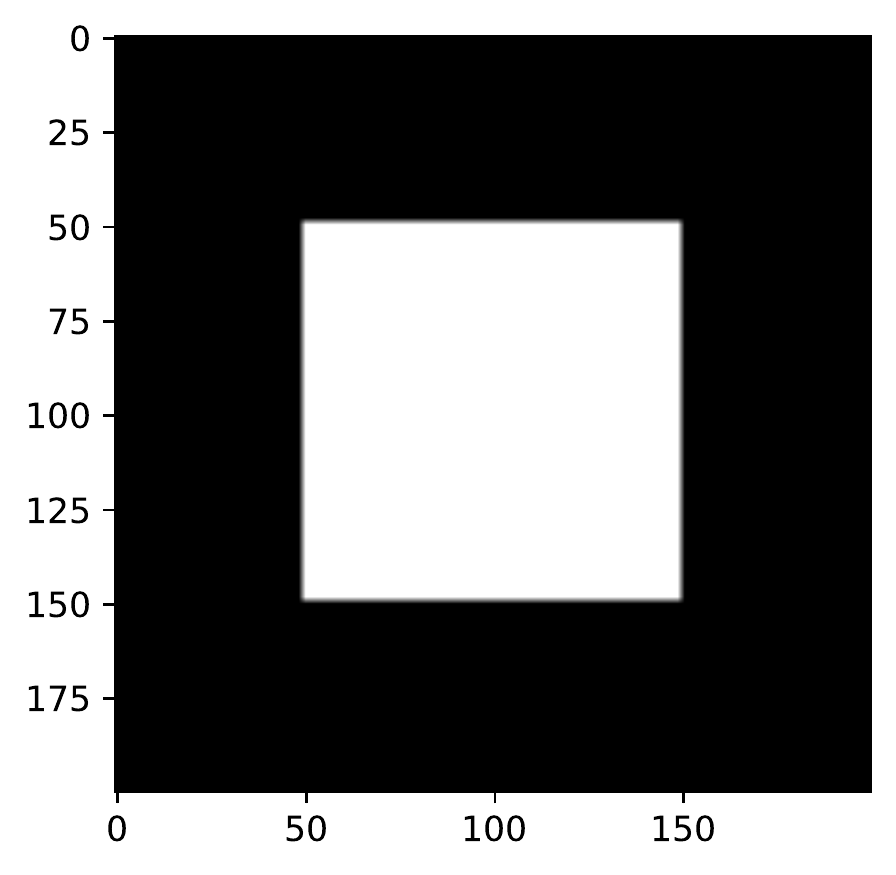}  
\end{subfigure}
\hfill
\begin{subfigure}{.11\textwidth}
  \centering
  \includegraphics[width=\linewidth]{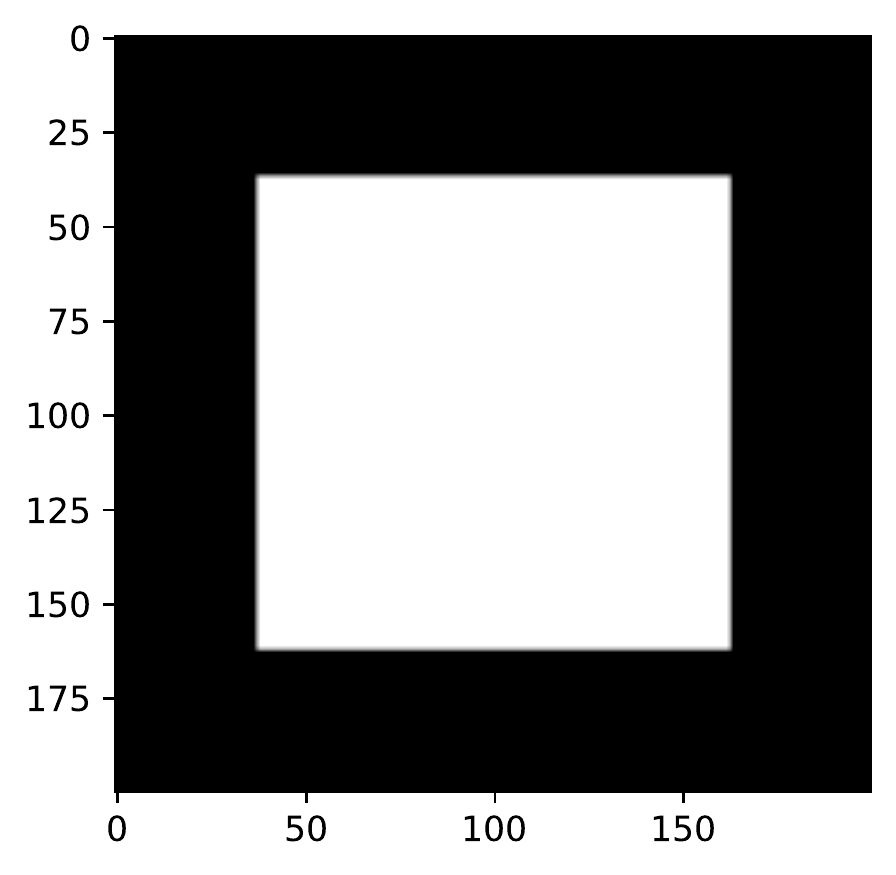}  
\end{subfigure}
\begin{subfigure}{.11\textwidth}
  \centering
  \includegraphics[width=\linewidth]{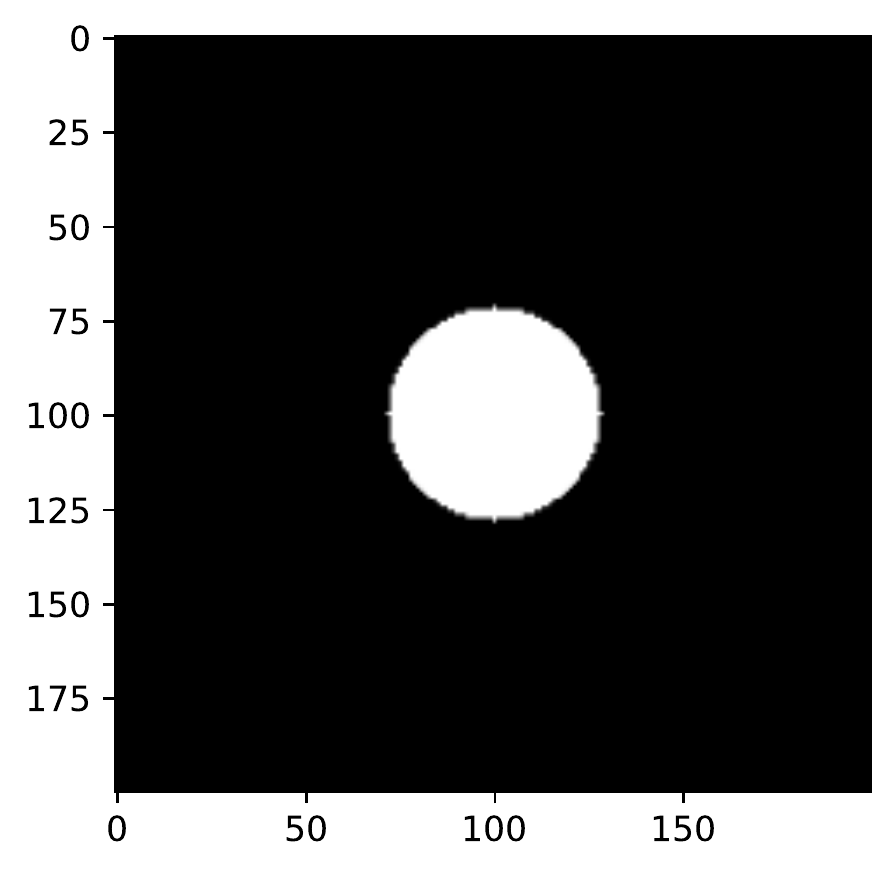}  
\end{subfigure}
\hfill
\begin{subfigure}{.11\textwidth}
  \centering
  \includegraphics[width=\linewidth]{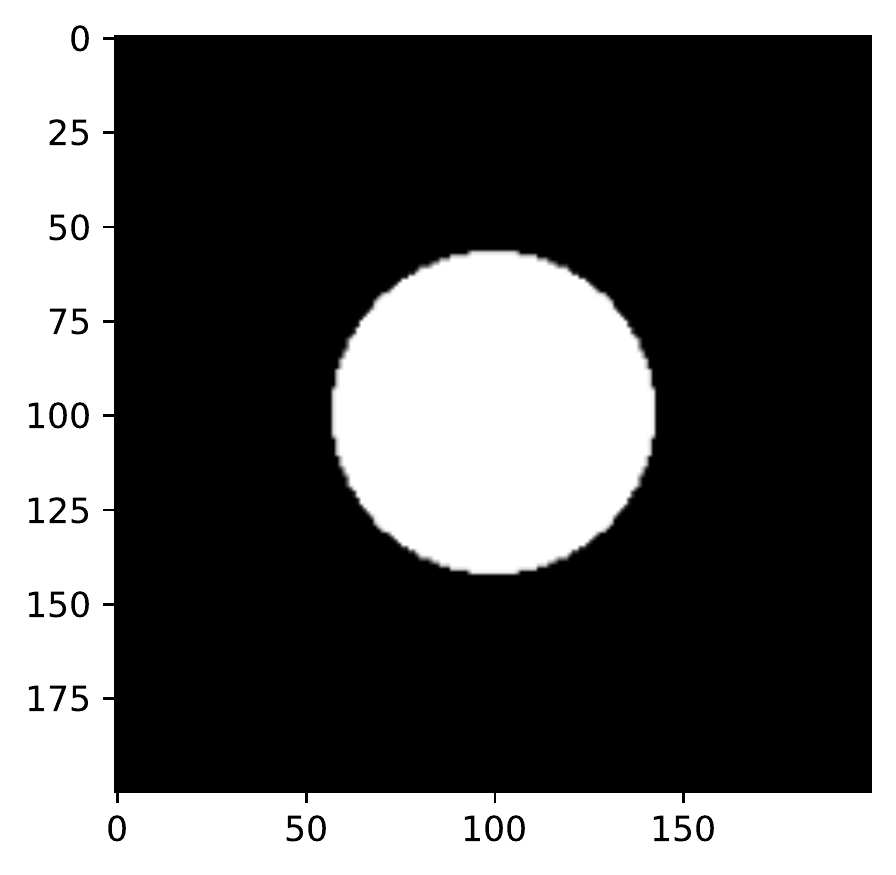}  
\end{subfigure}
\hfill
\begin{subfigure}{.11\textwidth}
  \centering
  \includegraphics[width=\linewidth]{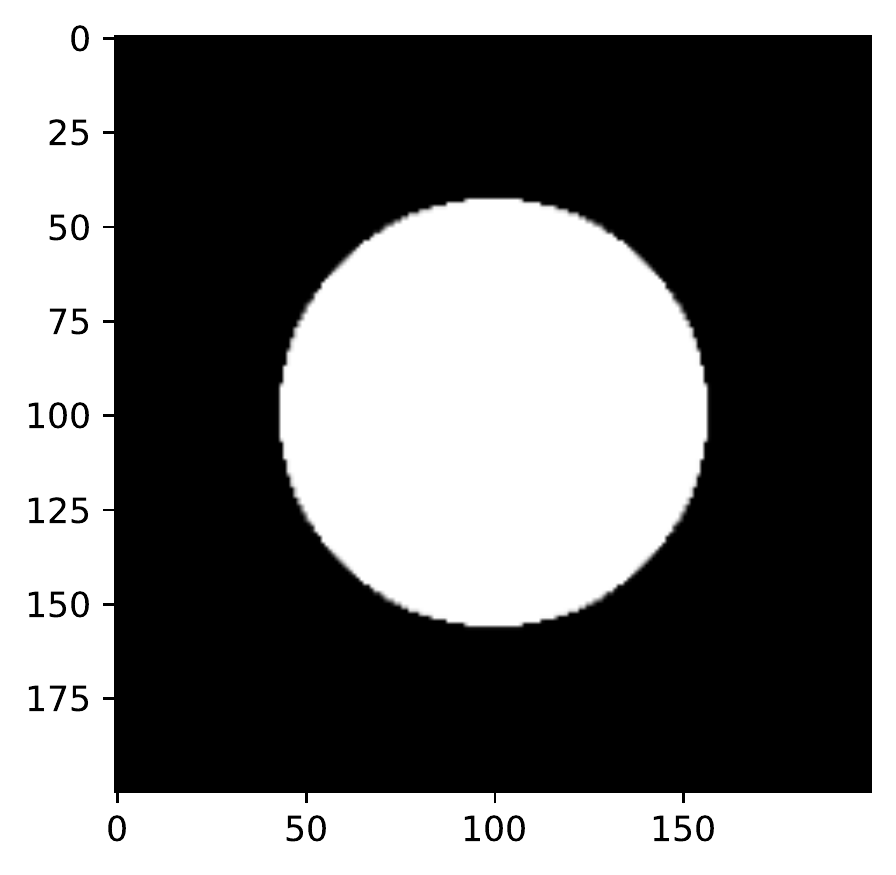}  
\end{subfigure}
\hfill
\begin{subfigure}{.11\textwidth}
  \centering
  \includegraphics[width=\linewidth]{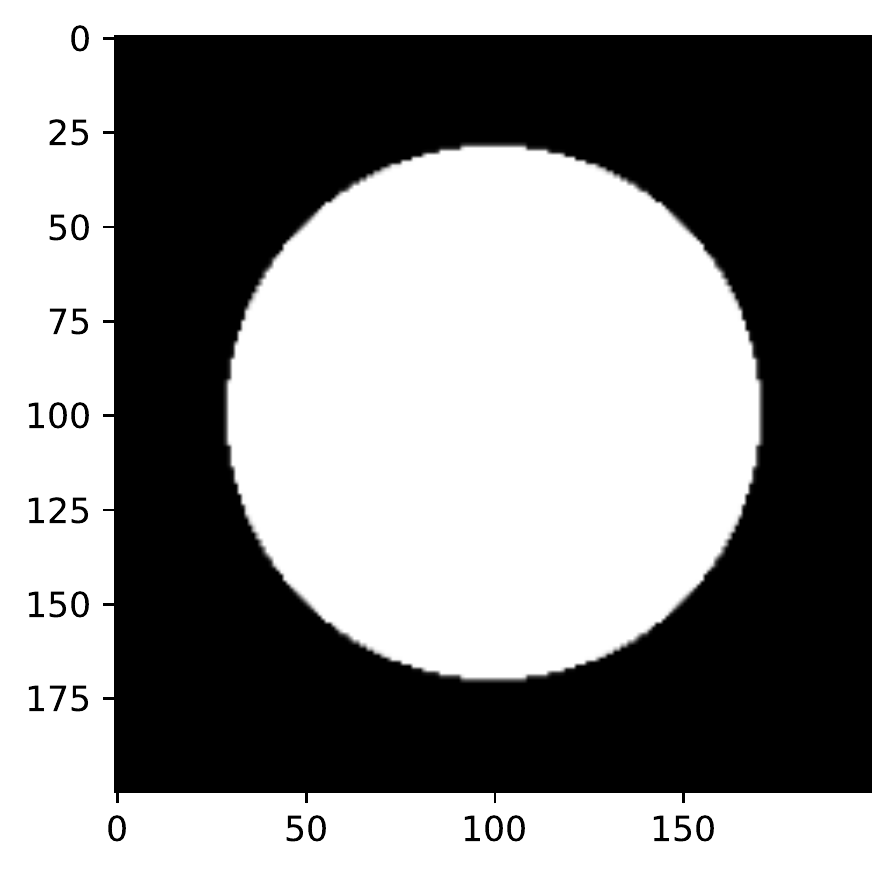}  
\end{subfigure}
\begin{subfigure}{.11\textwidth}
  \centering
  \includegraphics[width=\linewidth]{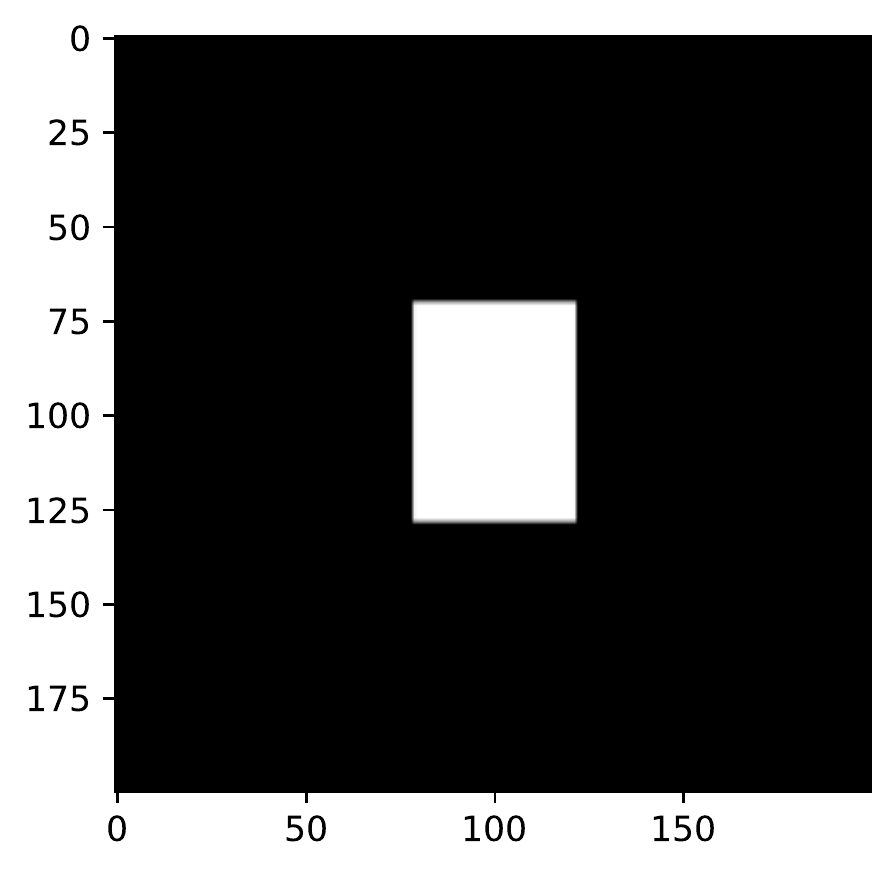}  
\end{subfigure}
\hfill
\begin{subfigure}{.11\textwidth}
  \centering
  \includegraphics[width=\linewidth]{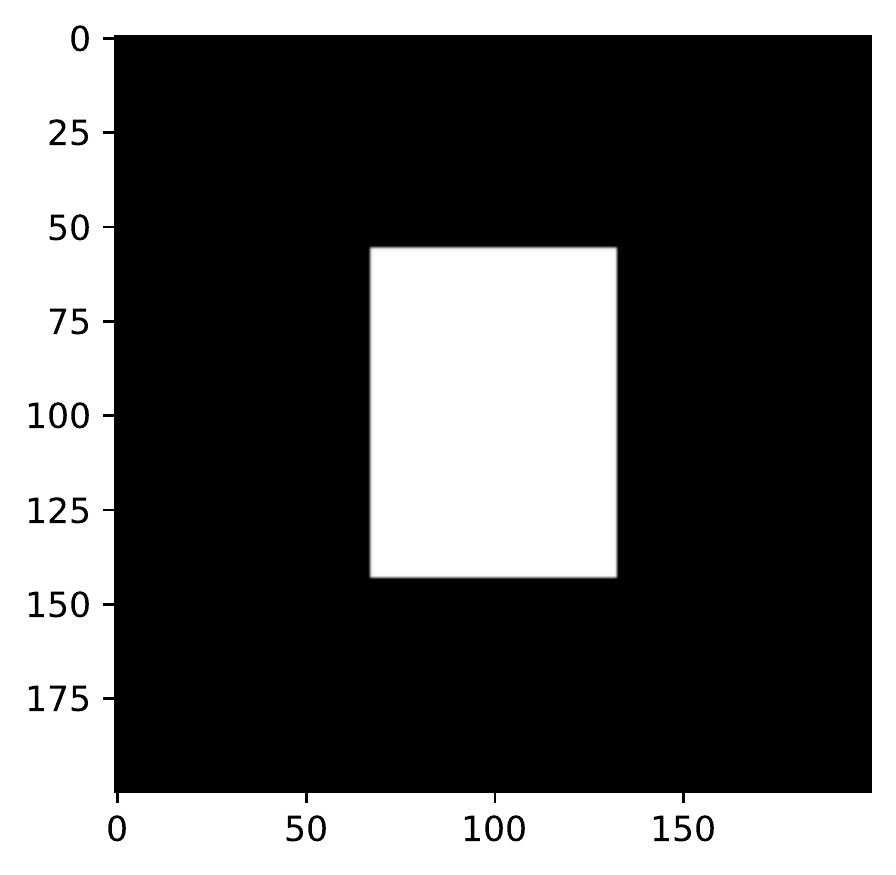}  
\end{subfigure}
\hfill
\begin{subfigure}{.11\textwidth}
  \centering
  \includegraphics[width=\linewidth]{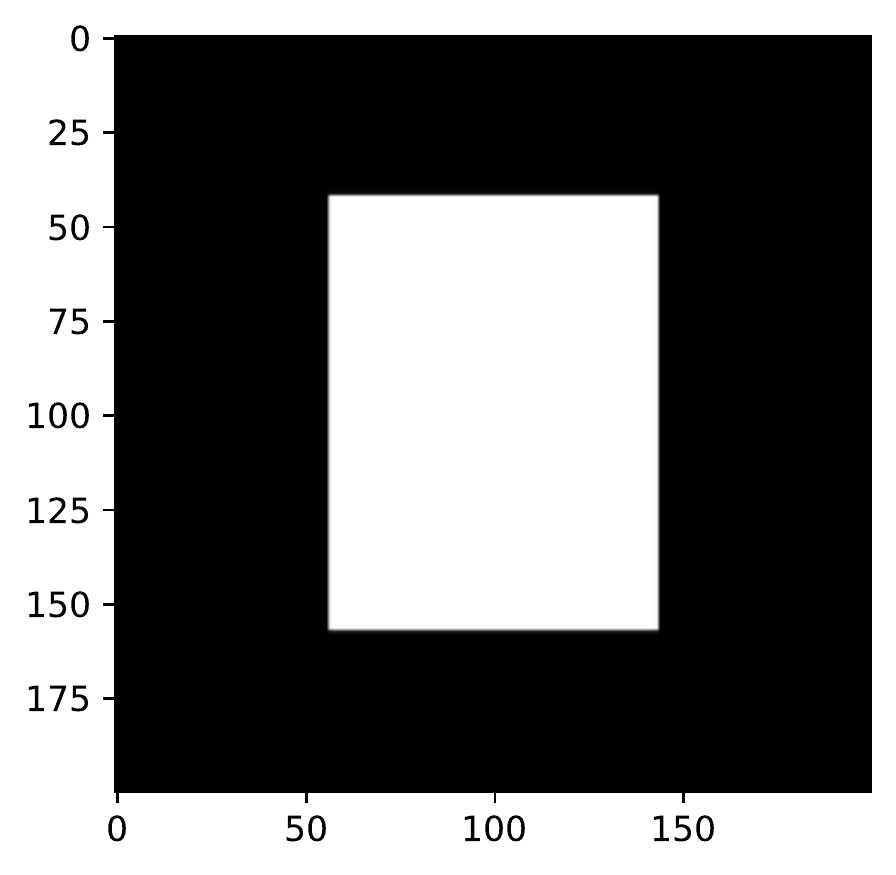}  
\end{subfigure}
\hfill
\begin{subfigure}{.11\textwidth}
  \centering
  \includegraphics[width=\linewidth]{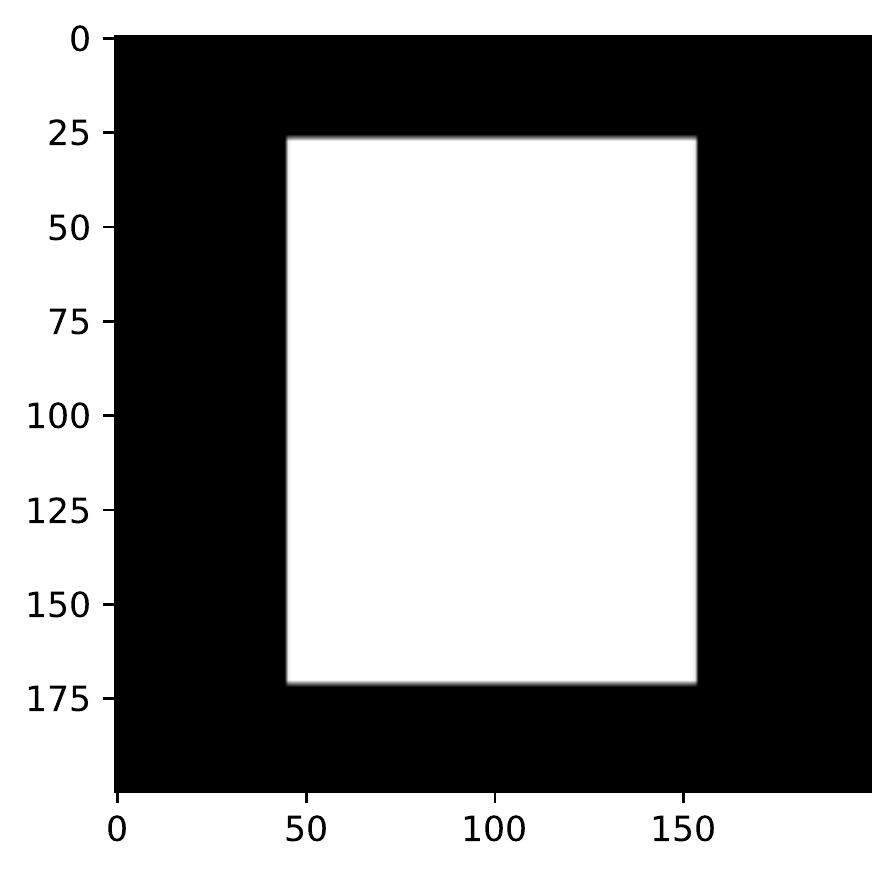}  
\end{subfigure}
\begin{subfigure}{.11\textwidth}
  \centering
  \includegraphics[width=\linewidth]{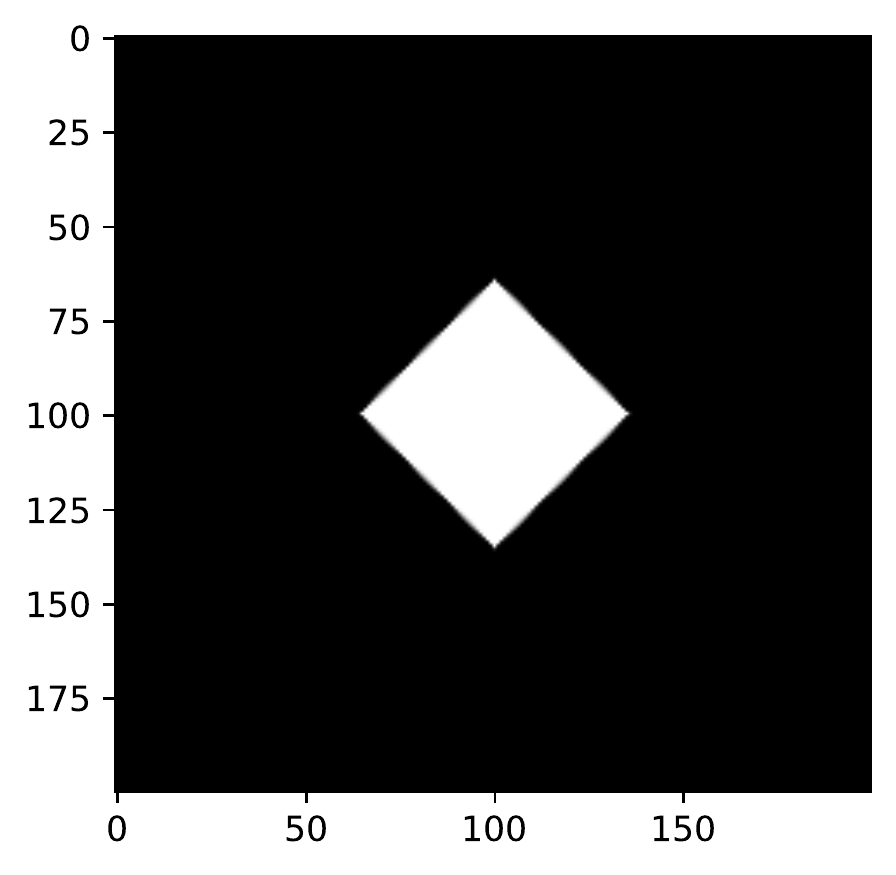}  
\end{subfigure}
\hfill
\begin{subfigure}{.11\textwidth}
  \centering
  \includegraphics[width=\linewidth]{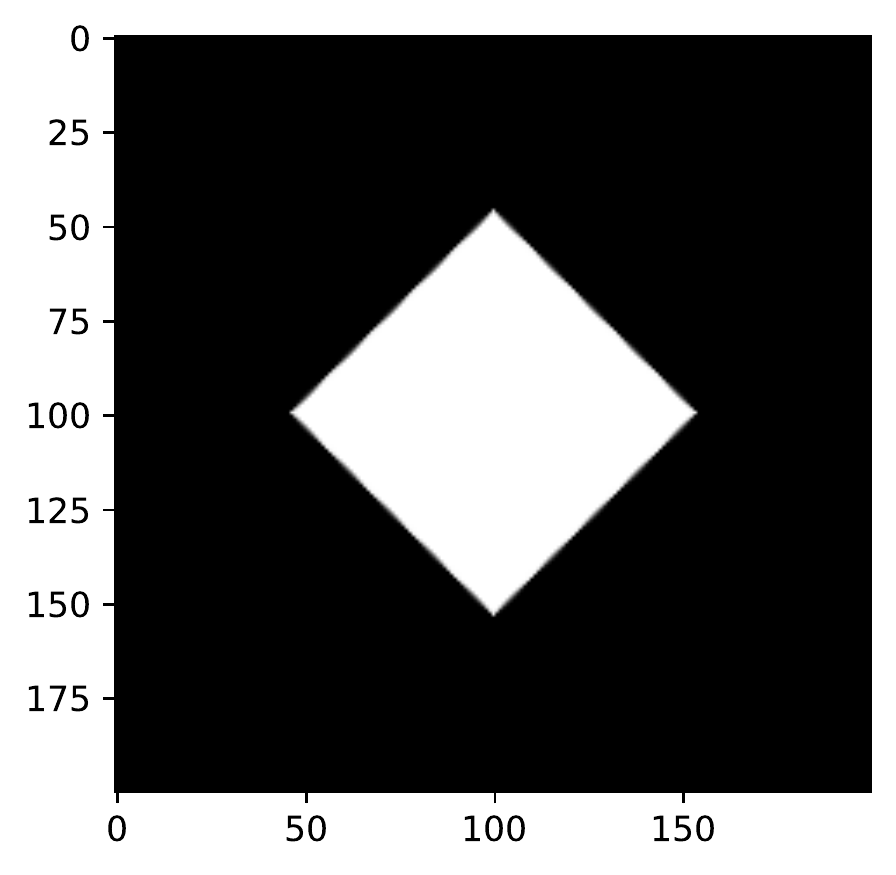}  
\end{subfigure}
\hfill
\begin{subfigure}{.11\textwidth}
  \centering
  \includegraphics[width=\linewidth]{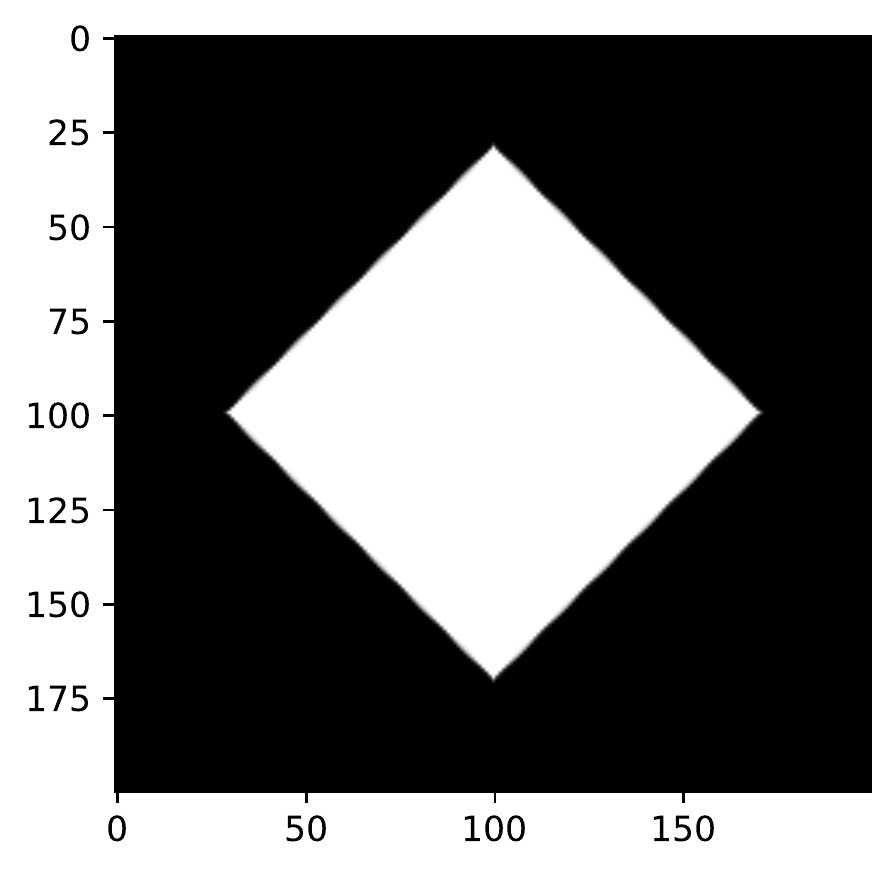}  
\end{subfigure}
\hfill
\begin{subfigure}{.11\textwidth}
  \centering
  \includegraphics[width=\linewidth]{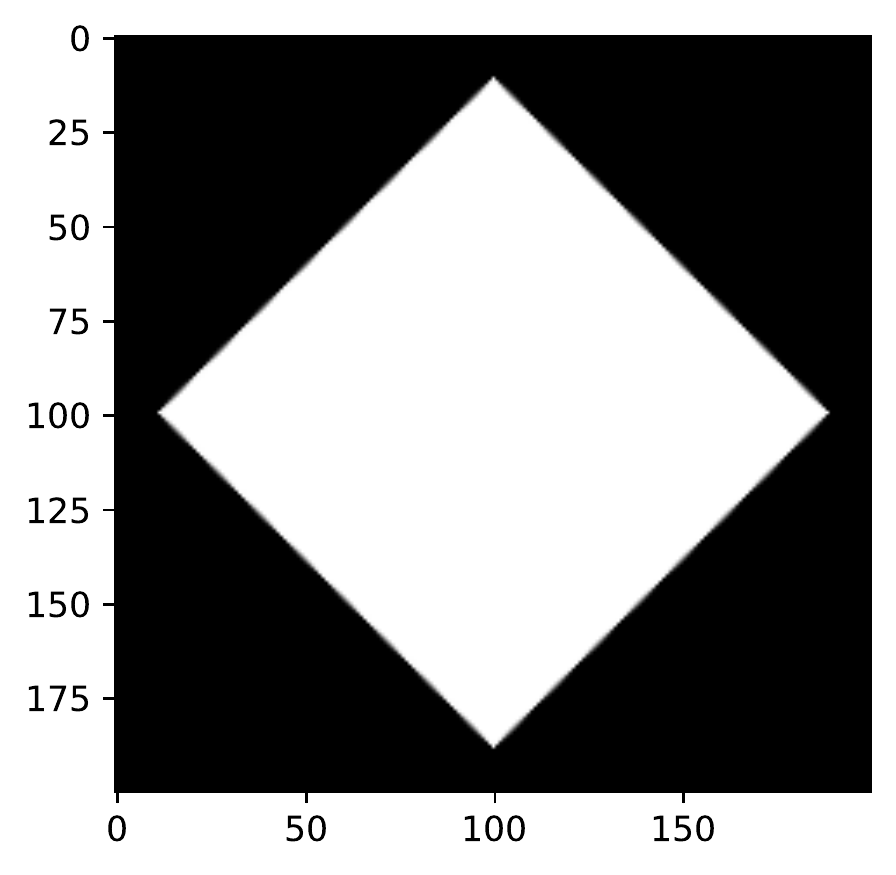}  
\end{subfigure}
\begin{subfigure}{.11\textwidth}
  \centering
  \includegraphics[width=\linewidth]{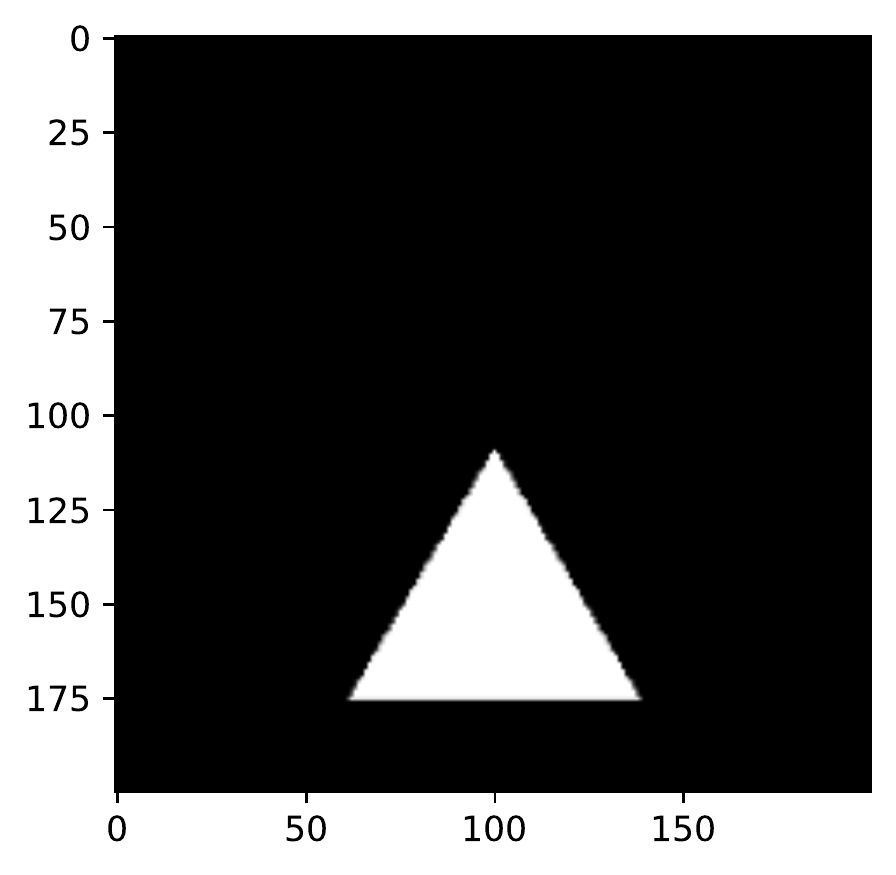}  
\end{subfigure}
\hfill
\begin{subfigure}{.11\textwidth}
  \centering
  \includegraphics[width=\linewidth]{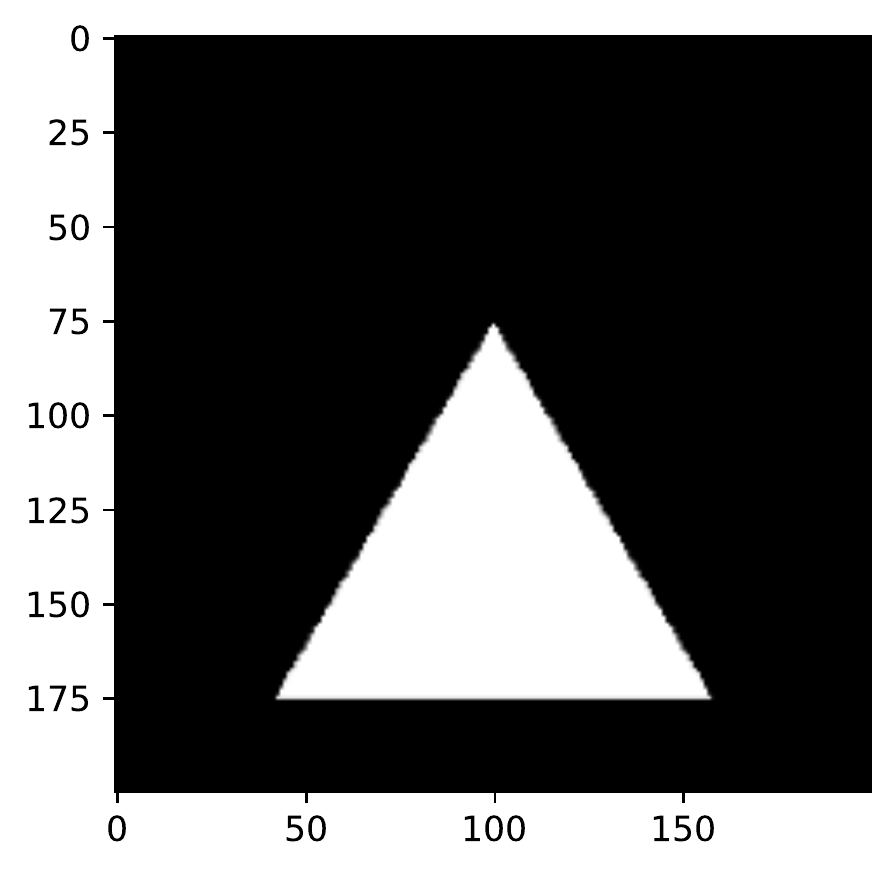}  
\end{subfigure}
\hfill
\begin{subfigure}{.11\textwidth}
  \centering
  \includegraphics[width=\linewidth]{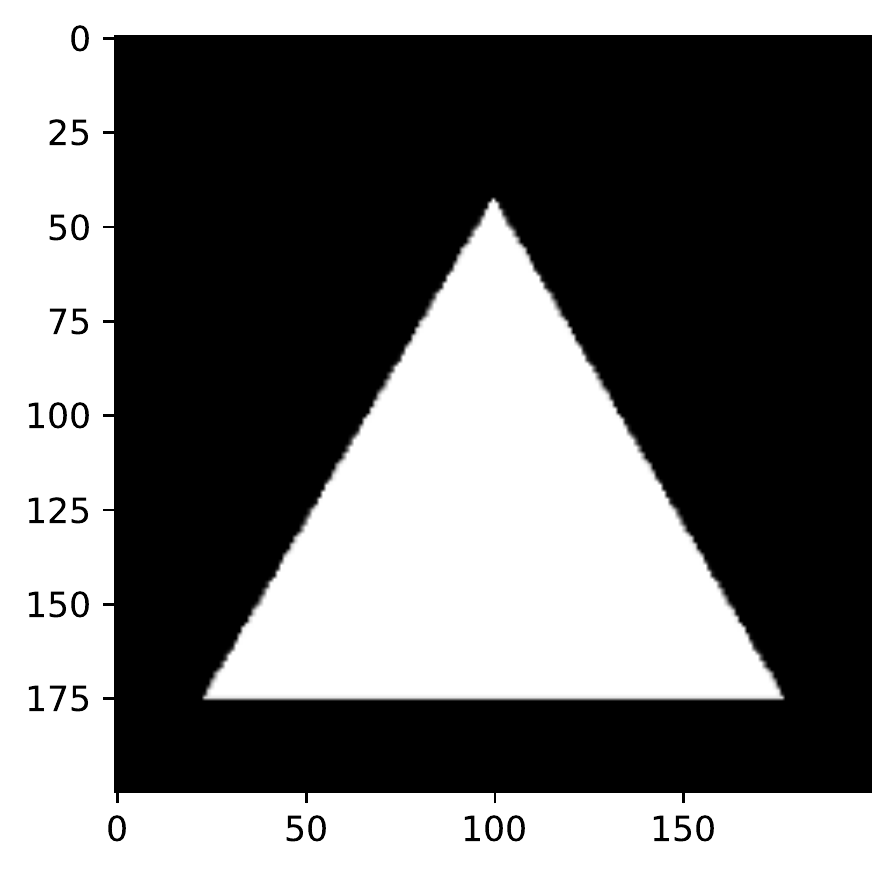}  
\end{subfigure}
\hfill
\begin{subfigure}{.11\textwidth}
  \centering
  \includegraphics[width=\linewidth]{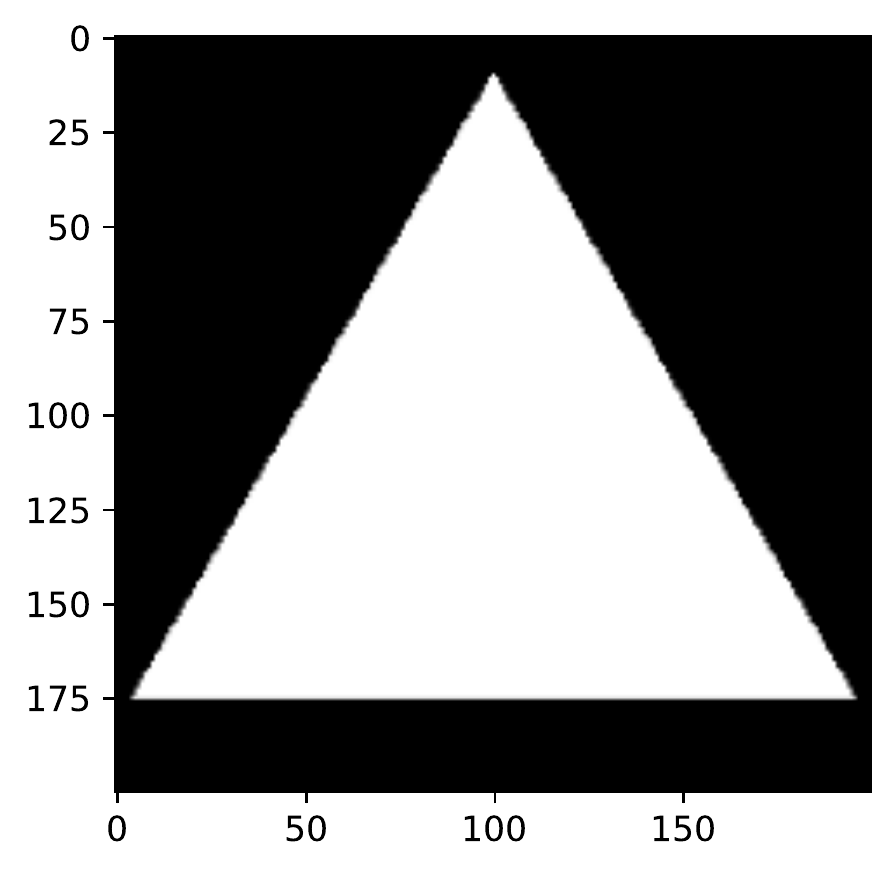}  
\end{subfigure}
\begin{subfigure}{.11\textwidth}
  \centering
  \includegraphics[width=\linewidth]{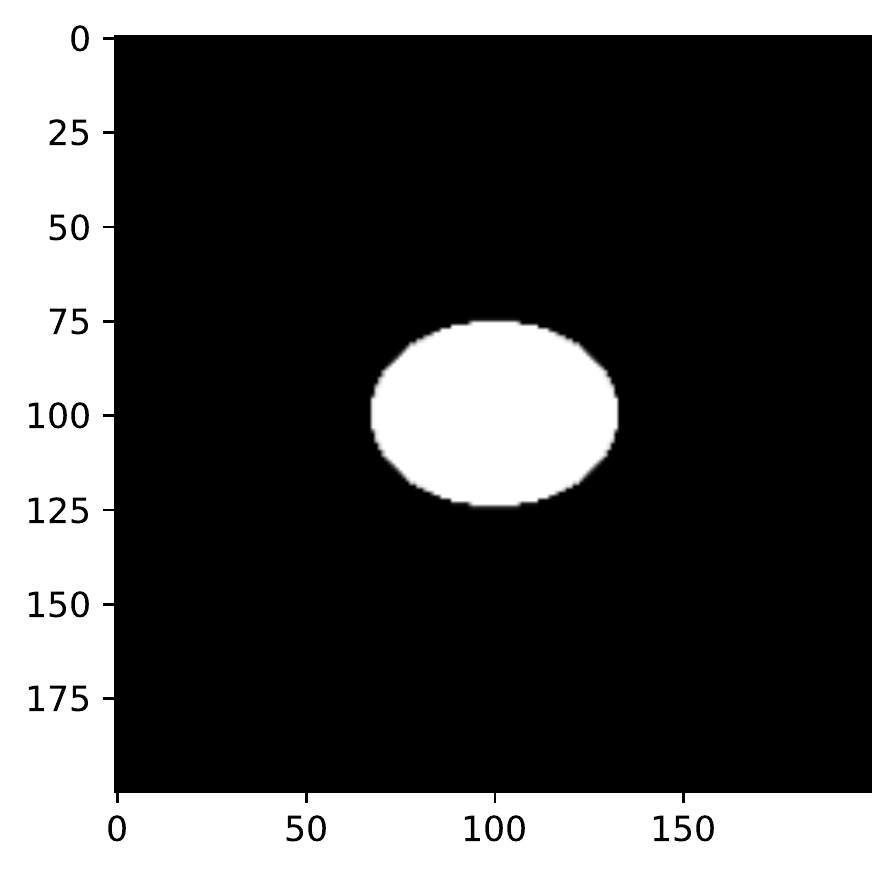} 
  \caption{$50\times 50$}
\end{subfigure}
\hfill
\begin{subfigure}{.11\textwidth}
  \centering
  \includegraphics[width=\linewidth]{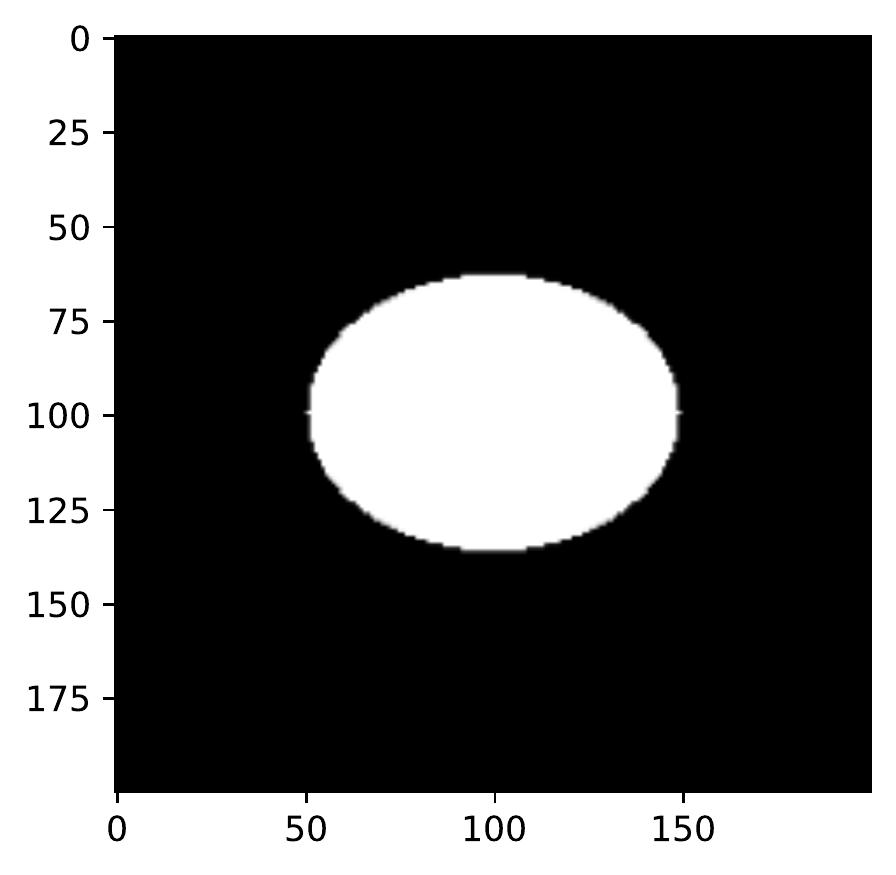} 
  \caption{$75\times 75$}
\end{subfigure}
\hfill
\begin{subfigure}{.11\textwidth}
  \centering
  \includegraphics[width=\linewidth]{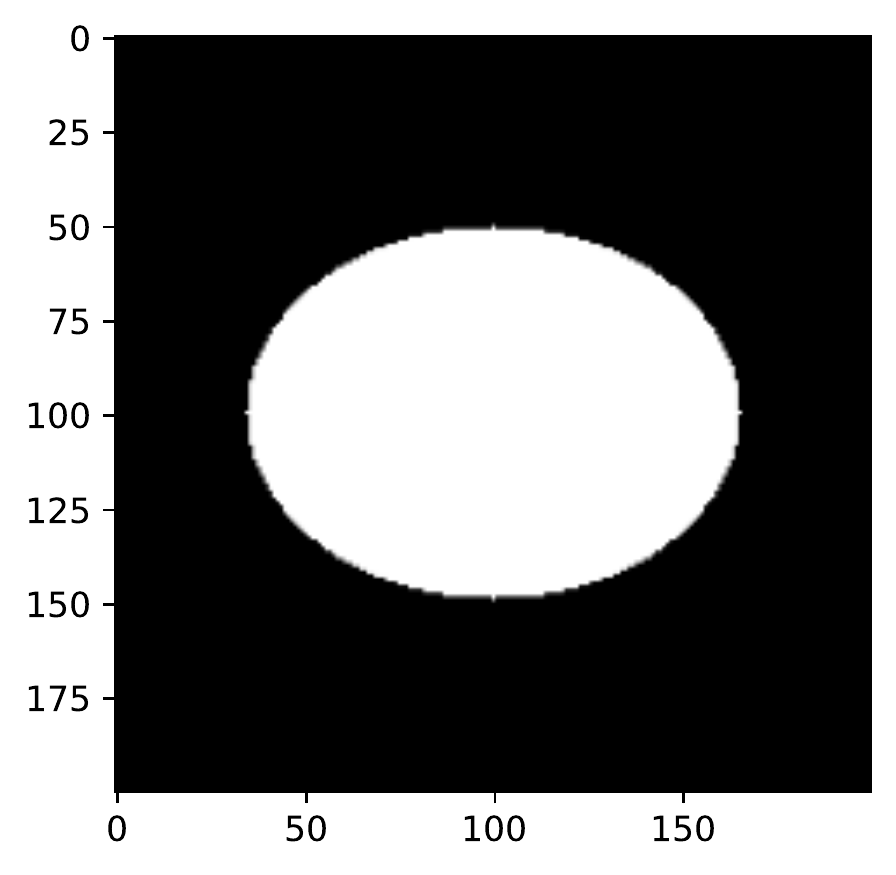}  
  \caption{$100\times 100$}
\end{subfigure}
\hfill
\begin{subfigure}{.11\textwidth}
  \centering
  \includegraphics[width=\linewidth]{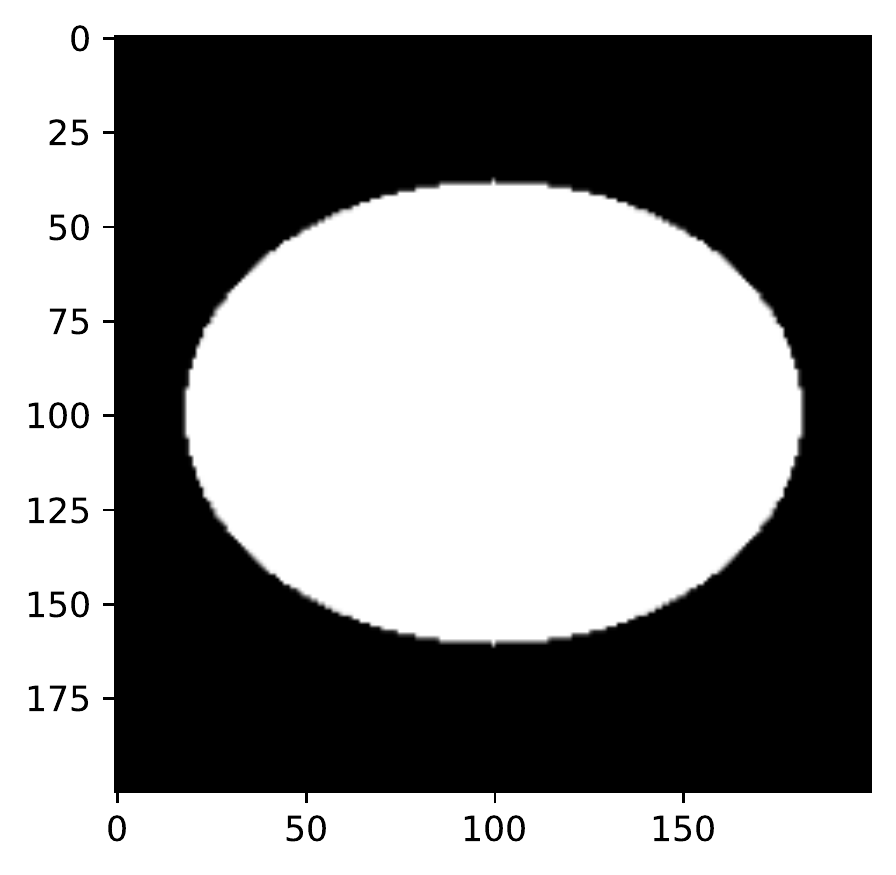}  
  \caption{$125\times 125$}
\end{subfigure}
\caption{Different shapes used for evaluating SAC. From top to bottom: square, circle, rectangle, diamond, triangle, and ellipse.  Shapes in each column have approximately the same $n\times n$ pixels, where $n\in \{50, 75, 100, 125\}$.}
\label{fig:shape}
\end{figure}

In the main paper, we demonstrate generalization to unseen patch shapes that were not considered in training the patch segmenter and in shape completion, obtaining surprisingly good robust performance. The shapes used for evaluating SAC are shown in~\cref{fig:shape}.  
\subsection{Evaluate SAC with SSD}
We use Faster R-CNN~\cite{10.5555/2969239.2969250} as our base object detector in the main paper. However, SAC is compatible with any object detector as it is a pre-processing defense. In this section, we show the performance of SAC using SSD~\cite{liu2016ssd} as the base object detector on the COCO dataset. The pre-trained SSD model is provided in~\texttt{torchvision}~\cite{10.1145/1873951.1874254}. We do not re-train the patch segmenter for SSD as the self adversarial training on the patch segmenter is object-detector agnostic. Results shown in~\Cref{tab:ssd} demonstrate that SAC can also provide strong robustness for SSD across different attack methods and patch sizes.

\begin{table}[h]
\centering
\setlength{\tabcolsep}{2pt}
\caption{mAP (\%) under different attack methods using SSD as the base object detector. The mAP on clean images is 44.5\% for the undefended model and 44.4\% for the SAC defended model.} 
\label{tab:ssd}
\begin{tabular}{ccccc} 
\toprule
 Attack & Method & 75$\times$75                       & 100$\times$100   & 125$\times$125                       \\ 
\midrule
\multirow{2}{*}{PGD~\cite{madry2017towards}}  & Undefended & 18.3$\pm$0.4  & 11.4$\pm$0.2  & 7.0$\pm$0.1\\
&SAC (\textbf{Ours})        & 39.1$\pm$0.3 & 38.8$\pm$0.2 & 34.2$\pm$0.1 \\ 

\multirow{2}{*}{DPatch~\cite{liu2018dpatch}}  & Undefended & 21.5$\pm$0.8  & 16.9$\pm$0.2  & 12.5$\pm$0.6\\
&SAC (\textbf{Ours})        & 39.9$\pm$0.2 & 39.1$\pm$0.1 & 35.4$\pm$0.3 \\ 
\multirow{2}{*}{MIM~\cite{dong2018boosting}}  & Undefended & 17.6$\pm$0.5  & 10.4$\pm$0.2  & 6.0$\pm$0.2\\
&SAC (\textbf{Ours})        & 37.9$\pm$0.2 & 38.5$\pm$0.1 & 35.0$\pm$0.3 \\
\bottomrule
\end{tabular}
\end{table}
\subsection{Shape Completion Details}
\subsubsection{Dynamic Programming for Shape Completion} \label{sec:dynprog}

Recall that our shape-completed mask is defined as:

\begin{equation} 
    \label{eq:completion-simple}
    \hat{M}_{SC\,(i,j)}  := \begin{cases}
    1&\text{    if}\,\,\,\,\exists\,i',\,j':\,\,\, M^{s,(i',j')}_{(i,j)} = 1 \text{ and }   \\
    & \frac{d_H(\hat{M}_{PS}, M^{s,(i',j')})}{s^2} \leq \gamma  \\
    0&\text{otherwise.}
    \end{cases}
\end{equation}

Here, we give a dynamic-programming based $O(H\times W)$ time algorithm for computing this mask.

We first need to define the following $O(H\times W)$ time subroutine: for an $H\times W$ binary matrix $M$, let $\text{Cuml.}(M)$ be defined as follows:
\begin{equation}
    \text{Cuml.}(M)_{(i,j)} := \sum_{i' = 1} ^i \sum_{j' = 1} ^j M_{(i',j')} 
\end{equation}
The entire matrix  $\text{Cuml.}(M)$ can be computed in  $O(H\times W)$ as follows. We first define $\text{Cuml.}^{x}(M)$ as:
\begin{equation}
    \text{Cuml.}^{x}(M)_{(i,j)} := \sum_{i' = 1} ^i M_{(i',j)} 
\end{equation}
Note that $\text{Cuml.}^{x}(M)_{(1,j)} = M_{(1,j)}$ and that, for $i> 1$, 
\begin{equation}
    \text{Cuml.}^{x}(M)_{(i,j)} := M_{(i,j)} + \text{Cuml.}^{x}(M)_{(i-1,j)}
\end{equation}
We can then construct $ \text{Cuml.}^{x}(M)$ row-by-row along the index $i$, with each cell taking constant time to fill: therefore $\text{Cuml.}^{x}(M)$ is constructed in $O(H\times W)$ time. $\text{Cuml.}(M)$ can then be constructed through two applications of this algorithm as:

\begin{equation}
    \text{Cuml.}(M) = (\text{Cuml.}^{x}((\text{Cuml.}^{x}(M))^T))^T
\end{equation}

We now apply this algorithm to $\hat{M}_{PS}:$
\begin{equation}
    \text{Cuml}\hat{M}_{PS}:= \text{Cuml.}(\hat{M}_{PS}).
\end{equation}
Note that, for each $i,j$:
\begin{equation}
\begin{split}
      &d_H(\hat{M}_{PS}, M^{s,(i,j)}) \\
      & = \sum_{\substack{i' \in [i,i+s)\\ j' \in [j,j+s)}} (1-\hat{M}_{PS, (i',j')}) + \sum_{\substack{i' \not\in [i,i+s) \vee \\ j' \not \in [j,j+s)}} \hat{M}_{PS, (i',j')}  \\
      & = s^2  - \sum_{\substack{i' \in [i,i+s)\\ j' \in [j,j+s)}} \hat{M}_{PS, (i',j')} + \sum_{\substack{i' \not\in [i,i+s) \vee \\ j' \not \in [j,j+s)}} \hat{M}_{PS, (i',j')}  \\
      & = s^2 + \sum_{(i',j')} \hat{M}_{PS, (i',j')}   - 2\sum_{\substack{i' \in [i,i+s)\\ j' \in [j,j+s)}} \hat{M}_{PS, (i',j')}  \\
      & = s^2 + \text{Cuml}\hat{M}_{PS(H,W)}   - 2\Bigg(\sum_{\substack{i' \in [1,i+s)\\ j' \in [1,j+s)}} \hat{M}_{PS, (i',j')} - \\
      & \sum_{\substack{i' \in [1,i)\\ j' \in [1,j+s)}} \hat{M}_{PS, (i',j')}- \sum_{\substack{i' \in [1,i+s)\\ j' \in [1,j)}} \hat{M}_{PS, (i',j')} + \sum_{\substack{i' \in [1,i)\\ j' \in [1,j)}} \hat{M}_{PS, (i',j')}\Bigg)\\ 
      &= s^2 + \text{Cuml}\hat{M}_{PS,(H,W)}   - 2\bigg(
      \text{Cuml}\hat{M}_{PS,(i+s-1,j+s-1)}\\&  - \text{Cuml}\hat{M}_{PS,(i-1,j+s-1)} -
      \text{Cuml}\hat{M}_{PS,(i+s-1,j-1)} \\& +
      \text{Cuml}\hat{M}_{PS,(i-1,j-1)}\bigg)
\end{split}
\end{equation}
(We are disregarding edge cases where $i+s > H$ or $j+s > W$: these can be easily reasoned about.) Using a pre-computed $ \text{Cuml}\hat{M}_{PS}$, we can then compute each of these Hamming distances in constant time. We can then, in $O(H\times W)$ time, compute the matrix $\hat{M}_\gamma$:
\begin{equation}
    \hat{M}_{\gamma, (i,j)} := \mathbbm{ 1}_{\frac{d_H(\hat{M}_{PS}, M^{s,(i,j)})}{s^2} \leq \gamma} \label{eq:threshold_1}
\end{equation}
where $\mathbbm{ 1}$ denotes an indicator function.
We also pre-compute the cumulative sums of this matrix:
\begin{equation}
    \text{Cuml}\hat{M}_{\gamma} := \text{Cuml.}( \hat{M}_{\gamma})
\end{equation}

Now, recall the condition of \cref{eq:completion-simple}:
\begin{equation}
    \begin{split}
        &\exists\,i',\,j':\,\,\, M^{s,(i',j')}_{(i,j)} = 1 \text{ and }  \frac{d_H(\hat{M}_{PS}, M^{s,(i',j')})}{s^2} \leq \gamma\\
        \iff  & \exists\,i',\,j':\,\,\, M^{s,(i',j')}_{(i,j)} = 1 \text{ and }   \hat{M}_{\gamma, (i',j')} = 1\\
        \iff  &\sum_{\substack{i' \in (i-s,i]\\ j' \in (j-s,j]}}  \hat{M}_{\gamma, (i',j')}  \geq 1\\
        \iff  &\Bigg( \sum_{\substack{i' \in [1,i]\\ j' \in [1,j]}}  \hat{M}_{\gamma, (i',j')} - \sum_{\substack{i' \in [1,i-s]\\ j' \in [1,j]}}  \hat{M}_{\gamma, (i',j')}\\& - \sum_{\substack{i' \in [1,i]\\ j' \in [1,j-s]}}  \hat{M}_{\gamma, (i',j')}+ \sum_{\substack{i' \in [1,i-s]\\ j' \in [1,j-s]}}  \hat{M}_{\gamma, (i',j')} \Bigg) \geq 1\\
        \iff \big( &\text{Cuml}\hat{M}_{\gamma,(i,j)} - \text{Cuml}\hat{M}_{\gamma,(i-s,j)} \\&-
      \text{Cuml}\hat{M}_{\gamma,(i,j-s)} +
      \text{Cuml}\hat{M}_{\gamma,(i-s,j-s)}  \big) \geq 1
    \end{split} \label{eq:equiv_condition}
\end{equation}
Again, this can be computed in constant time for each index. Let $\hat{C}_{\gamma,(i,j)}:=\text{Cuml}\hat{M}_{\gamma,(i,j)} - \text{Cuml}\hat{M}_{\gamma,(i-s,j)} -
      \text{Cuml}\hat{M}_{\gamma,(i,j-s)} +
      \text{Cuml}\hat{M}_{\gamma,(i-s,j-s)}$, then~\cref{eq:completion-simple} becomes simply:
\begin{equation}
     \hat{M}_{SC\,(i,j)}  := \mathbbm{1}_{ \hat{C}_{\gamma,(i,j)}\geq 1} \label{eq:threshold_2}
\end{equation}
This gives us an overall runtime of $O(H\times W)$ as desired. Note that in our PyTorch implementation, we are able to use tensor operations such that no explicit iteration over indices is necessary at any point in the algorithm.
\subsubsection{Adjusting $\gamma$} \label{sec:gamma_search}
\begin{figure}
    \centering
    \includegraphics[width=3.25in]{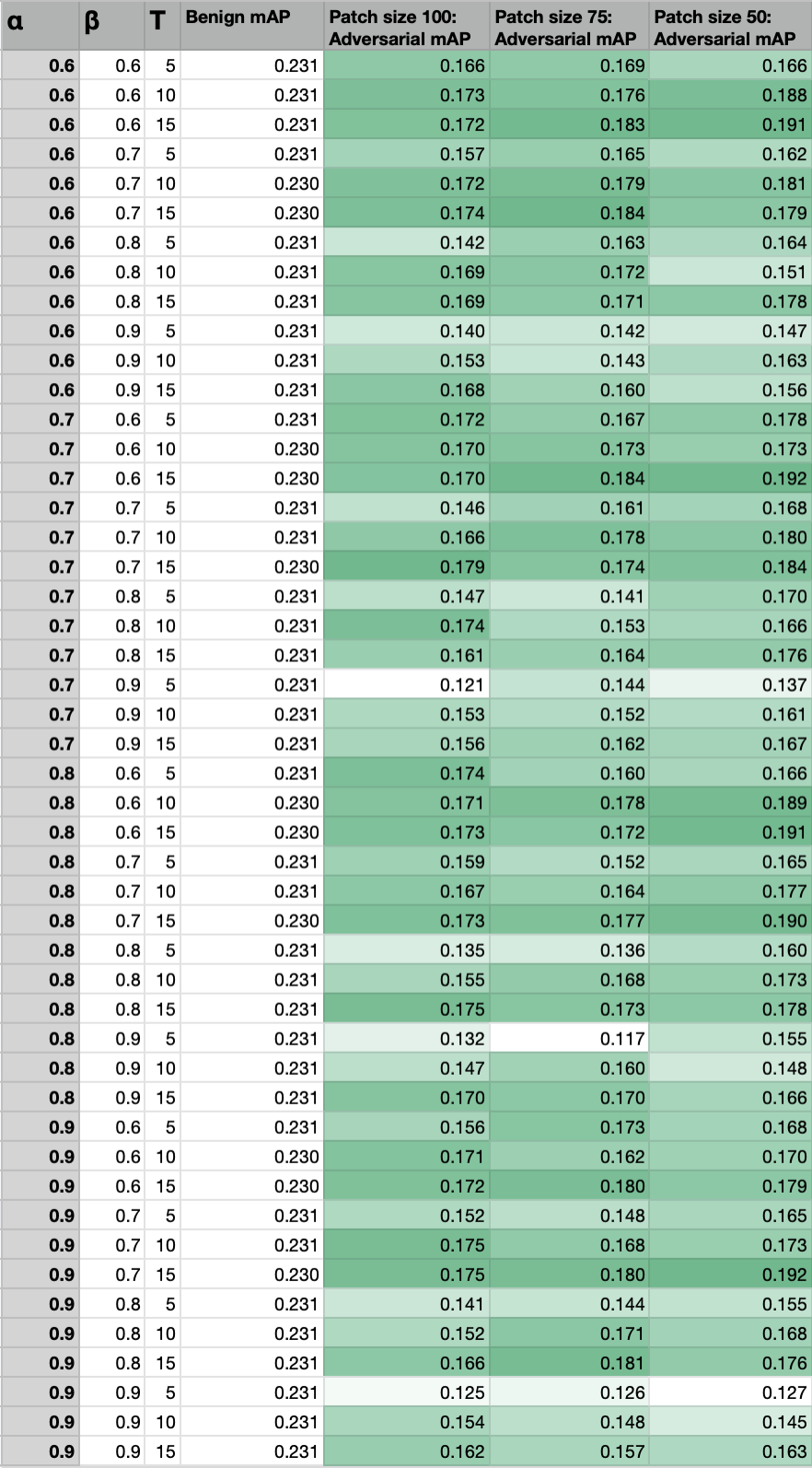}
    \caption{Validation set performance on xView under adaptive attack, as a function of defense hyperparameters $\alpha,\beta,T$ used for searching over $\gamma$. Within each column, more green shading indicates higher mAP. }
    \label{fig:Grid_search_a_b_t}
\end{figure}

In practice, the method described above can be highly sensitive to the hyperparameter $\gamma$. If $\gamma$ is set too low, then no candidate mask $M^{s,(i',j')}$ will be sufficiently close to  $\hat{M}_{PS}$, so the detector will return nothing. However, if $\gamma$ is set too high, then the shape completion will be too conservative, masking a large area of possible candidate patches. (Note that $\gamma \geq 1$ is not usable, because it would cover an image entirely with a mask even when  $\hat{M}_{PS} = \textbf{0}$.) To deal with this issue, we initially use low values of $\gamma$, and then gradually increase $\gamma$ if no mask is initially returned -- stopping when either some mask is returned or a maximum value is reached, at which point we assume that there is no ground-truth adversarial patch. 
Specifically, for iteration $t = 1,..., T$, we set 
\[\gamma_t := 1 - \alpha\beta^{(t-1)},\]
where $T \in \mathbb{N},$ and  $\alpha, \beta < 1 $.
We then return the first nonzero  $ \hat{M}_{SC}(S,\gamma_t)$, or an empty mask if this does not occur. We set $\alpha = 0.9$, $\beta = 0.7$, $T = 15$. The values of $\alpha$, $\beta$ and $T$ are tuned using grid search on a validation set with 200 images from the xView dataset (See Figure \ref{fig:Grid_search_a_b_t}).
\subsubsection{Adaptive attacks on Shape Completion}
To attack the patch segmenter, we use a straight-through estimator (STE)~\cite{bengio2013estimating} at the thresholding step: $\hat{M}_{PS} = PS_\theta(x) > 0.5.$ To attack the shape completion algorithm, we have tried the following attacks:
\paragraph{BPDA Attack}
Note that the algorithm described in Section \ref{sec:dynprog} involves two non-differentiable thresholding steps (\cref{eq:threshold_1} and \cref{eq:threshold_2}). In order to implement an adaptive attack, at these steps, we use BPDA, using a STE for the gradient at each thresholding step. 
When aggregating masks which assume patches of different sizes (Equation 9 in the main text) we also use a straight-through estimator on a thresholded sum of masks. This is the strongest adaptive attack for SAC that we found and we use this attack in the main paper. 
\paragraph{$\gamma$-Search STE Attack} 

There is an additional non-differentiable step in the defense, however: the search over values of $\gamma$ described in Section \ref{sec:gamma_search}. In order to deal with this, we attempted to use BPDA as well, using the following recursive formulation:
\begin{equation}
\begin{split}
       &\hat{M}_{SC}(S)_{\alpha,\beta,0} := \mathbf{0}\\
       & \hat{M}_{SC}(S)_{\alpha,\beta,T} :=   \hat{M}_{SC}(S,1-\alpha) \\
        &+ \mathbbm{1}_{\frac{\Sigma  \hat{M}_{SC}(S,1-\alpha)}{C}  < 1 } \hat{M}_{SC}(S)_{\alpha*\beta,\beta,T-1} \,\,\,\,\text{(for $T\geq1$)}
\end{split}
\label{eq:gamma_STE}
\end{equation} 
Where $C$ is the area of the smallest considered patch size in $S$ (i.e., the minimum nonzero shape completion output ).

\begin{figure}
    \centering
    \includegraphics[width=3.25in]{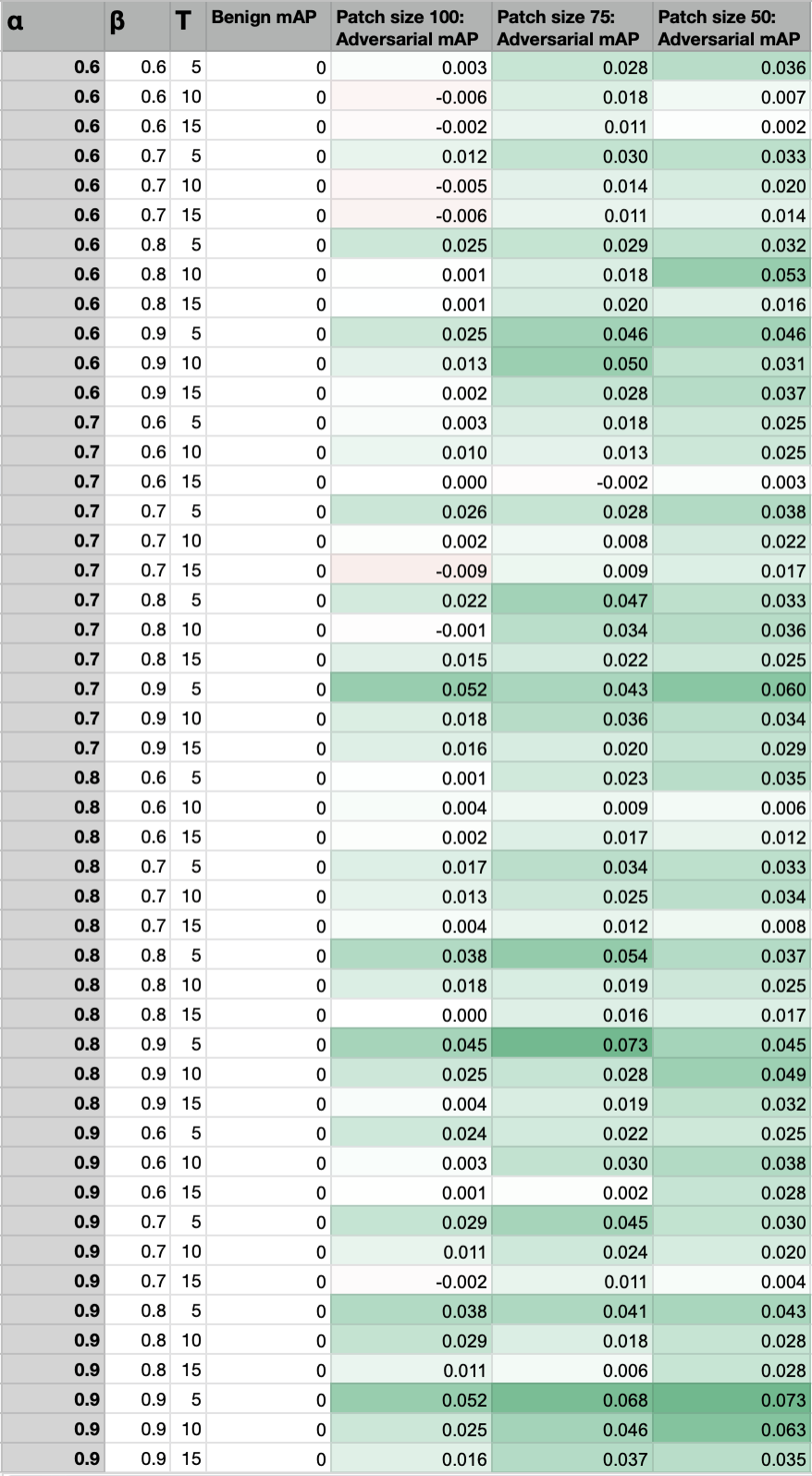}
    \caption{Difference in mAP under BPDA attack using STE gradients for the search over $\gamma$ (as in \cref{eq:gamma_STE}) versus simply treating the search as non-differentiable,  on 200-image xView validation set. Positive numbers (green) indicate that the non-differentiable treatment yielded a more successful attack, while negative numbers (red) indicate that the STE treatment was more successful. We see that in most hyperparameter settings, the STE treatment of the search over $\gamma$ made the attack less successful, and in no setting did it make the attack substantially more successful.  }
    \label{fig:bpda_v2}
\end{figure}

We can then use a STE for the indicator function. However, this technique turns out to yield worse performance in practice than simply treating the search over $\gamma$ as non-differentiable   (See \cref{fig:bpda_v2}). Therefore, in our main results, we treat this search over $\gamma$ as non-differentiable, rather than using an STE.
\paragraph{Log-Sum-Exp Transfer Attack} 
We were also initially concerned that the simple straight-through estimation approach for the  algorithm described in Section \ref{sec:dynprog} might fail, specifically at the point of \cref{eq:threshold_2}, where the threshold takes the form (see \cref{eq:equiv_condition}):
\begin{equation}
      \sum_{\substack{i' \in (i-s,i]\\ j' \in (j-s,j]}}  \hat{M}_{\gamma, (i',j')}  \geq 1
\end{equation}
where $\hat{M}_{\gamma, (i',j')}$ is a 0/1 indicator of whether a patch should be added to the final output mask with upper-left corner $(i',j')$. We were concerned that a straight-through estimator would propagate gradients to \textit{the sum directly}, affecting \textit{every potential} patch which could cover a location $(i,j)$, rather than concentrating the gradient only on those patches that \textit{actually}  contribute to the pixel $(i,j)$ being masked.

To mitigate this, we first considered the equivalent threshold condition:
\begin{equation}
      \max_{\substack{i' \in (i-s,i]\\ j' \in (j-s,j]}}  \hat{M}_{\gamma, (i',j')}  \geq 1
\end{equation}
While logically equivalent, the gradient propagated by the STE to the LHS  would now only propagate on to the values $\hat{M}_{\gamma, (i',j')} $ which are equal to 1. However, unfortunately, this formulation is not compatible with the dynamic programming algorithm described in Section \ref{sec:dynprog}: due to computational limitations, we do not want to compute the maximum over every pair $(i',j')$, \textit{for each pair} $(i,j)$. 

To solve this problem, we instead used the following proxy function when generating attack gradients (including during the forward pass):
\begin{equation} 
     \log \Bigg(\sum_{\substack{i' \in (i-s,i]\\ j' \in (j-s,j]}}  e^{C \cdot \hat{M}_{\gamma, (i',j')}} \Bigg)/C \geq 1 \label{eq:approx_condition}
\end{equation}
where $C$ is a large constant (we use $C=10\log(100)$). This is the ``LogSumExp'' softmax function:  note that the LHS is approximately 1 if any $\hat{M}_{\gamma, (i',j')}$ is one and approximately zero otherwise. Also note that the derivative of the LHS with respect to each $\hat{M}_{\gamma, (i',j')}$ is similarly approximately 1 if $\hat{M}_{\gamma, (i',j')}$ is one and zero otherwise. Crucially, we can compute this in the above DP framework, simply by replacing $\hat{M}_{\gamma, (i',j')}$ with its exponent (and taking the log before thresholding).

\begin{figure}
    \centering
    \includegraphics[width=3.25in]{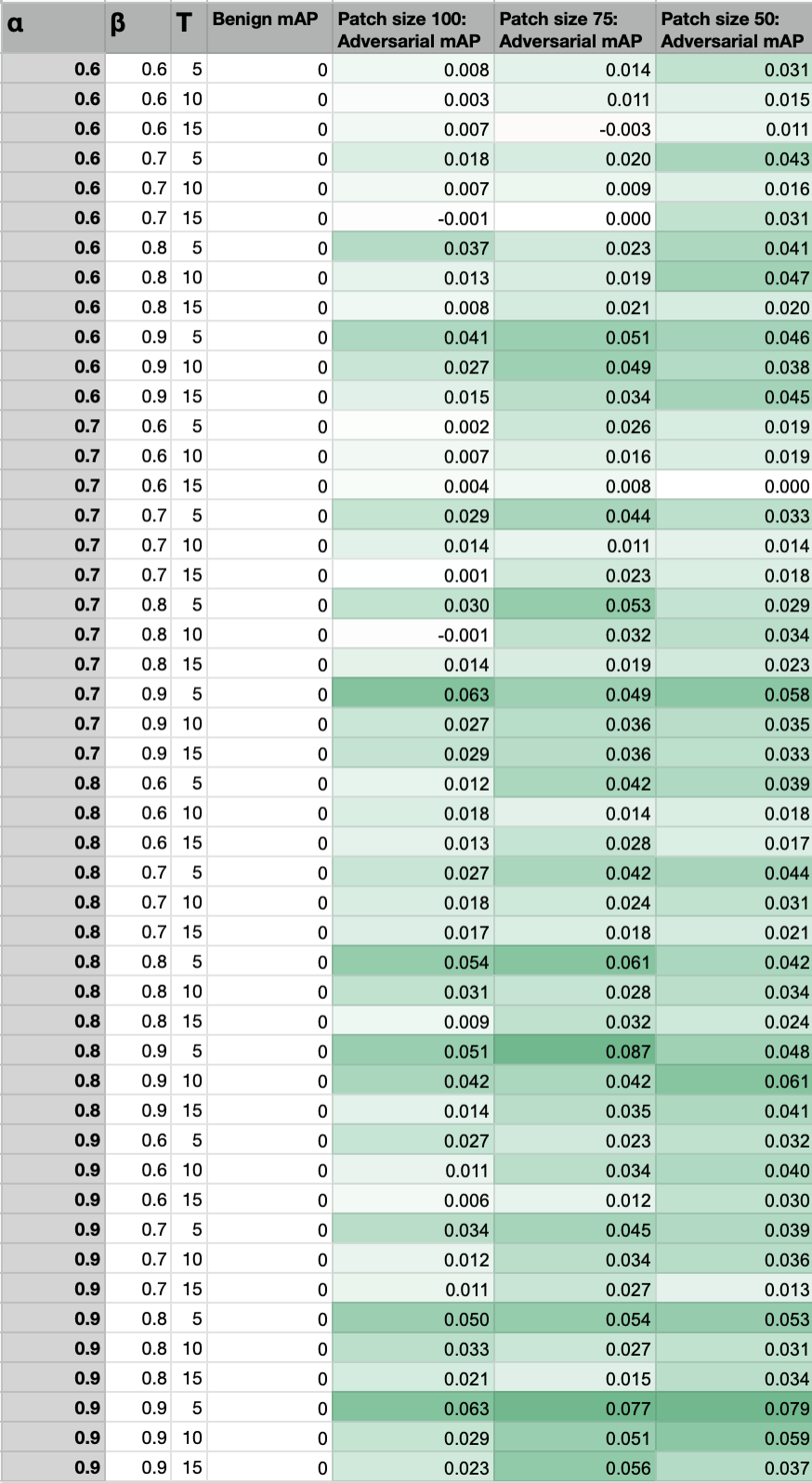}
    \caption{Difference in mAP using Log-Sum-Exp approximation for \cref{eq:threshold_2} as described in \cref{eq:approx_condition} versus the naive BPDA attack we ultimately used, on 200-image xView validation set. Positive numbers (green) indicate that the naive BPDA attack yielded a more successful attack, while negative numbers (red) indicate that the Log-Sum-Exp treatment was more successful. We see that in most hyperparameter settings, the Log-Sum-Exp technique made the attack less successful, and in no setting did it make the attack substantially more successful.  }
    \label{fig:old_attack}
\end{figure}

However, in practice,  the naive BPDA attack outperforms this adaptive attack (\cref{fig:old_attack}). This is likely because the condition in \cref{eq:approx_condition} is an inexact approximation, so the function being attacked differs from the true objective. (In both attacks, we treat the search over $\gamma$ as nondifferentiable, as described above.)

\subsubsection{Patch Visualization}
\begin{figure}[htbp]
    \centering
    \begin{subfigure}[b]{0.23\textwidth}
        \centering
        \includegraphics[width=\textwidth]{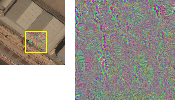}
        \caption{{\small Patch for SAC without SC.}}    
    \end{subfigure}
    \hfill
    \begin{subfigure}[b]{0.23\textwidth}  
        \centering 
        \includegraphics[width=\textwidth]{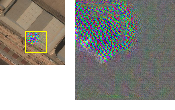}
        \caption{{\small Patch for SAC with SC.}}
    \end{subfigure}
    \caption{$100\times100$ adversarial patches generated by adaptive attacks on xView dataset. Patches for SAC without shape completion (SC) have widespread noises in the square bounded area, while patches for SAC with SC have structured noises.} 
    \label{fig:patch}
\end{figure}

We find that adaptive attacks on models with SC
would force the attacker to generate patches that have more structured noises trying to fool SC (see \cref{fig:patch}).

\subsubsection{Visualization of Shape Completion Outputs}
\begin{figure*}
\begin{subfigure}{.24\textwidth}
  \centering
  \includegraphics[width=\linewidth]{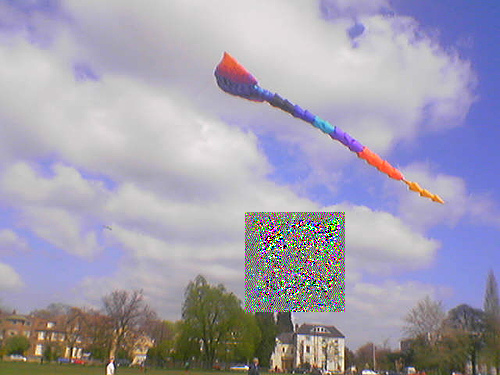}
\end{subfigure}
\hfill
\begin{subfigure}{.24\textwidth}
  \centering
  \includegraphics[width=\linewidth]{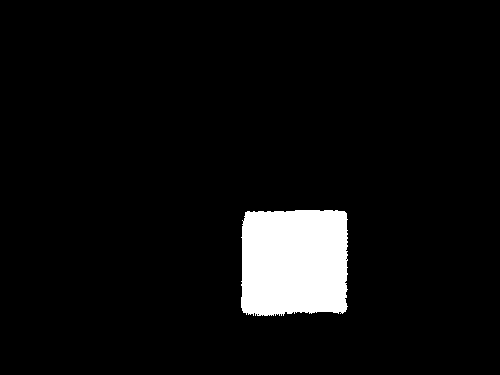}
\end{subfigure}
\hfill
\begin{subfigure}{.24\textwidth}
  \centering
  \includegraphics[width=\linewidth]{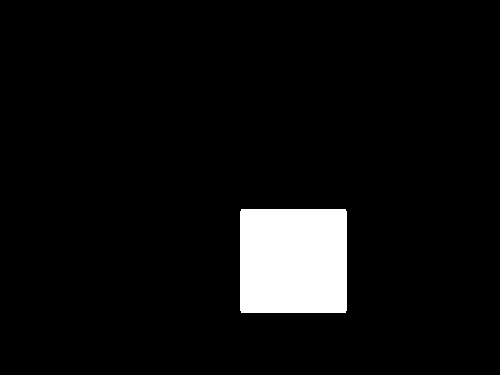} 
\end{subfigure}
\hfill
\begin{subfigure}{.24\textwidth}
  \centering
  \includegraphics[width=\linewidth]{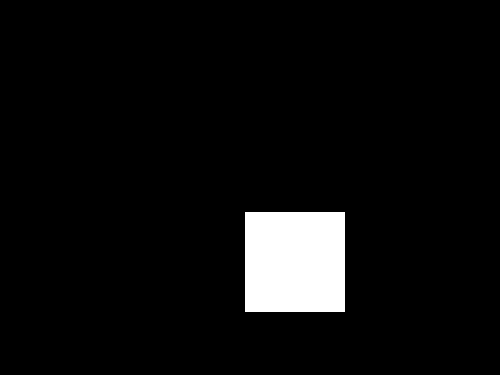}  
\end{subfigure}
\begin{subfigure}{.24\textwidth}
  \centering
  \includegraphics[width=\linewidth]{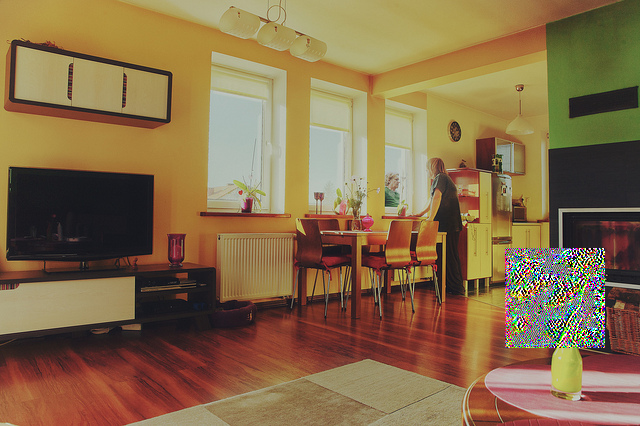}  
\end{subfigure}
\hfill
\begin{subfigure}{.24\textwidth}
  \centering
  \includegraphics[width=\linewidth]{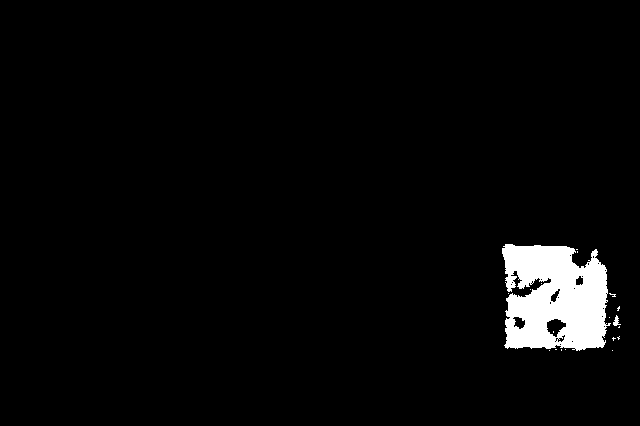}  
\end{subfigure}
\hfill
\begin{subfigure}{.24\textwidth}
  \centering
  \includegraphics[width=\linewidth]{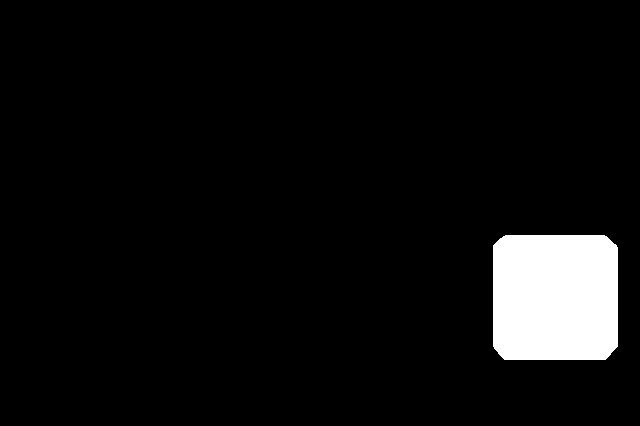}  
\end{subfigure}
\hfill
\begin{subfigure}{.24\textwidth}
  \centering
  \includegraphics[width=\linewidth]{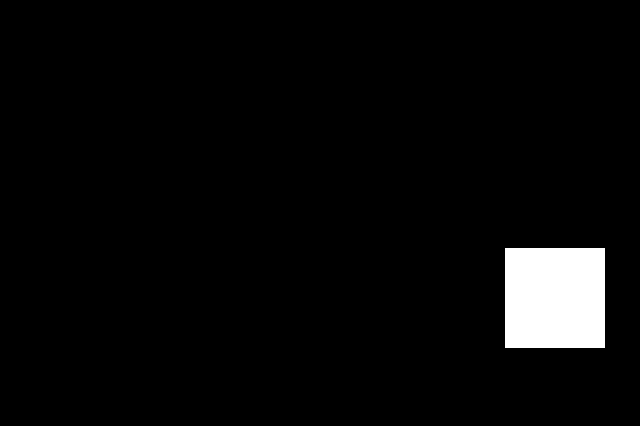}  
\end{subfigure}
\hfill
\begin{subfigure}{.24\textwidth}
  \centering
  \includegraphics[width=\linewidth]{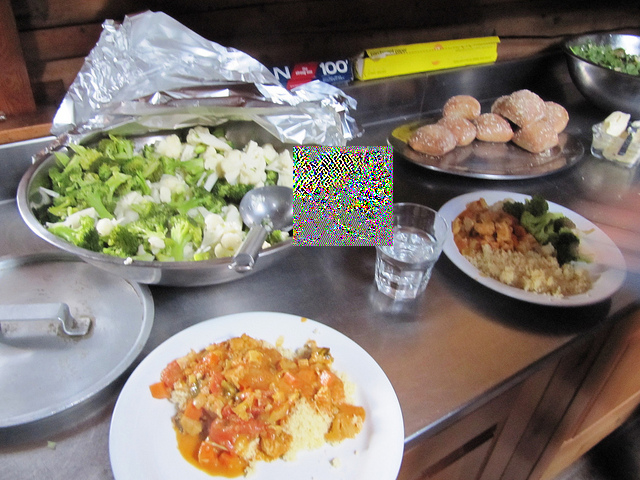}  
\end{subfigure}
\hfill
\begin{subfigure}{.24\textwidth}
  \centering
  \includegraphics[width=\linewidth]{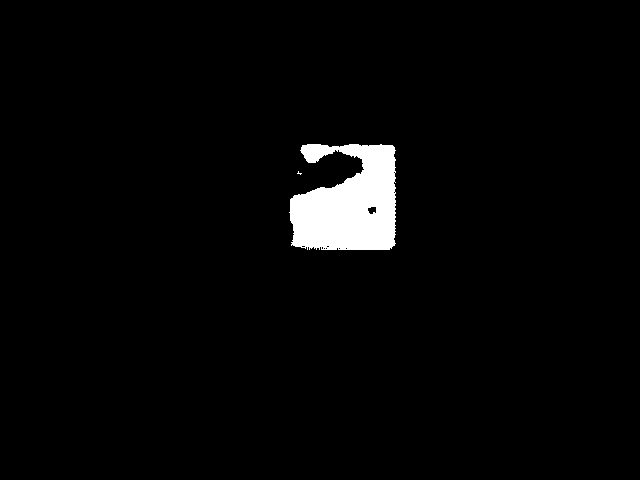}  
\end{subfigure}
\hfill
\begin{subfigure}{.24\textwidth}
  \centering
  \includegraphics[width=\linewidth]{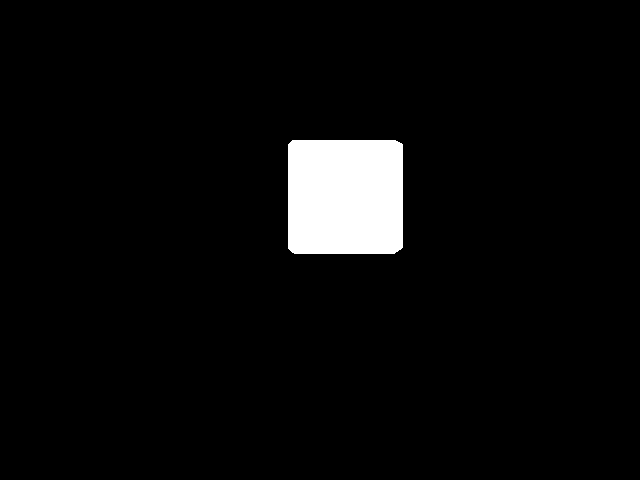}  
\end{subfigure}
\hfill
\begin{subfigure}{.24\textwidth}
  \centering
  \includegraphics[width=\linewidth]{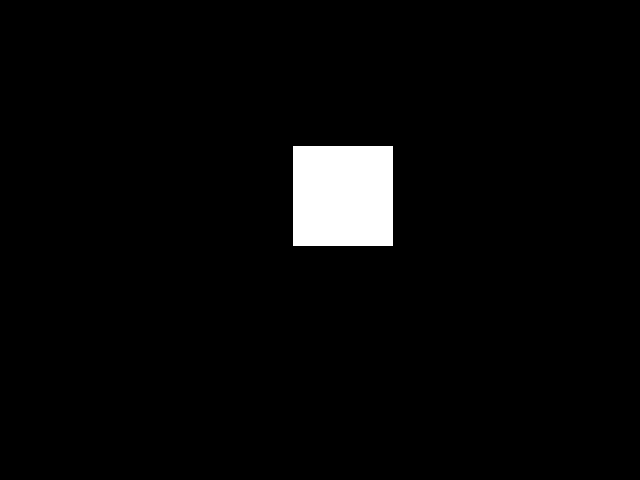}  
\end{subfigure}
\hfill
\begin{subfigure}{.24\textwidth}
  \centering
  \includegraphics[width=\linewidth]{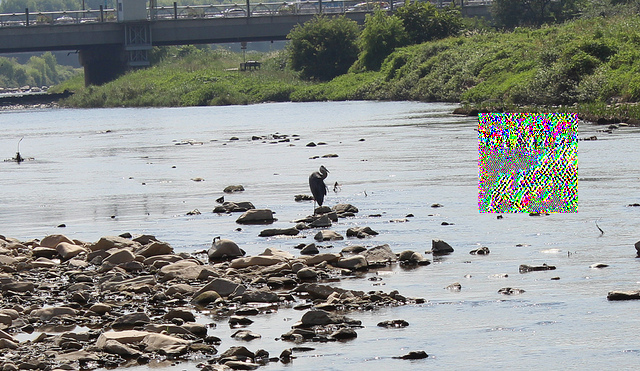}
  \caption{Adversarial images.}
\end{subfigure}
\hfill
\begin{subfigure}{.24\textwidth}
  \centering
  \includegraphics[width=\linewidth]{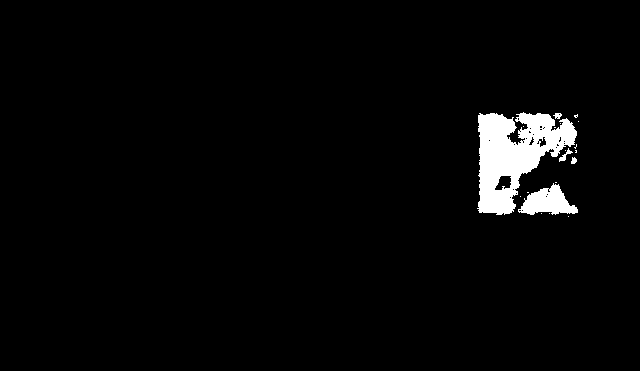}
  \caption{Outputs of patch segmentation $\hat{M}_{PS}$.}
\end{subfigure}
\hfill
\begin{subfigure}{.24\textwidth}
  \centering
  \includegraphics[width=\linewidth]{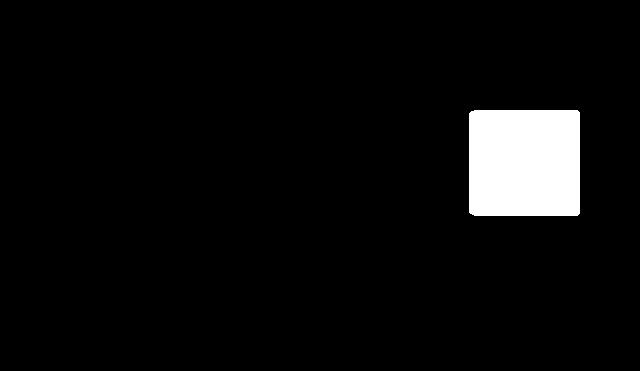} 
  \caption{Outputs of shape completion $\hat{M}_{SC}$.}
\end{subfigure}
\hfill
\begin{subfigure}{.24\textwidth}
  \centering
  \includegraphics[width=\linewidth]{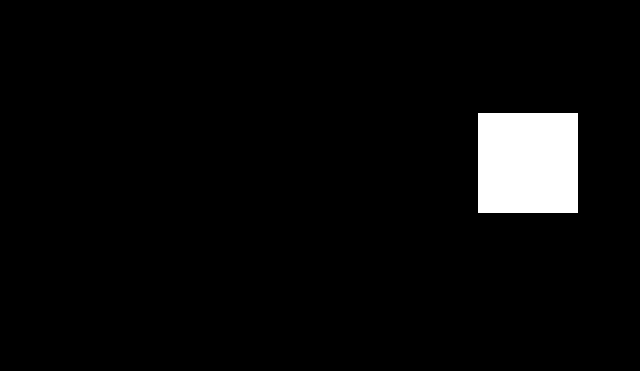}  
  \caption{Ground-truth patch masks $M$.}
\end{subfigure}

\caption{Visualization of shape completion outputs. Given the output of the patch segmenter, the proposed shape completion algorithm generates a ``completed patch mask" to cover the entire adversarial patches.}
\label{fig:sc}
\end{figure*}

We provide several examples of shape completion outputs in~\cref{fig:sc}. The outputs of the patch segmenter can be disturbed by the attacker such that some parts of the adversarial patches are not detected, especially under adaptive attacks. Given the output mask of the patch segmenter, the proposed shape completion algorithm generates a ``completed patch mask" to cover the entire adversarial patches.

\subsection{Visualization of Detection Results}
\subsubsection{SAC under Adaptive Attacks}
\begin{figure*}
\begin{subfigure}[t]{.24\textwidth}
  \centering
  \includegraphics[width=\linewidth]{fig/vis_SAC_coco/1_gt_det.png}  
\end{subfigure}
\hfill
\begin{subfigure}[t]{.24\textwidth}
  \centering
  \includegraphics[width=\linewidth]{fig/vis_SAC_coco/1_pred_clean.png}
\end{subfigure}
\hfill
\begin{subfigure}[t]{.24\textwidth}
  \centering
  \includegraphics[width=\linewidth]{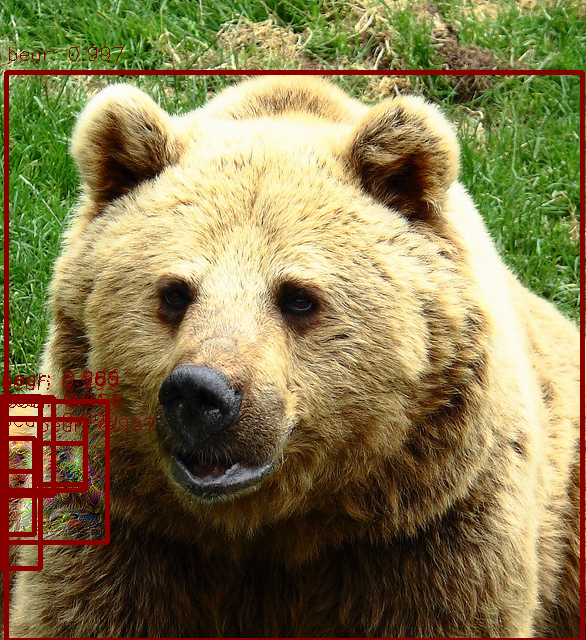}
\end{subfigure}
\hfill
\begin{subfigure}[t]{.24\textwidth}
  \centering
  \includegraphics[width=\linewidth]{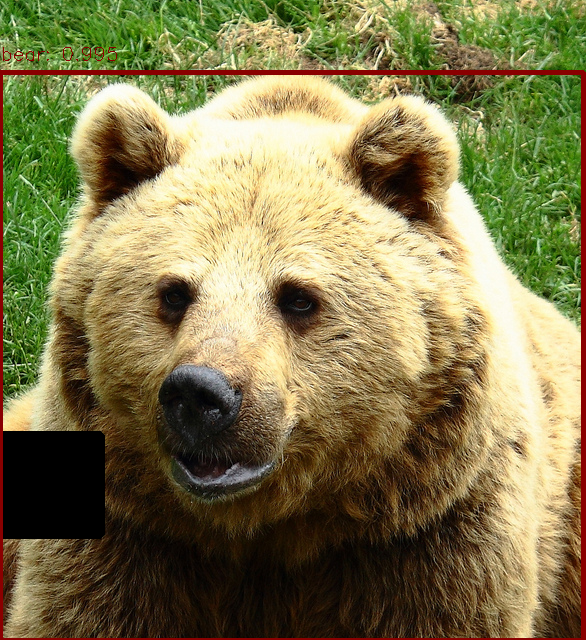}  
\end{subfigure}

\begin{subfigure}[t]{.24\textwidth}
  \centering
  \includegraphics[width=\linewidth]{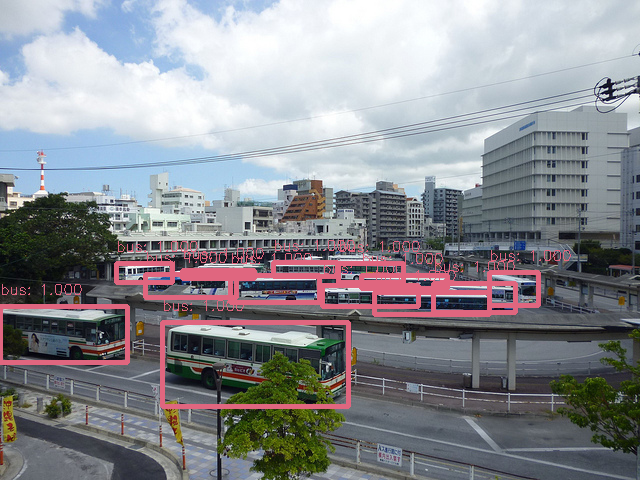}  
\end{subfigure}
\hfill
\begin{subfigure}[t]{.24\textwidth}
  \centering
  \includegraphics[width=\linewidth]{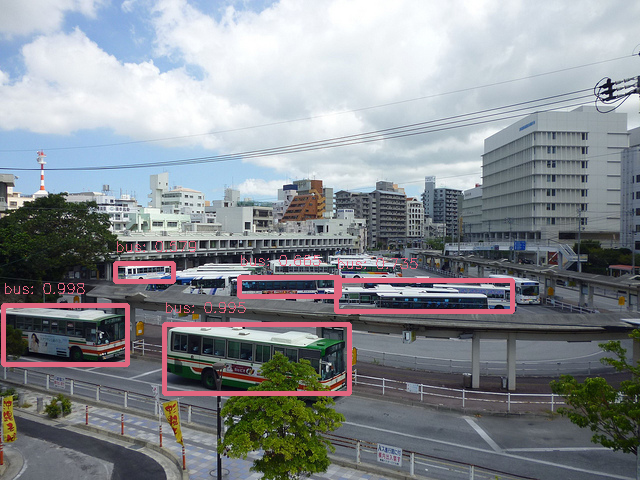}
\end{subfigure}
\hfill
\begin{subfigure}[t]{.24\textwidth}
  \centering
  \includegraphics[width=\linewidth]{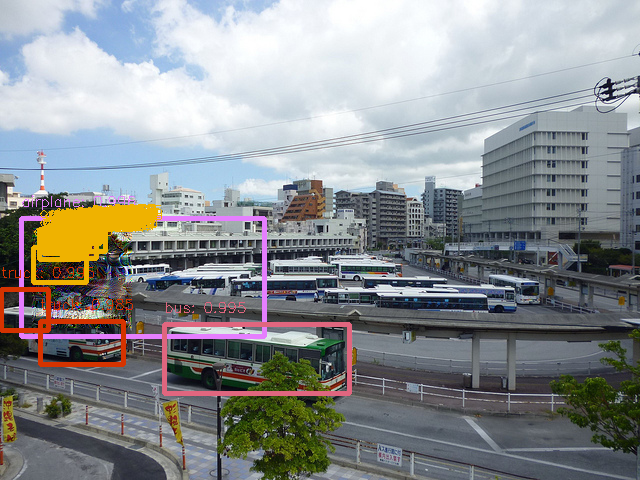}
\end{subfigure}
\hfill
\begin{subfigure}[t]{.24\textwidth}
  \centering
  \includegraphics[width=\linewidth]{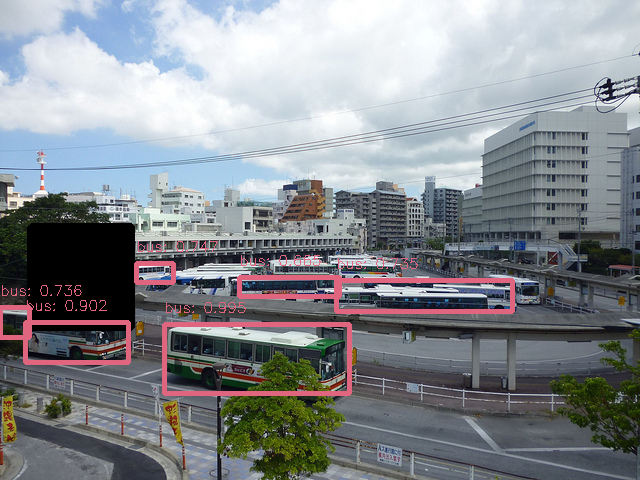}  
\end{subfigure}
\begin{subfigure}[t]{.24\textwidth}
  \centering
  \includegraphics[width=\linewidth]{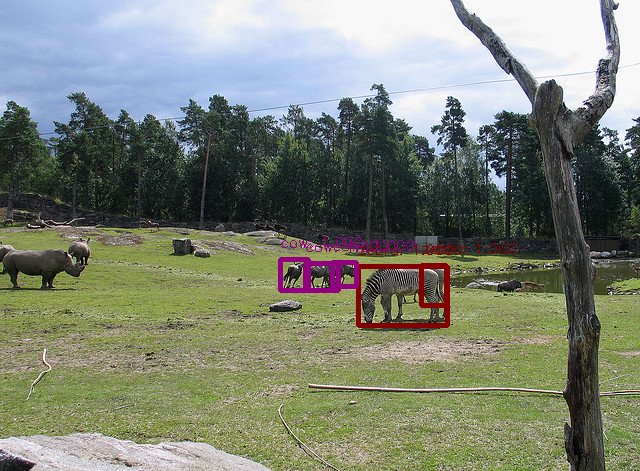}  
\end{subfigure}
\hfill
\begin{subfigure}[t]{.24\textwidth}
  \centering
  \includegraphics[width=\linewidth]{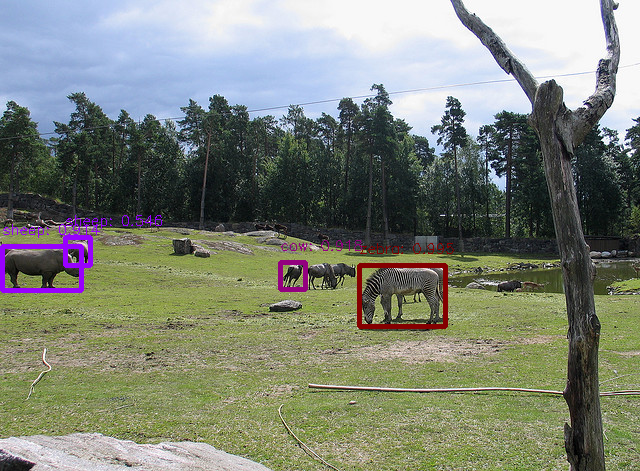}
\end{subfigure}
\hfill
\begin{subfigure}[t]{.24\textwidth}
  \centering
  \includegraphics[width=\linewidth]{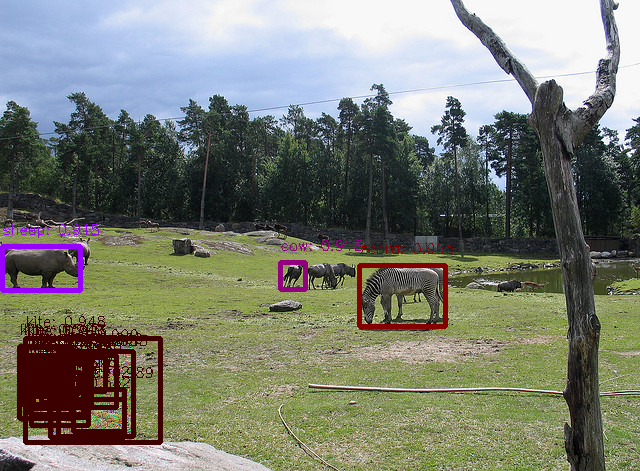}
\end{subfigure}
\hfill
\begin{subfigure}[t]{.24\textwidth}
  \centering
  \includegraphics[width=\linewidth]{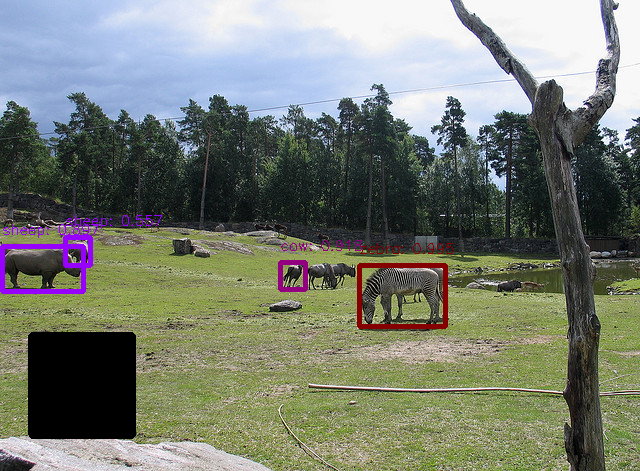}  
\end{subfigure}

\begin{subfigure}[t]{.24\textwidth}
  \centering
  \includegraphics[width=\linewidth]{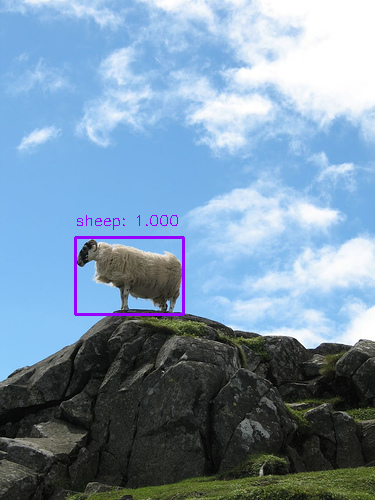}  
\end{subfigure}
\hfill
\begin{subfigure}[t]{.24\textwidth}
  \centering
  \includegraphics[width=\linewidth]{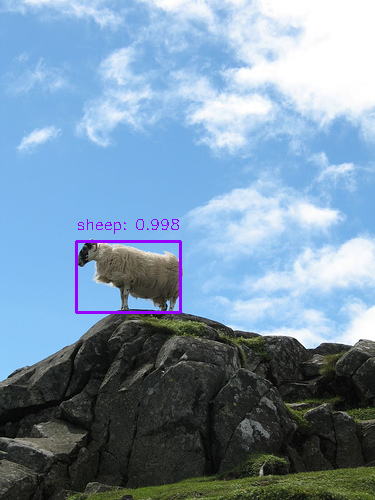}
\end{subfigure}
\hfill
\begin{subfigure}[t]{.24\textwidth}
  \centering
  \includegraphics[width=\linewidth]{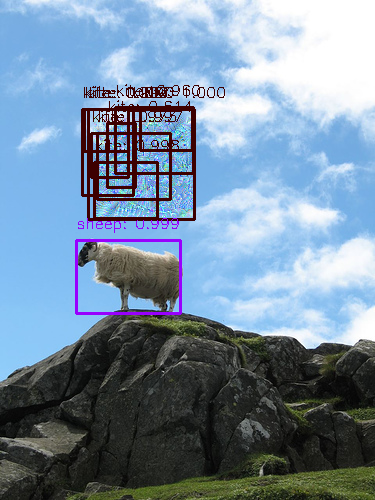}
\end{subfigure}
\hfill
\begin{subfigure}[t]{.24\textwidth}
  \centering
  \includegraphics[width=\linewidth]{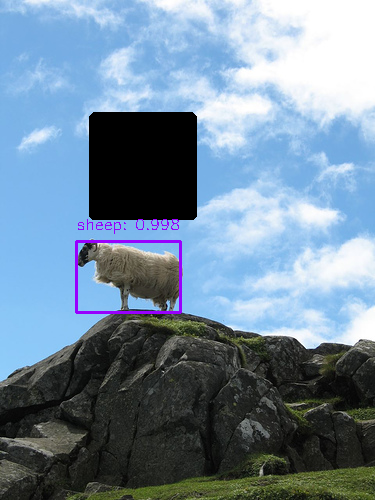}  
\end{subfigure}

\begin{subfigure}[t]{.24\textwidth}
  \centering
  \includegraphics[width=\linewidth]{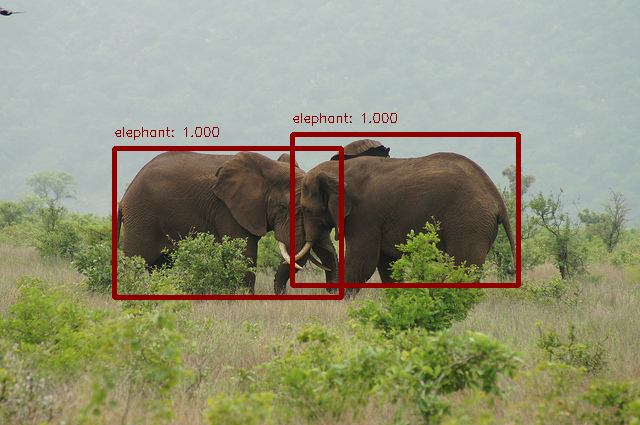}  
\end{subfigure}
\hfill
\begin{subfigure}[t]{.24\textwidth}
  \centering
  \includegraphics[width=\linewidth]{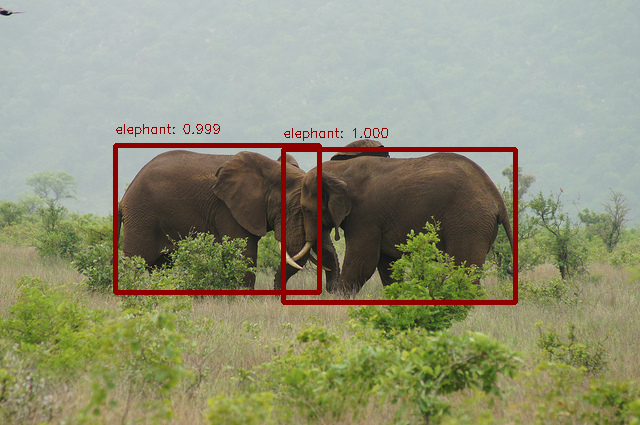}
\end{subfigure}
\hfill
\begin{subfigure}[t]{.24\textwidth}
  \centering
  \includegraphics[width=\linewidth]{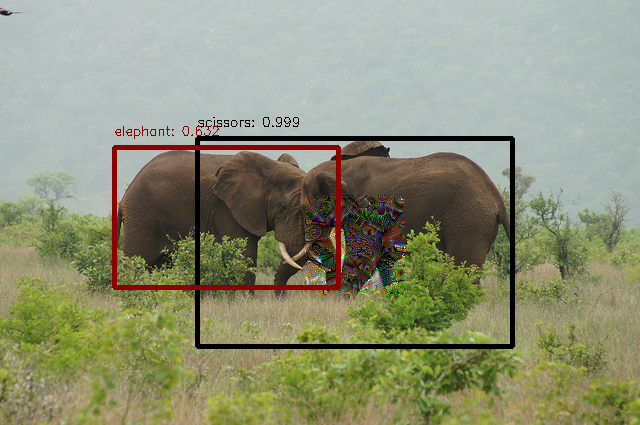}
\end{subfigure}
\hfill
\begin{subfigure}[t]{.24\textwidth}
  \centering
  \includegraphics[width=\linewidth]{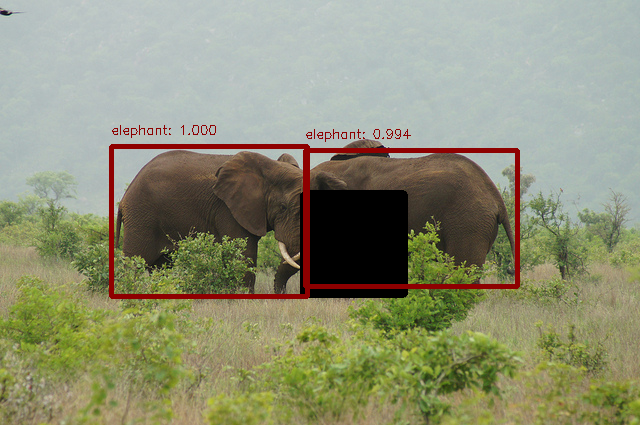}  
\end{subfigure}

\begin{subfigure}[t]{.24\textwidth}
  \centering
  \includegraphics[width=\linewidth]{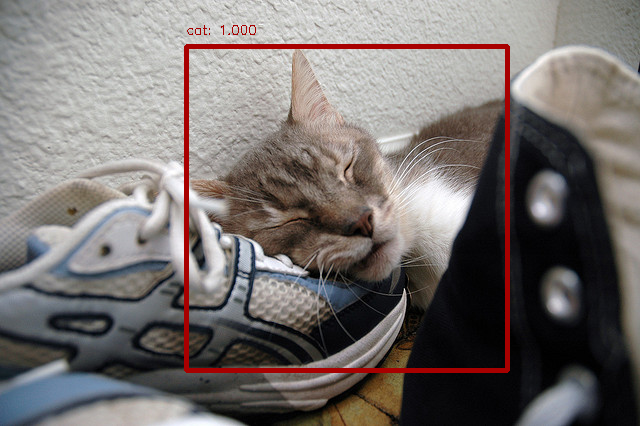}  
  \caption{Ground-truth on clean images.}
\end{subfigure}
\hfill
\begin{subfigure}[t]{.24\textwidth}
  \centering
  \includegraphics[width=\linewidth]{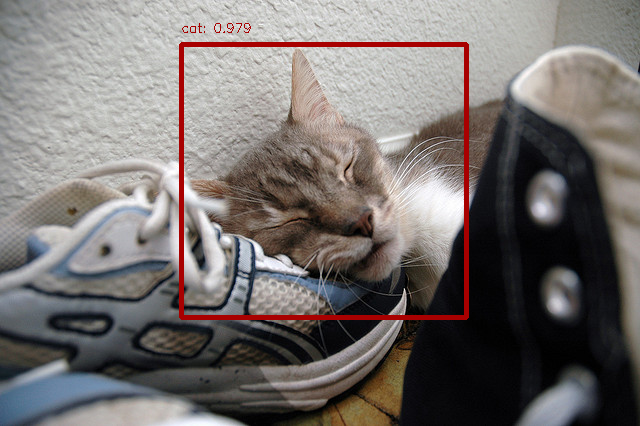}
  \caption{Predictions on clean images}
\end{subfigure}
\hfill
\begin{subfigure}[t]{.24\textwidth}
  \centering
  \includegraphics[width=\linewidth]{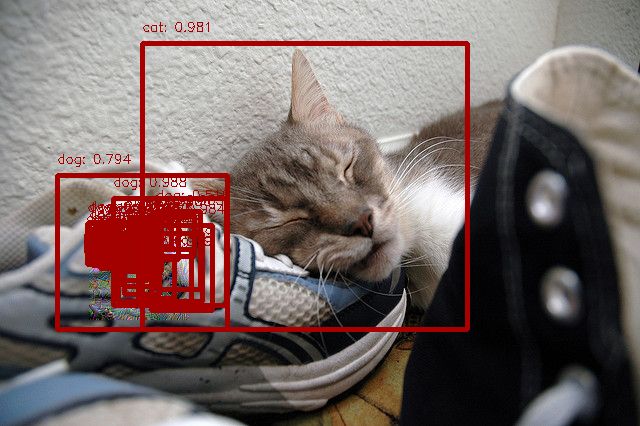}
  \caption{Predictions on adversarial images.}
\end{subfigure}
\hfill
\begin{subfigure}[t]{.24\textwidth}
  \centering
  \includegraphics[width=\linewidth]{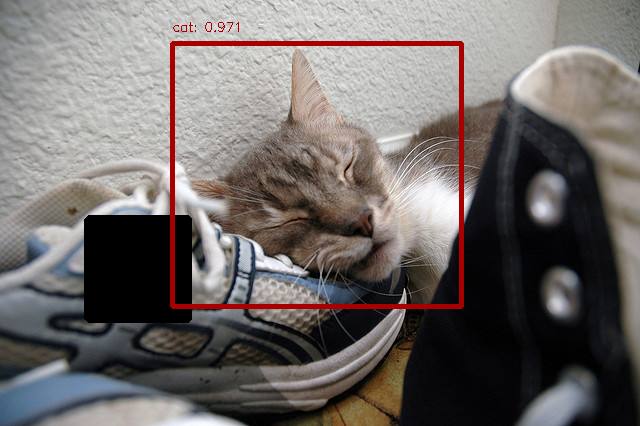}  
  \caption{Predictions on SAC masked images.}
\end{subfigure}

\caption{Examples on the COCO dataset. The adversarial patches are $100\times 100$ squares generated by PGD adaptive attacks.}
\label{fig:more_coco}
\end{figure*}

\begin{figure*}
\begin{subfigure}[t]{.24\textwidth}
  \centering
  \includegraphics[width=\linewidth]{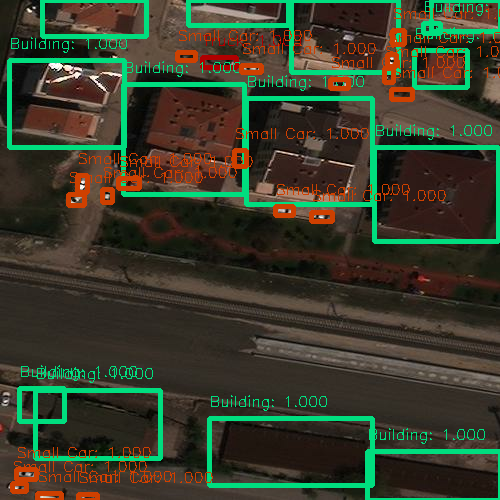}  
\end{subfigure}
\hfill
\begin{subfigure}[t]{.24\textwidth}
  \centering
  \includegraphics[width=\linewidth]{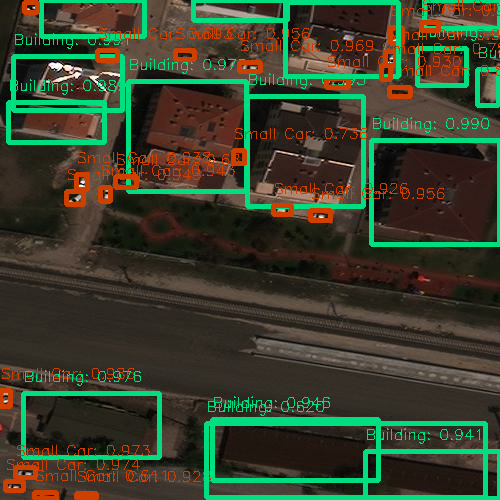}
\end{subfigure}
\hfill
\begin{subfigure}[t]{.24\textwidth}
  \centering
  \includegraphics[width=\linewidth]{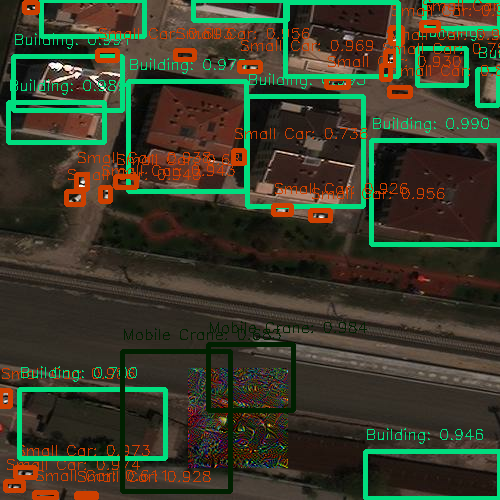}
\end{subfigure}
\hfill
\begin{subfigure}[t]{.24\textwidth}
  \centering
  \includegraphics[width=\linewidth]{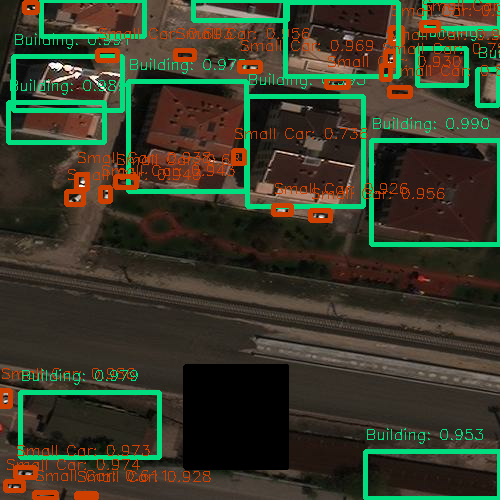}  
\end{subfigure}

\begin{subfigure}[t]{.24\textwidth}
  \centering
  \includegraphics[width=\linewidth]{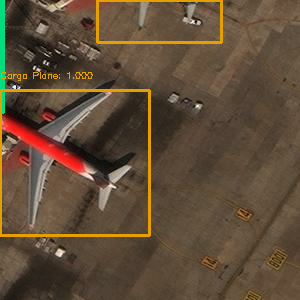}  
\end{subfigure}
\hfill
\begin{subfigure}[t]{.24\textwidth}
  \centering
  \includegraphics[width=\linewidth]{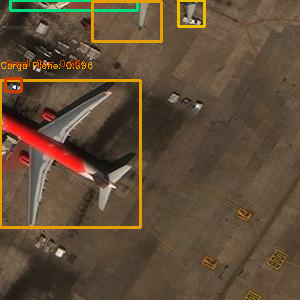}
\end{subfigure}
\hfill
\begin{subfigure}[t]{.24\textwidth}
  \centering
  \includegraphics[width=\linewidth]{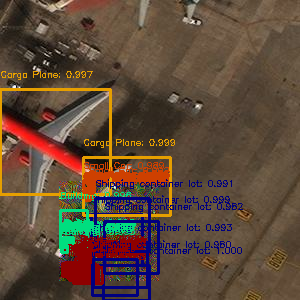}
\end{subfigure}
\hfill
\begin{subfigure}[t]{.24\textwidth}
  \centering
  \includegraphics[width=\linewidth]{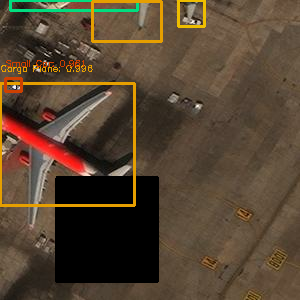}  
\end{subfigure}

\begin{subfigure}[t]{.24\textwidth}
  \centering
  \includegraphics[width=\linewidth]{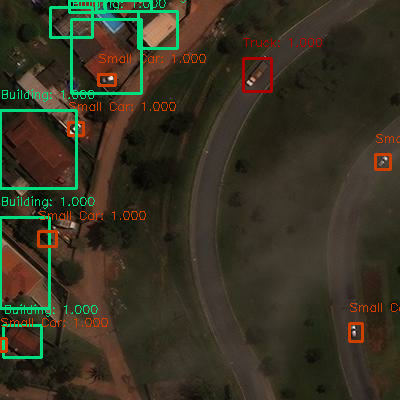}  
\end{subfigure}
\hfill
\begin{subfigure}[t]{.24\textwidth}
  \centering
  \includegraphics[width=\linewidth]{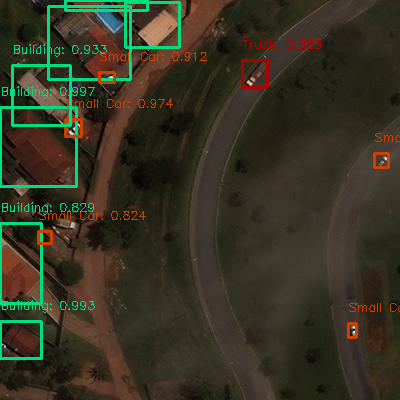}
\end{subfigure}
\hfill
\begin{subfigure}[t]{.24\textwidth}
  \centering
  \includegraphics[width=\linewidth]{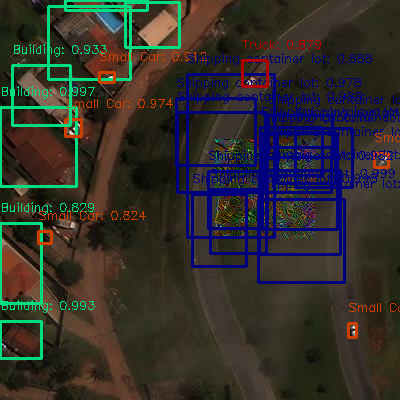}
\end{subfigure}
\hfill
\begin{subfigure}[t]{.24\textwidth}
  \centering
  \includegraphics[width=\linewidth]{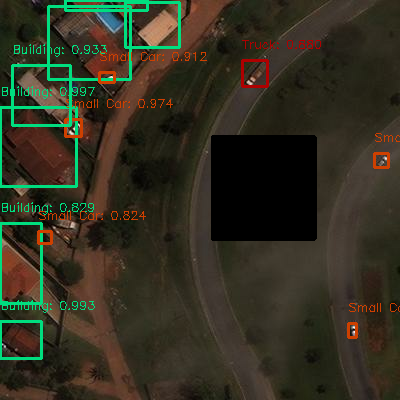}  
\end{subfigure}

\begin{subfigure}[t]{.24\textwidth}
  \centering
  \includegraphics[width=\linewidth]{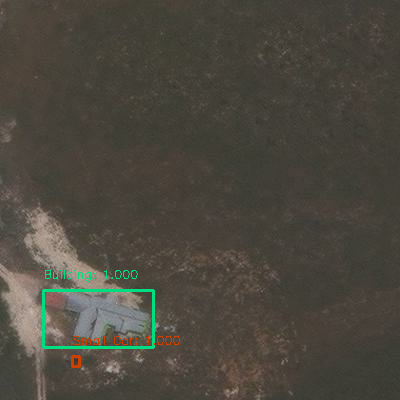}  
\end{subfigure}
\hfill
\begin{subfigure}[t]{.24\textwidth}
  \centering
  \includegraphics[width=\linewidth]{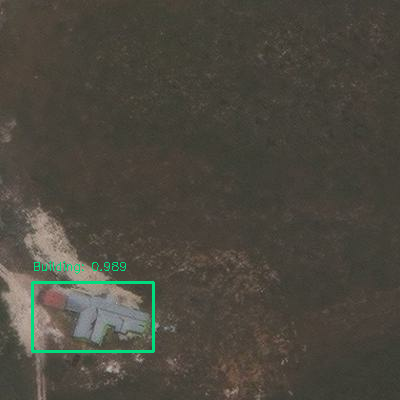}
\end{subfigure}
\hfill
\begin{subfigure}[t]{.24\textwidth}
  \centering
  \includegraphics[width=\linewidth]{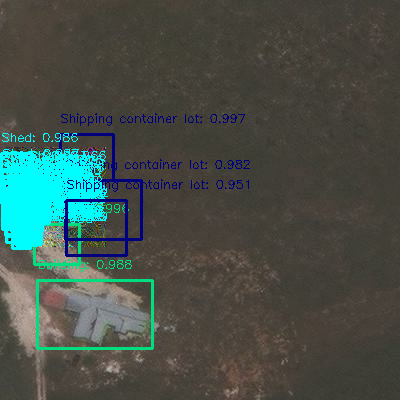}
\end{subfigure}
\hfill
\begin{subfigure}[t]{.24\textwidth}
  \centering
  \includegraphics[width=\linewidth]{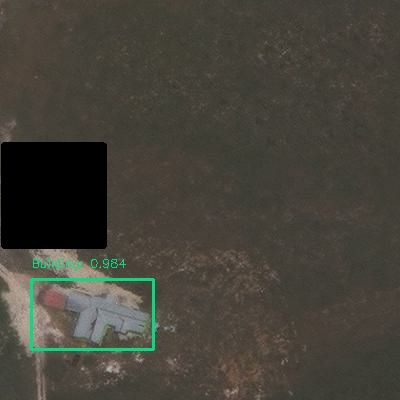}  
\end{subfigure}

\begin{subfigure}[t]{.24\textwidth}
  \centering
  \includegraphics[width=\linewidth]{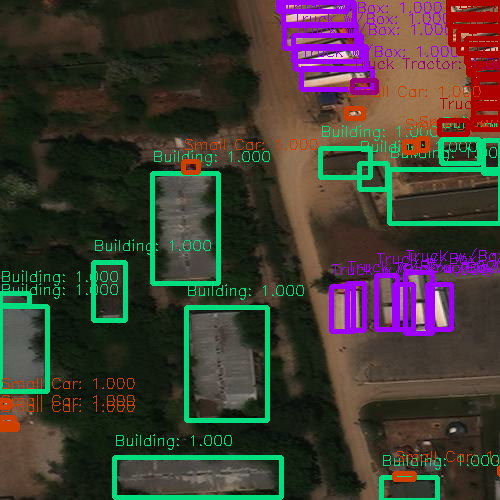}  
  \caption{Ground-truth on clean images.}
\end{subfigure}
\hfill
\begin{subfigure}[t]{.24\textwidth}
  \centering
  \includegraphics[width=\linewidth]{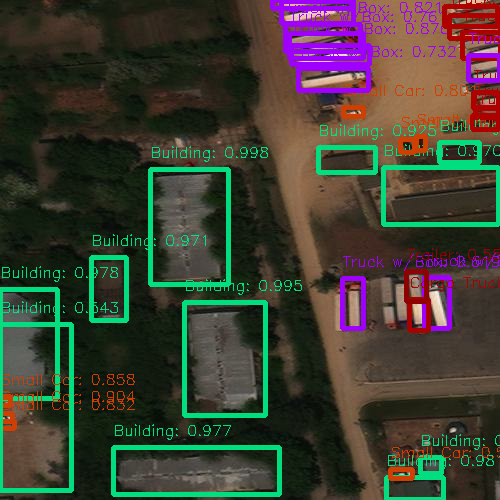}
  \caption{Predictions on clean images}
\end{subfigure}
\hfill
\begin{subfigure}[t]{.24\textwidth}
  \centering
  \includegraphics[width=\linewidth]{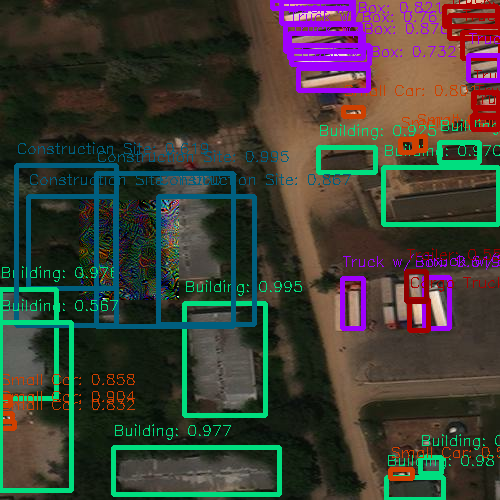}
  \caption{Predictions on adversarial images.}
\end{subfigure}
\hfill
\begin{subfigure}[t]{.24\textwidth}
  \centering
  \includegraphics[width=\linewidth]{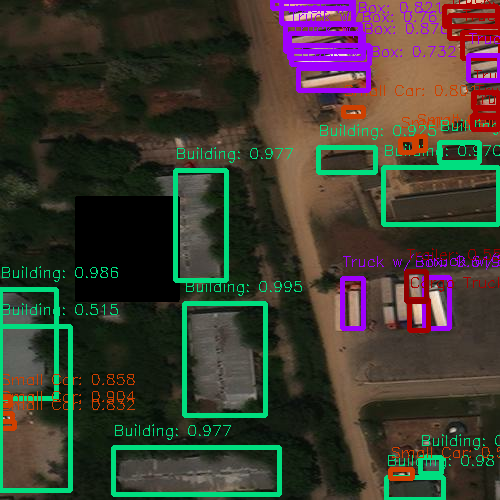}  
  \caption{Predictions on SAC masked images.}
\end{subfigure}

\caption{Examples on the xView dataset. The adversarial patches are $100\times 100$ squares generated by PGD adaptive attacks. Adversarial patches create spurious detections, and make the detector ignore the ground-truth objects. SAC can detect and remove the patches even under strong adaptive attacks, and therefore restore model predictions.}
\label{fig:more_xview}
\end{figure*}

We provide several examples of SAC under adaptive attacks in~\cref{fig:more_coco} and~\cref{fig:more_xview}. Adversarial patches create spurious detections, and make the detector ignore the ground-truth objects. SAC can detect and remove the adversarial patches even under strong adaptive attacks, and therefore restore model predictions.

\subsubsection{SAC v.s. Baselines}
\begin{figure*}
\begin{subfigure}[t]{.24\textwidth}
  \centering
  \includegraphics[width=\linewidth]{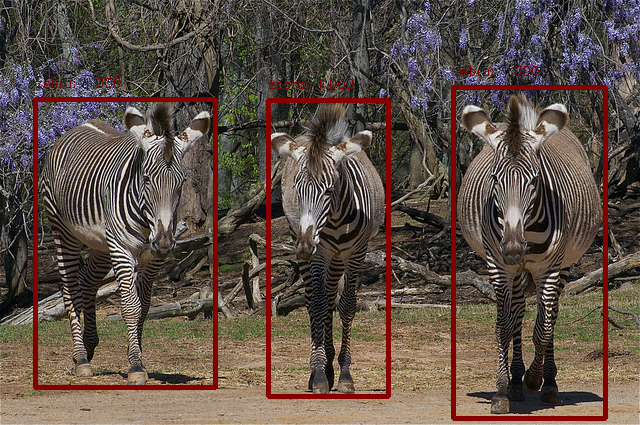}
  \caption{Ground-truth on the clean image.}
\end{subfigure}
\hfill
\begin{subfigure}[t]{.24\textwidth}
  \centering
  \includegraphics[width=\linewidth]{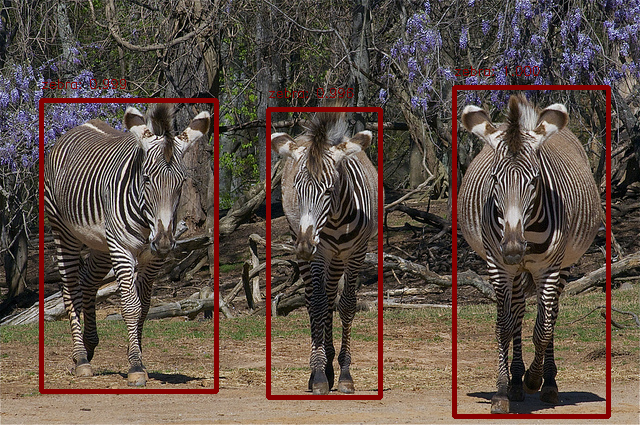}
  \caption{Predictions on the clean image.}
\end{subfigure}
\hfill
\begin{subfigure}[t]{.24\textwidth}
  \centering
  \includegraphics[width=\linewidth]{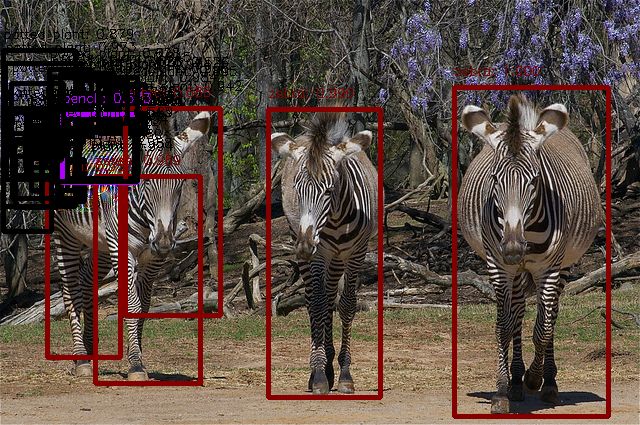}
  \caption{Predictions on the adversarial image.}
\end{subfigure}
\hfill
\begin{subfigure}[t]{.24\textwidth}
  \centering
  \includegraphics[width=\linewidth]{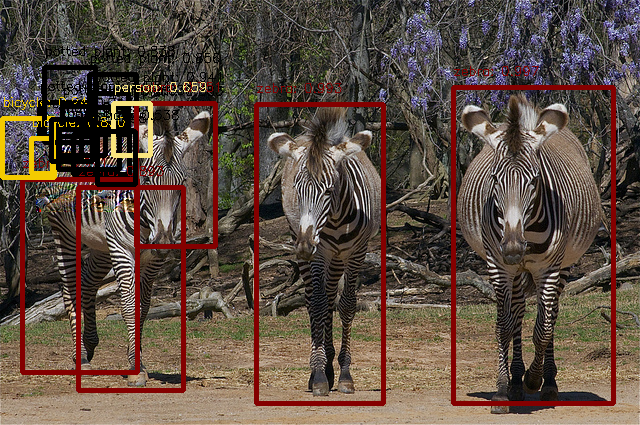}
  \caption{Predictions of the AT model on the adversarial image.}
\end{subfigure}
\hfill
\begin{subfigure}[t]{.24\textwidth}
  \centering
  \includegraphics[width=\linewidth]{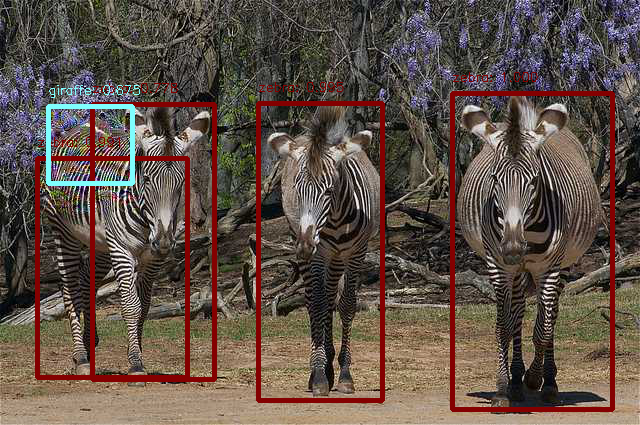}
  \caption{Predictions on the JPEG~\cite{dziugaite2016study} processed adversarial image (non-adaptive attack).}
\end{subfigure}
\hfill
\begin{subfigure}[t]{.24\textwidth}
  \centering
  \includegraphics[width=\linewidth]{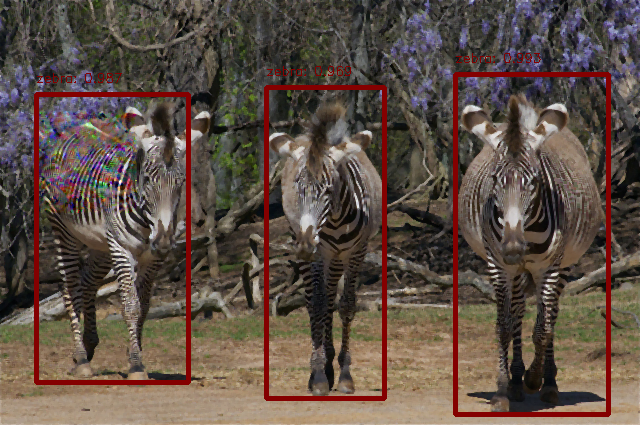}
  \caption{Predictions on the Spatial Smoothing~\cite{xu2017feature} processed adversarial image (non-adaptive attack).}
\end{subfigure}
\hfill
\begin{subfigure}[t]{.24\textwidth}
  \centering
  \includegraphics[width=\linewidth]{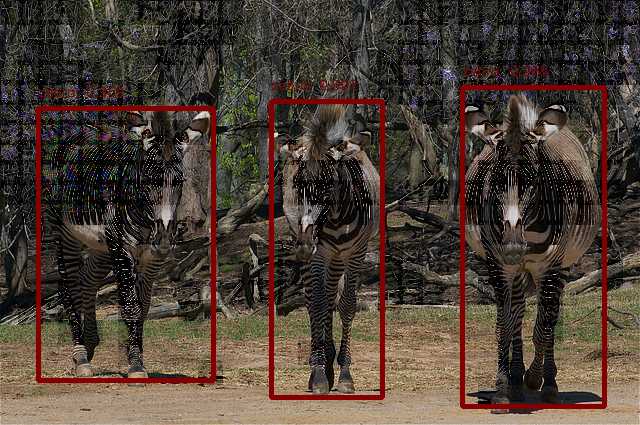}
  \caption{Predictions on the LGS~\cite{naseer2019local} processed adversarial image (non-adaptive attack).}
\end{subfigure}
\hfill
\begin{subfigure}[t]{.24\textwidth}
  \centering
  \includegraphics[width=\linewidth]{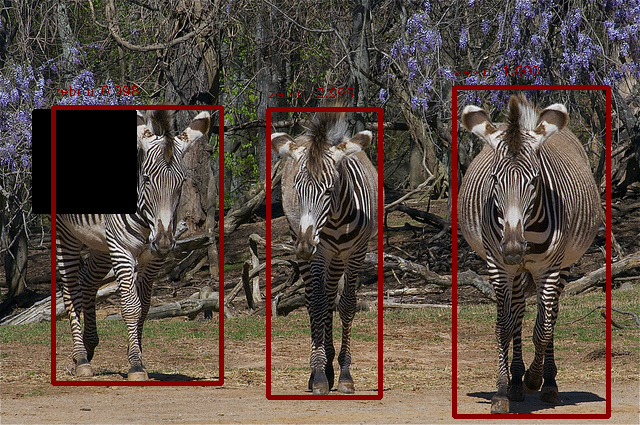}
  \caption{Predictions on the SAC masked adversarial image (non-adaptive attack).}
\end{subfigure}
\hfill
\begin{subfigure}[t]{.24\textwidth}
  \centering
  \includegraphics[width=\linewidth]{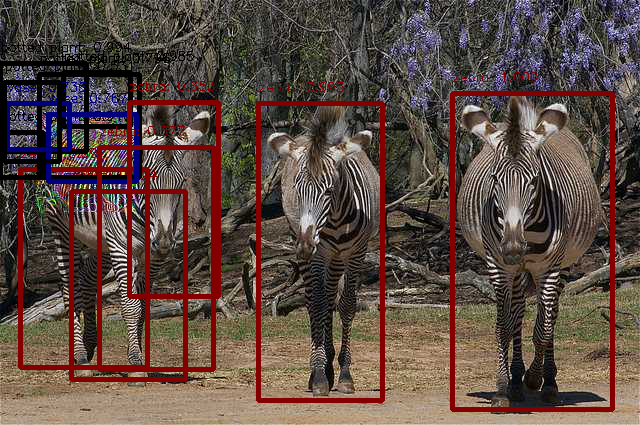}
  \caption{Predictions on the JPEG~\cite{dziugaite2016study} processed adversarial image (adaptive attack).}
\end{subfigure}
\hfill
\begin{subfigure}[t]{.24\textwidth}
  \centering
  \includegraphics[width=\linewidth]{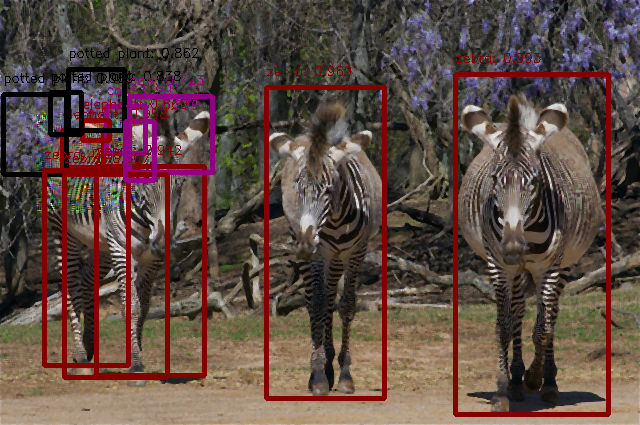}
  \caption{Predictions on the Spatial Smoothing~\cite{xu2017feature} processed adversarial image (adaptive attack).}
\end{subfigure}
\hfill
\begin{subfigure}[t]{.24\textwidth}
  \centering
  \includegraphics[width=\linewidth]{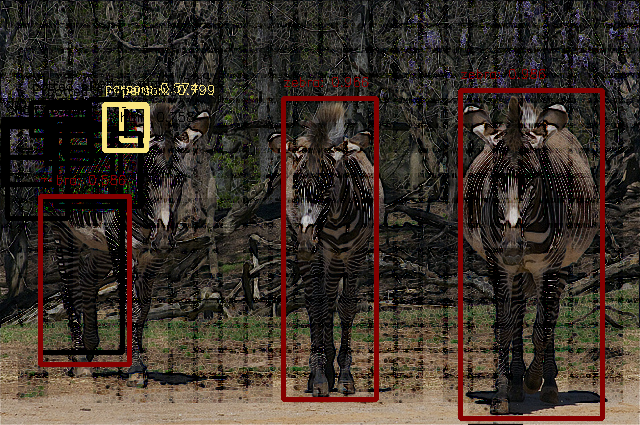}
  \caption{Predictions on the LGS~\cite{naseer2019local} processed adversarial image (adaptive attack).}
\end{subfigure}
\hfill
\begin{subfigure}[t]{.24\textwidth}
  \centering
  \includegraphics[width=\linewidth]{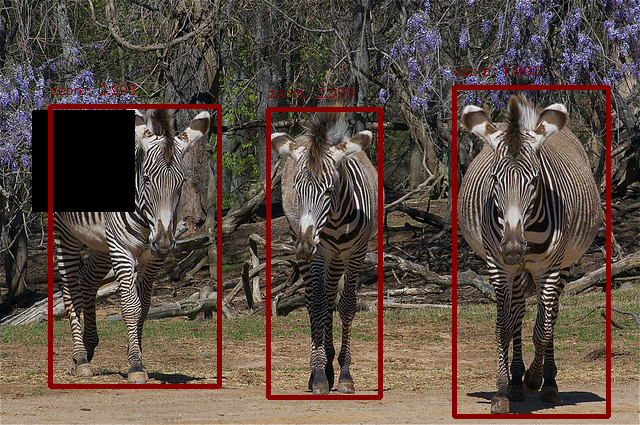}
  \caption{Predictions on the SAC masked adversarial image (adaptive attack).}
\end{subfigure}
\caption{Detection results of different defense methods on the COCO dataset. The adversarial patches are $100 \times 100$ squares and placed at the same location. JPEG~\cite{dziugaite2016study}, Spatial Smoothing~\cite{xu2017feature}, LGS~\cite{naseer2019local} have reasonable performance under non-adaptive attacks, but can not defend adaptive attacks where the adversary also attacks the pre-processing functions. In addition, they degrade image quality, especially LGS. SAC can defend both non-adaptive and adaptive attacks and maintains high image quality.}
\label{fig:baselines_coco}
\end{figure*}

\begin{figure*}
\begin{subfigure}[t]{.24\textwidth}
  \centering
  \includegraphics[width=\linewidth]{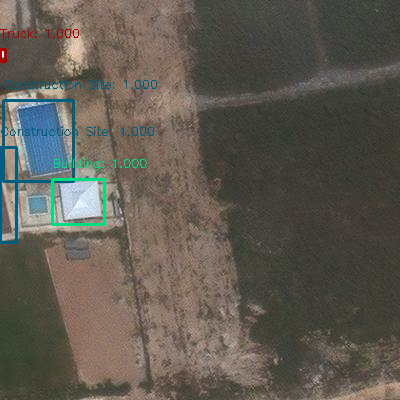}
  \caption{Ground-truth on the clean image.}
\end{subfigure}
\hfill
\begin{subfigure}[t]{.24\textwidth}
  \centering
  \includegraphics[width=\linewidth]{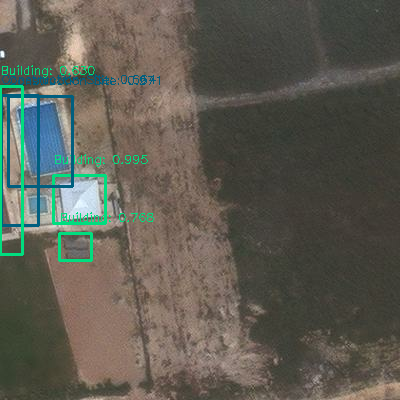}
  \caption{Predictions on the clean image.}
\end{subfigure}
\hfill
\begin{subfigure}[t]{.24\textwidth}
  \centering
  \includegraphics[width=\linewidth]{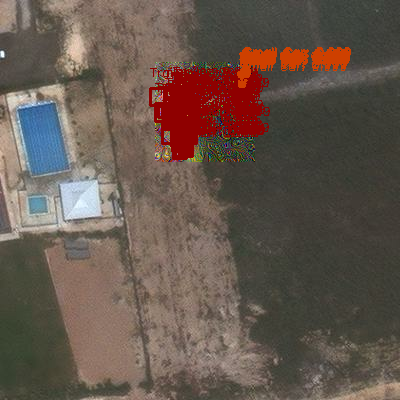}
  \caption{Predictions on the adversarial image.}
\end{subfigure}
\hfill
\begin{subfigure}[t]{.24\textwidth}
  \centering
  \includegraphics[width=\linewidth]{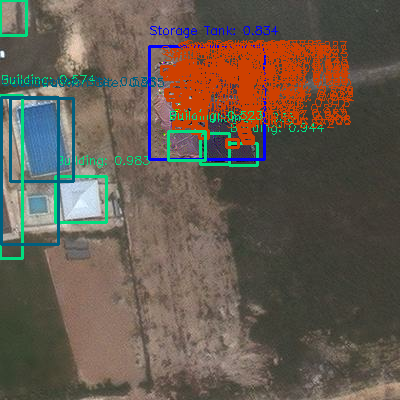}
  \caption{Predictions of the AT model on the adversarial image.}
\end{subfigure}
\hfill
\begin{subfigure}[t]{.24\textwidth}
  \centering
  \includegraphics[width=\linewidth]{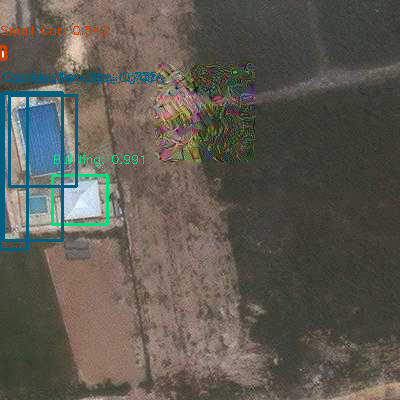}
  \caption{Predictions on the JPEG~\cite{dziugaite2016study} processed adversarial image (non-adaptive attack).}
\end{subfigure}
\hfill
\begin{subfigure}[t]{.24\textwidth}
  \centering
  \includegraphics[width=\linewidth]{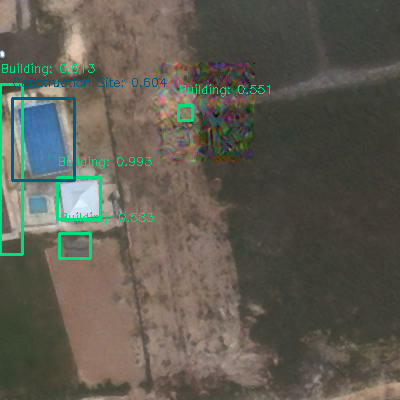}
  \caption{Predictions on the Spatial Smoothing~\cite{xu2017feature} processed adversarial image (non-adaptive attack).}
\end{subfigure}
\hfill
\begin{subfigure}[t]{.24\textwidth}
  \centering
  \includegraphics[width=\linewidth]{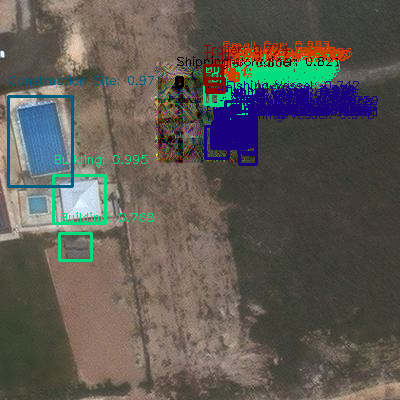}
  \caption{Predictions on the LGS~\cite{naseer2019local} processed adversarial image (non-adaptive attack).}
\end{subfigure}
\hfill
\begin{subfigure}[t]{.24\textwidth}
  \centering
  \includegraphics[width=\linewidth]{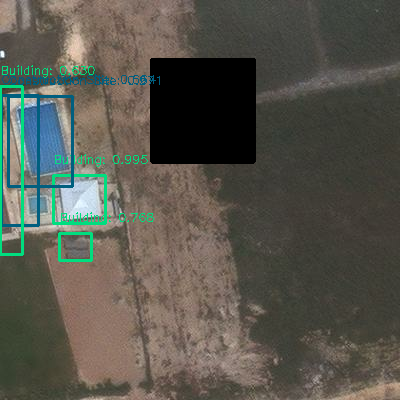}
  \caption{Predictions on the SAC masked adversarial image (non-adaptive attack).}
\end{subfigure}
\hfill
\begin{subfigure}[t]{.24\textwidth}
  \centering
  \includegraphics[width=\linewidth]{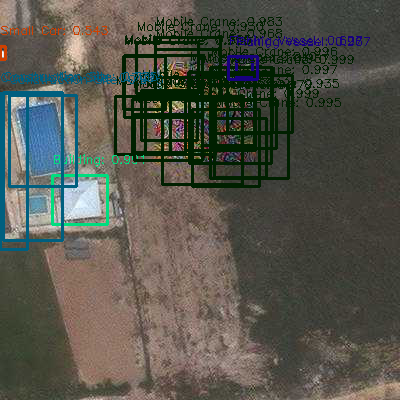}
  \caption{Predictions on the JPEG~\cite{dziugaite2016study} processed adversarial image (adaptive attack).}
\end{subfigure}
\hfill
\begin{subfigure}[t]{.24\textwidth}
  \centering
  \includegraphics[width=\linewidth]{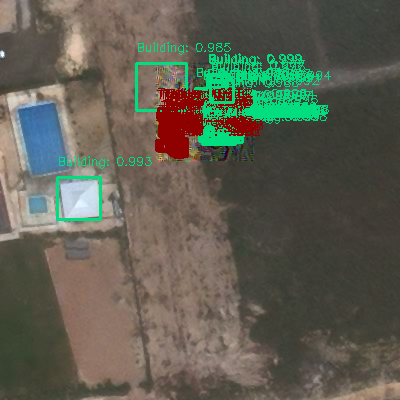}
  \caption{Predictions on the Spatial Smoothing~\cite{xu2017feature} processed adversarial image (adaptive attack).}
\end{subfigure}
\hfill
\begin{subfigure}[t]{.24\textwidth}
  \centering
  \includegraphics[width=\linewidth]{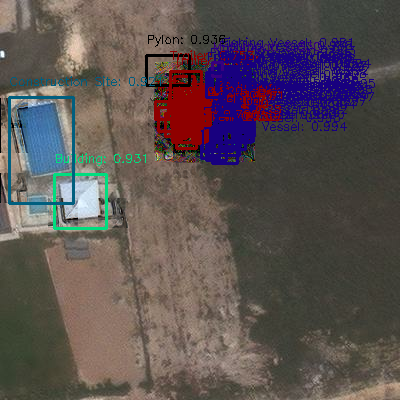}
  \caption{Predictions on the LGS~\cite{naseer2019local} processed adversarial image (adaptive attack).}
\end{subfigure}
\hfill
\begin{subfigure}[t]{.24\textwidth}
  \centering
  \includegraphics[width=\linewidth]{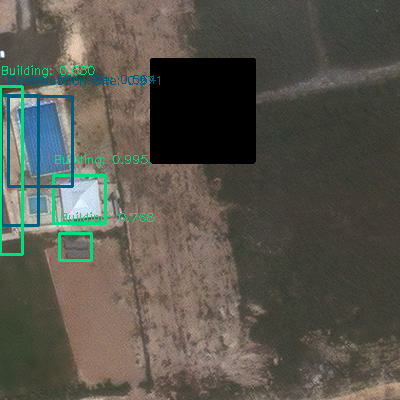}
  \caption{Predictions on the SAC masked adversarial image (adaptive attack).}
\end{subfigure}
\caption{Detection results of different defense methods on the xView dataset. The adversarial patches are $100 \times 100$ squares and placed at the same location. JPEG~\cite{dziugaite2016study}, Spatial Smoothing~\cite{xu2017feature}, LGS~\cite{naseer2019local} have reasonable performance under non-adaptive attacks, but can not defend adaptive attacks where the adversary also attacks the pre-processing functions. In addition, they degrade image quality, especially LGS. SAC can defend both non-adaptive and adaptive attacks and maintains high image quality.}
\label{fig:baselines_xview}
\end{figure*}

In this paper, we compare SAC with JPEG~\cite{dziugaite2016study}, Spatial Smoothing~\cite{xu2017feature}, LGS~\cite{naseer2019local}, and vanilla adversarial training (AT)~\cite{madry2017towards}. Visual comparisons are shown in~\cref{fig:baselines_coco} and \cref{fig:baselines_xview}. JPEG, Spatial Smoothing, LGS are pre-processing defenses that aim to remove the high-frequency information of adversarial patches. They have reasonable performance under non-adaptive attacks, but can not defend adaptive attacks where the adversary also attacks the pre-process functions. In addition, they degrade image quality, especially LGS, which degrades their performance on clean images. SAC can defend both non-adaptive and adaptive attacks. In addition, SAC does not degrade image quality, and therefore can maintain high performance on clean images.

\subsubsection{SAC under Different Attack Methods}
\begin{figure*}
\begin{subfigure}[t]{.24\textwidth}
  \centering
  \includegraphics[width=\linewidth]{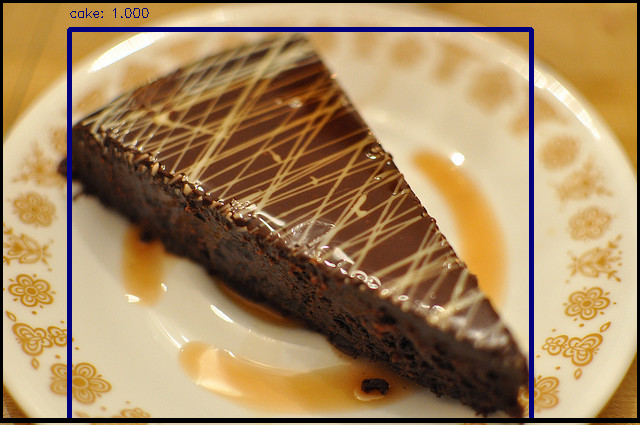}  
  \caption{Ground-truth on the clean image.}
\end{subfigure}
\hfill
\begin{subfigure}[t]{.24\textwidth}
  \centering
  \includegraphics[width=\linewidth]{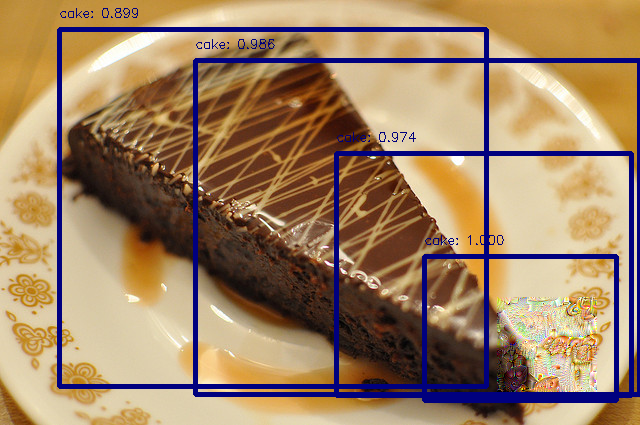}  
  \caption{Predictions on the PGD adversarial image.}
\end{subfigure}
\hfill
\begin{subfigure}[t]{.24\textwidth}
  \centering
  \includegraphics[width=\linewidth]{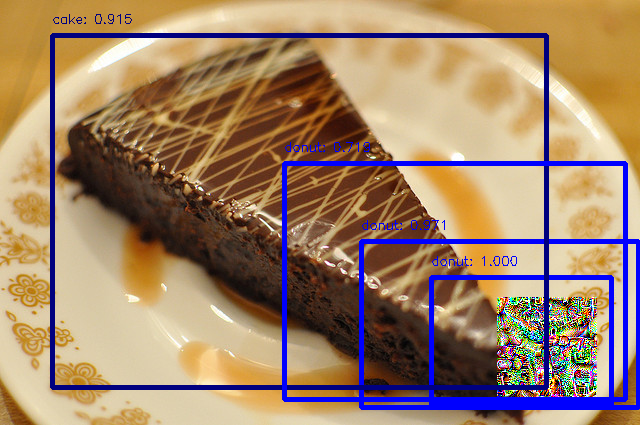}  
  \caption{Predictions on the MIM adversarial image.}
\end{subfigure}
\hfill
\begin{subfigure}[t]{.24\textwidth}
  \centering
  \includegraphics[width=\linewidth]{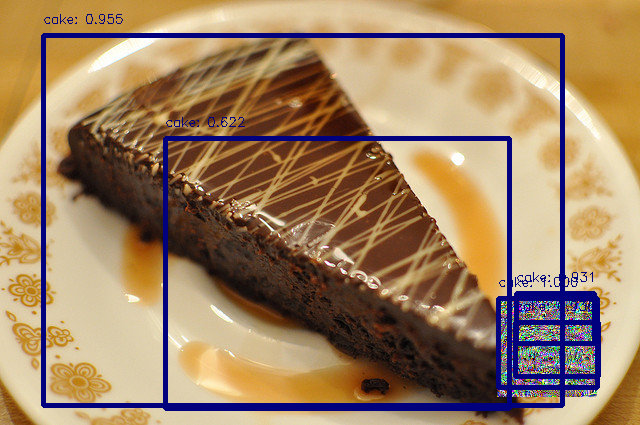}  
\caption{Prediction on the DPatch adversarial image (undefended).}
\end{subfigure}

\begin{subfigure}[t]{.24\textwidth}
  \centering
  \includegraphics[width=\linewidth]{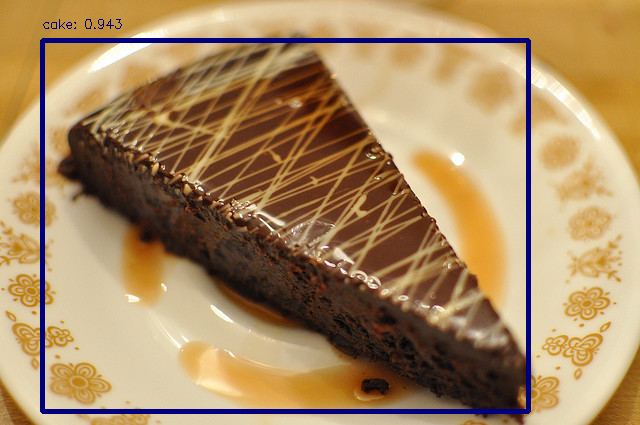} 
  \caption{Predictions on the clean image.}
\end{subfigure}
\hfill
\begin{subfigure}[t]{.24\textwidth}
  \centering
  \includegraphics[width=\linewidth]{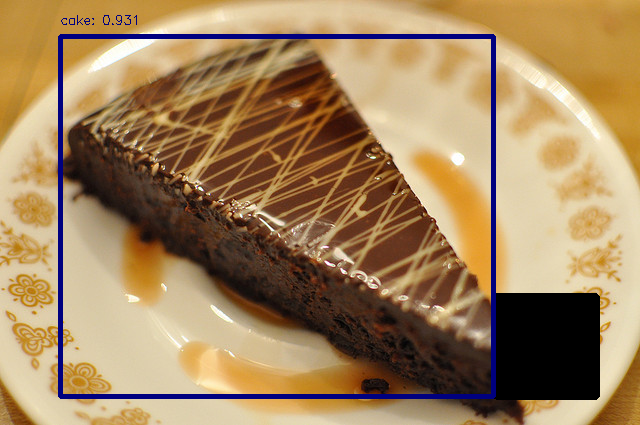}  
  \caption{Predictions on the SAC masked PGD adversarial image.}
\end{subfigure}
\hfill
\begin{subfigure}[t]{.24\textwidth}
  \centering
  \includegraphics[width=\linewidth]{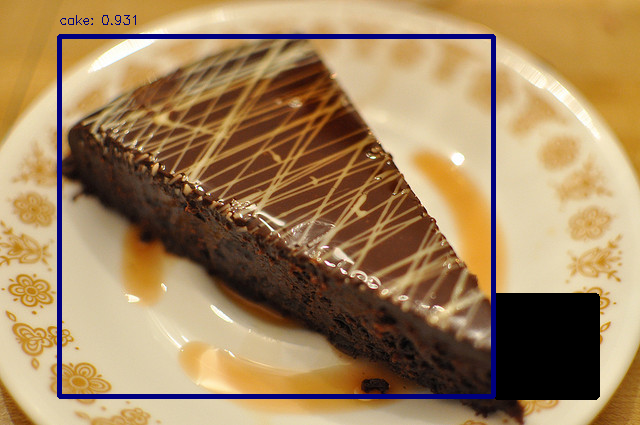}  
  \caption{Predictions on the SAC masked MIM adversarial image.}
\end{subfigure}
\hfill
\begin{subfigure}[t]{.24\textwidth}
  \centering
  \includegraphics[width=\linewidth]{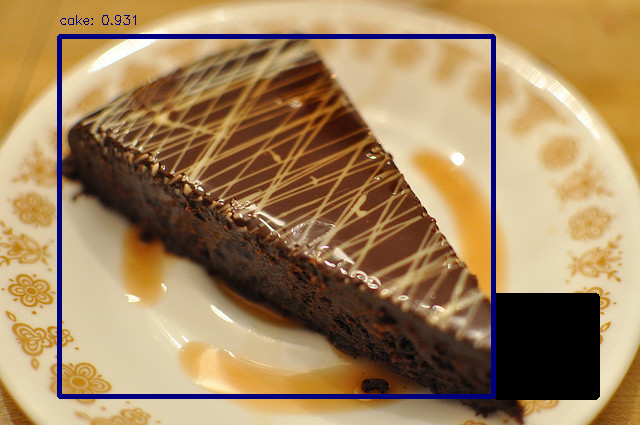}  
  \caption{Predictions on the SAC masked DPatch adversarial image.}
\end{subfigure}

\caption{Detection results on a clean image and corresponding adversarial images generated by different attack methods. The image is taken from the COCO dataset. The adversarial patches are $100 \times 100$ squares and placed at the same location.}
\label{fig:attacks_coco}
\end{figure*}

\begin{figure*}
\begin{subfigure}[t]{.24\textwidth}
  \centering
  \includegraphics[width=\linewidth]{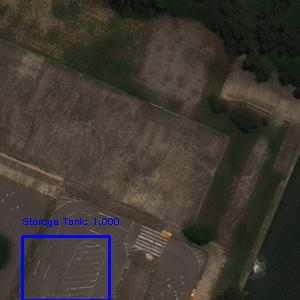}  
  \caption{Ground-truth on the clean image.}
\end{subfigure}
\hfill
\begin{subfigure}[t]{.24\textwidth}
  \centering
  \includegraphics[width=\linewidth]{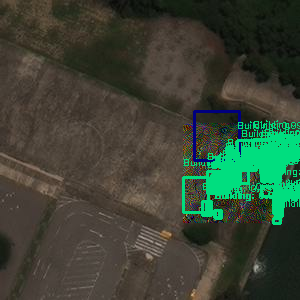}  
  \caption{Predictions on the PGD adversarial image.}
\end{subfigure}
\hfill
\begin{subfigure}[t]{.24\textwidth}
  \centering
  \includegraphics[width=\linewidth]{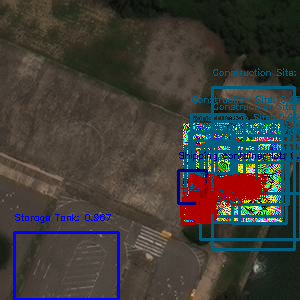}  
  \caption{Predictions on the MIM adversarial image.}
\end{subfigure}
\hfill
\begin{subfigure}[t]{.24\textwidth}
  \centering
  \includegraphics[width=\linewidth]{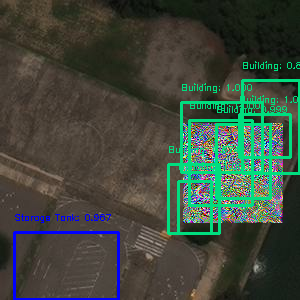}  
\caption{Prediction on the DPatch adversarial image (undefended).}
\end{subfigure}

\begin{subfigure}[t]{.24\textwidth}
  \centering
  \includegraphics[width=\linewidth]{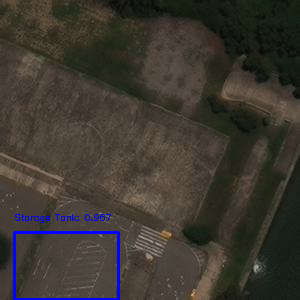} 
  \caption{Predictions on the clean image.}
\end{subfigure}
\hfill
\begin{subfigure}[t]{.24\textwidth}
  \centering
  \includegraphics[width=\linewidth]{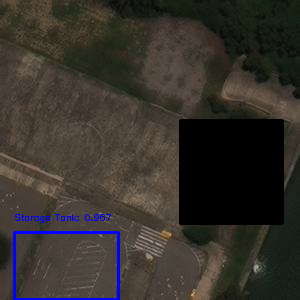}  
  \caption{Predictions on the SAC masked PGD adversarial image.}
\end{subfigure}
\hfill
\begin{subfigure}[t]{.24\textwidth}
  \centering
  \includegraphics[width=\linewidth]{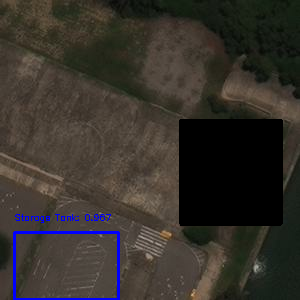}  
  \caption{Predictions on the SAC masked MIM adversarial image.}
\end{subfigure}
\hfill
\begin{subfigure}[t]{.24\textwidth}
  \centering
  \includegraphics[width=\linewidth]{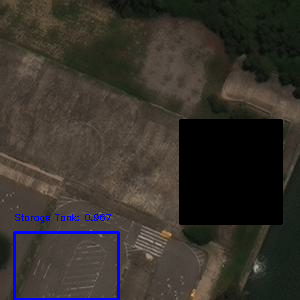}  
  \caption{Predictions on the SAC masked DPatch adversarial image.}
\end{subfigure}

\caption{Detection results on a clean image and corresponding adversarial images generated by different attack methods. The image is taken from the xView dataset. The adversarial patches are $100 \times 100$ squares and placed at the same location.}
\label{fig:attacks_xview}
\end{figure*}

We visualize the detection results of SAC under different attacks in~\cref{fig:attacks_coco} and~\cref{fig:attacks_xview}, including PGD~\cite{madry2017towards}, MIM~\cite{dong2018boosting} and DPatch~\cite{liu2018dpatch}. SAC can effectively detect and remove the adversarial patches under different attacks and restore the model predictions. We also notice that the adversarial patches generated by different methods has different styles. PGD generated adversarial patches are less visible, even though it has the same $\epsilon=1$ attack budget.

\subsubsection{SAC under Different Patch Shapes}
\begin{figure*}
\begin{subfigure}[t]{.24\textwidth}
  \centering
  \includegraphics[width=\linewidth]{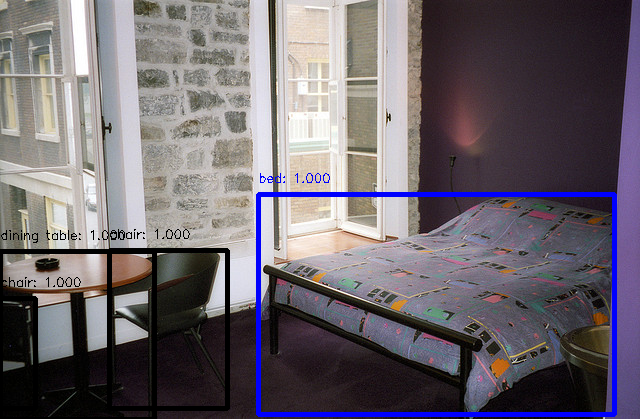}  
  \caption{Ground-truth on the clean image.}
\end{subfigure}
\hfill
\begin{subfigure}[t]{.24\textwidth}
  \centering
  \includegraphics[width=\linewidth]{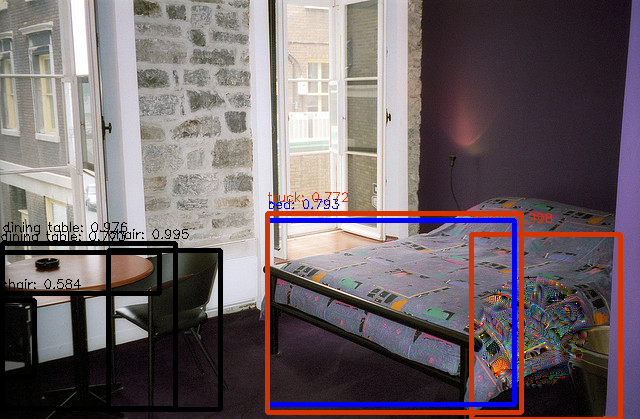}  
  \caption{Predictions on the adversarial image with a circle patch.}
\end{subfigure}
\hfill
\begin{subfigure}[t]{.24\textwidth}
  \centering
  \includegraphics[width=\linewidth]{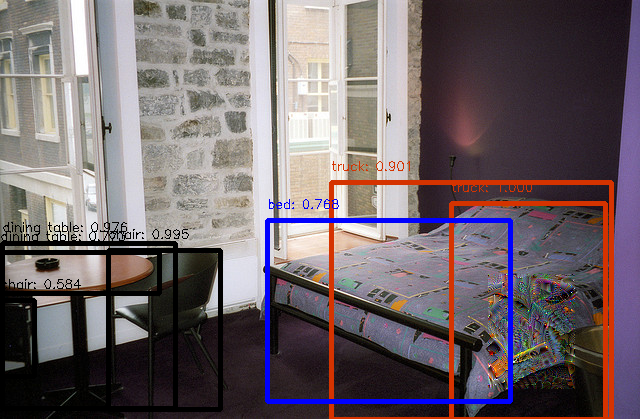}  
  \caption{Predictions on the adversarial image with a rectangle patch.}
\end{subfigure}
\hfill
\begin{subfigure}[t]{.24\textwidth}
  \centering
  \includegraphics[width=\linewidth]{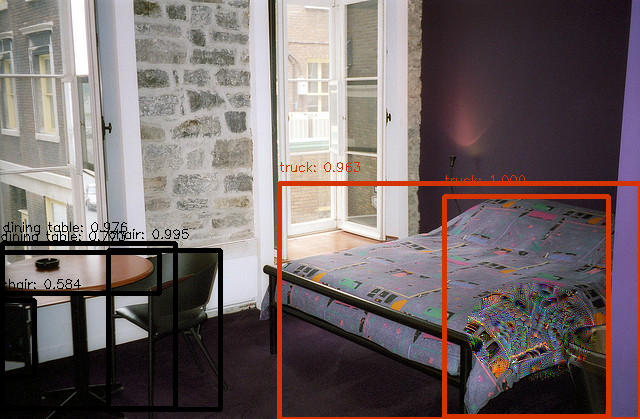}  
  \caption{Predictions on the adversarial image with a ellipse patch.}
\end{subfigure}

\begin{subfigure}[t]{.24\textwidth}
  \centering
  \includegraphics[width=\linewidth]{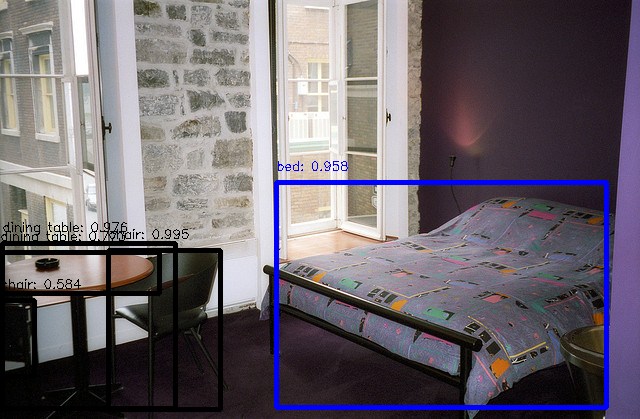} 
  \caption{Predictions on the clean image.}
\end{subfigure}
\hfill
\begin{subfigure}[t]{.24\textwidth}
  \centering
  \includegraphics[width=\linewidth]{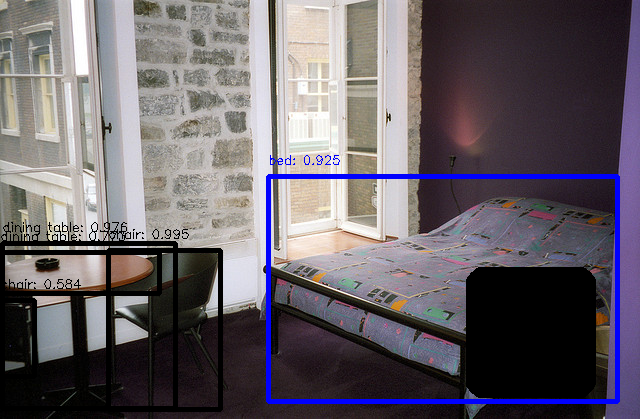}  
  \caption{Predictions on the SAC masked adversarial image with a circle patch.}
\end{subfigure}
\hfill
\begin{subfigure}[t]{.24\textwidth}
  \centering
  \includegraphics[width=\linewidth]{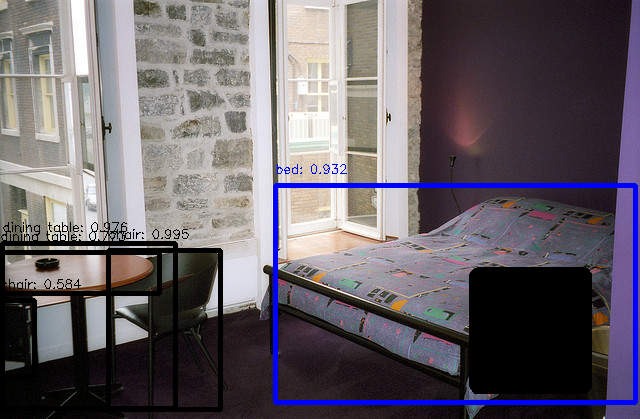}  
  \caption{Predictions on the SAC masked adversarial image with a rectangle patch.}
\end{subfigure}
\hfill
\begin{subfigure}[t]{.24\textwidth}
  \centering
  \includegraphics[width=\linewidth]{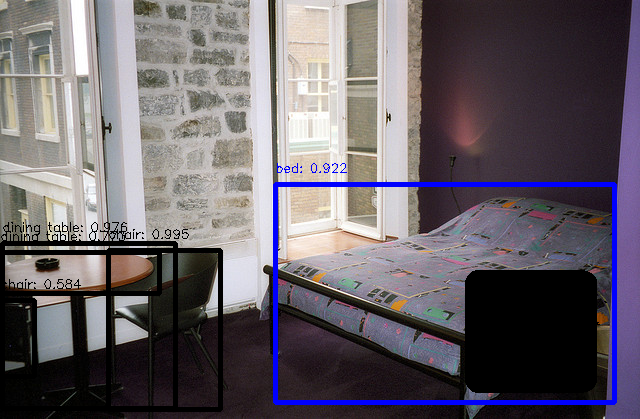}  
  \caption{Predictions on the SAC masked adversarial image with an ellipse patch.}
\end{subfigure}

\caption{Detection results on adversarial images with different patch shapes. The image is taken from the COCO dataset. The adversarial patches have $100 \times 100$ pixels and placed at the same location.}
\label{fig:shapes_coco}
\end{figure*}

\begin{figure*}
\begin{subfigure}[t]{.24\textwidth}
  \centering
  \includegraphics[width=\linewidth]{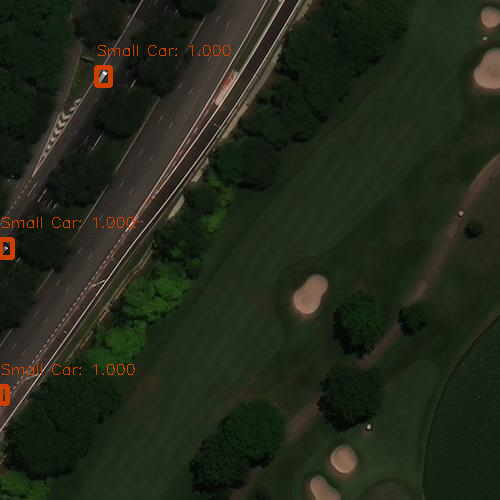}  
  \caption{Ground-truth.}
\end{subfigure}
\hfill
\begin{subfigure}[t]{.24\textwidth}
  \centering
  \includegraphics[width=\linewidth]{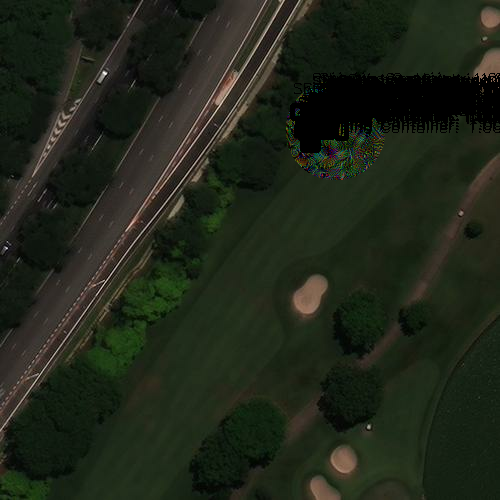}  
  \caption{Predictions on the adversarial image with a circle patch.}
\end{subfigure}
\hfill
\begin{subfigure}[t]{.24\textwidth}
  \centering
  \includegraphics[width=\linewidth]{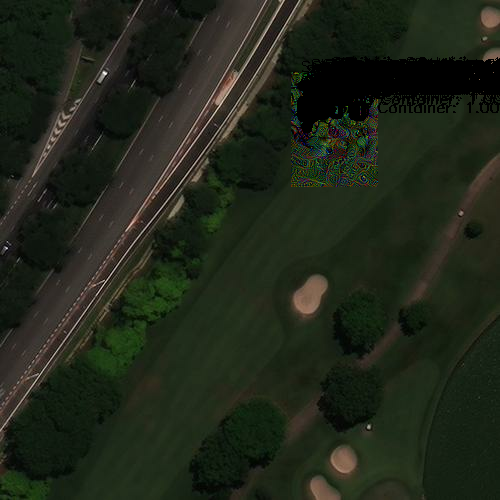}  
  \caption{Predictions on the adversarial image with a rectangle patch.}
\end{subfigure}
\hfill
\begin{subfigure}[t]{.24\textwidth}
  \centering
  \includegraphics[width=\linewidth]{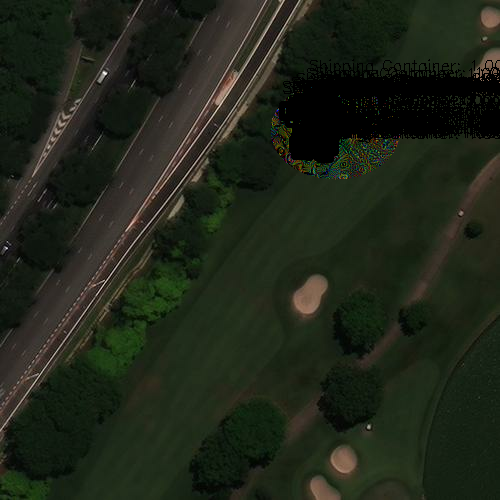}  
  \caption{Predictions on the adversarial image with a ellipse patch.}
\end{subfigure}

\begin{subfigure}[t]{.24\textwidth}
  \centering
  \includegraphics[width=\linewidth]{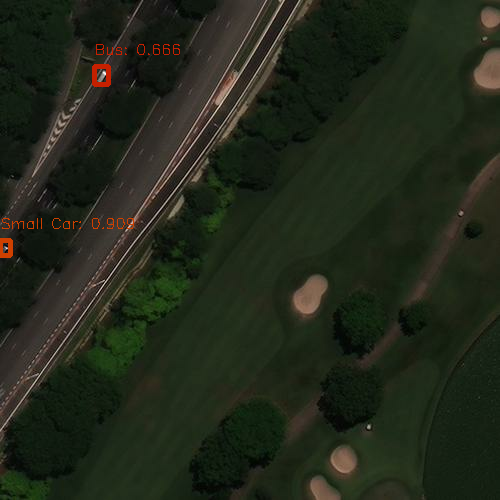} 
  \caption{Predictions on the clean image.}
\end{subfigure}
\hfill
\begin{subfigure}[t]{.24\textwidth}
  \centering
  \includegraphics[width=\linewidth]{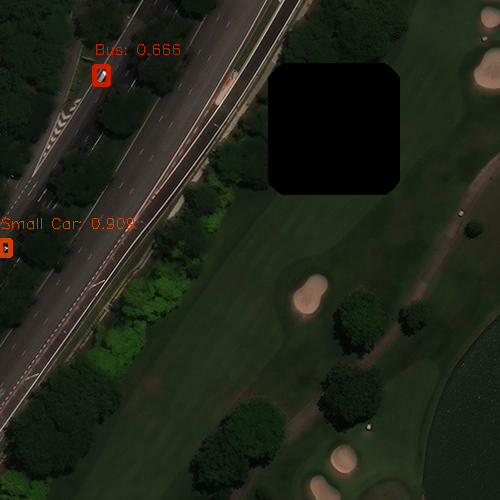}  
  \caption{Predictions on the SAC masked adversarial image with a circle patch.}
\end{subfigure}
\hfill
\begin{subfigure}[t]{.24\textwidth}
  \centering
  \includegraphics[width=\linewidth]{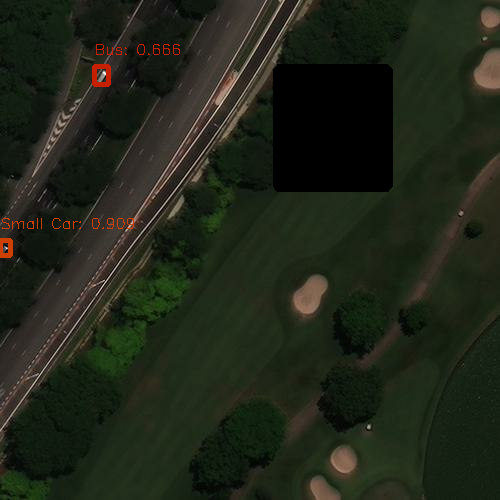}  
  \caption{Predictions on the SAC masked adversarial image with a rectangle patch.}
\end{subfigure}
\hfill
\begin{subfigure}[t]{.24\textwidth}
  \centering
  \includegraphics[width=\linewidth]{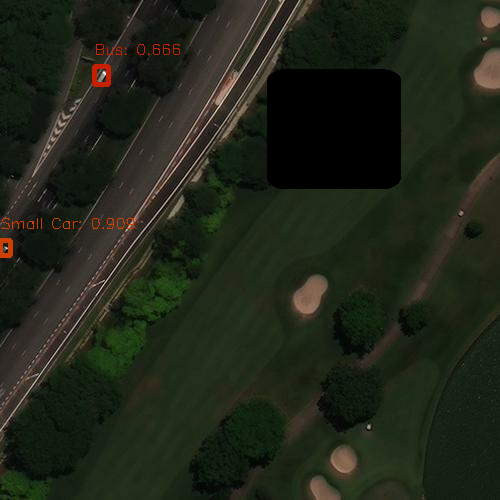}  
  \caption{Predictions on the SAC masked adversarial image with an ellipse patch.}
\end{subfigure}

\caption{Detection results on adversarial images with different patch shapes. The image is taken from the xView dataset. The adversarial patches have $100 \times 100$ pixels and placed at the same location.}
\label{fig:shapes_xview}
\end{figure*}

We visualize the detection results of SAC under PGD attacks with unseen patch shapes in~\cref{fig:shapes_coco} and~\cref{fig:shapes_xview}, including circle, rectangle and ellipse. SAC can effectively detect and remove the adversarial patches of different shapes and restore the model predictions, even though those shapes are used in training the patch segmenter and mismatch the square shape prior in shape completion. However, we do notice that masked region can be larger than the original patches as SAC tries to cover the patch with square shapes.

\subsubsection{Failure Cases}
\label{sec:failure}
\begin{figure*}
\begin{subfigure}[t]{.24\textwidth}
  \centering
  \includegraphics[width=\linewidth]{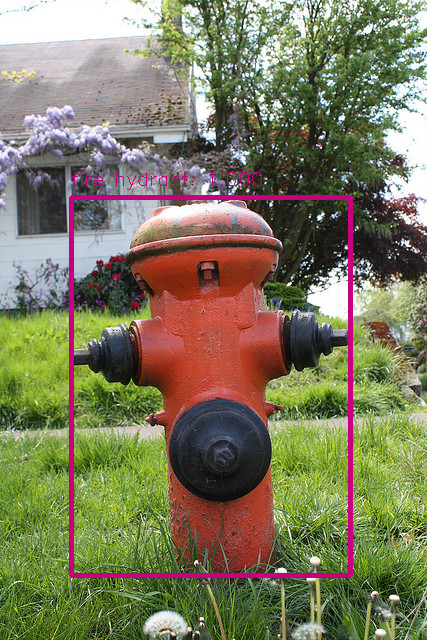}  
\end{subfigure}
\hfill
\begin{subfigure}[t]{.24\textwidth}
  \centering
  \includegraphics[width=\linewidth]{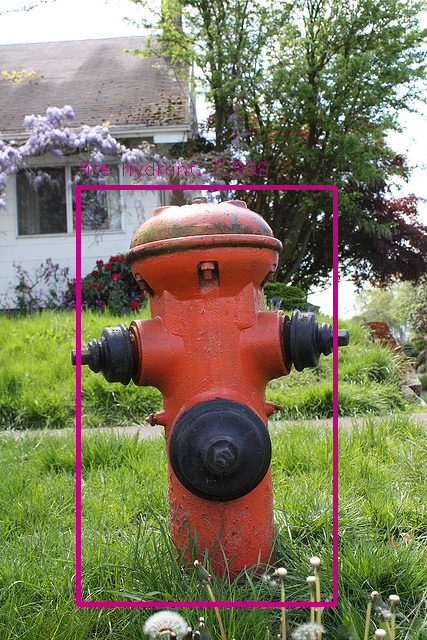}
\end{subfigure}
\hfill
\begin{subfigure}[t]{.24\textwidth}
  \centering
  \includegraphics[width=\linewidth]{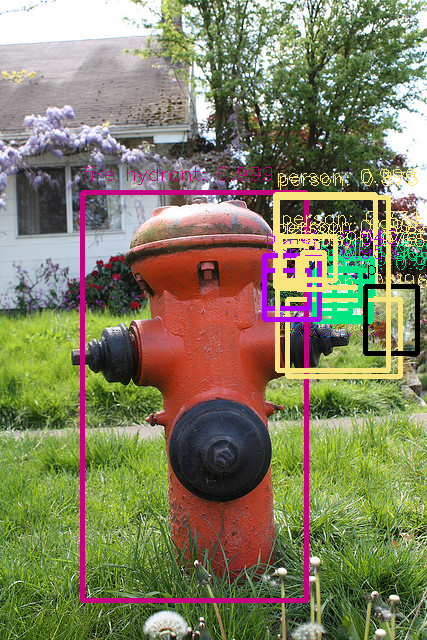}
\end{subfigure}
\hfill
\begin{subfigure}[t]{.24\textwidth}
  \centering
  \includegraphics[width=\linewidth]{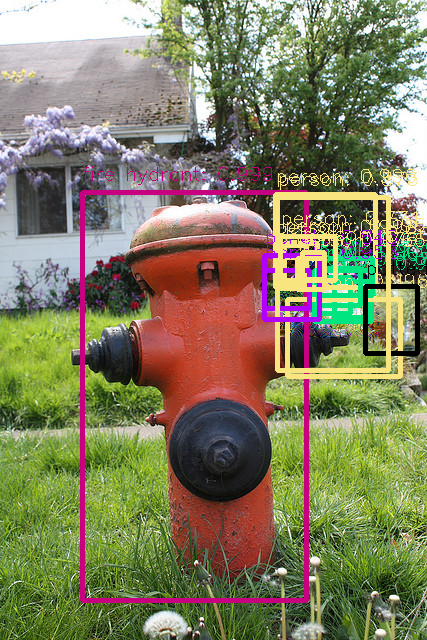}  
\end{subfigure}

\begin{subfigure}[t]{.24\textwidth}
  \centering
  \includegraphics[width=\linewidth]{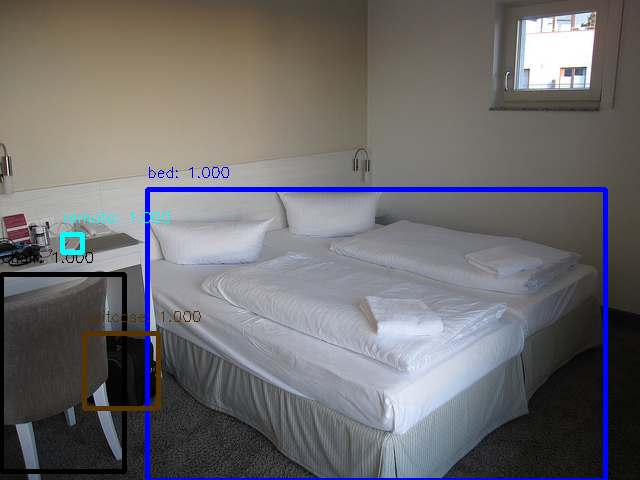}  
\end{subfigure}
\hfill
\begin{subfigure}[t]{.24\textwidth}
  \centering
  \includegraphics[width=\linewidth]{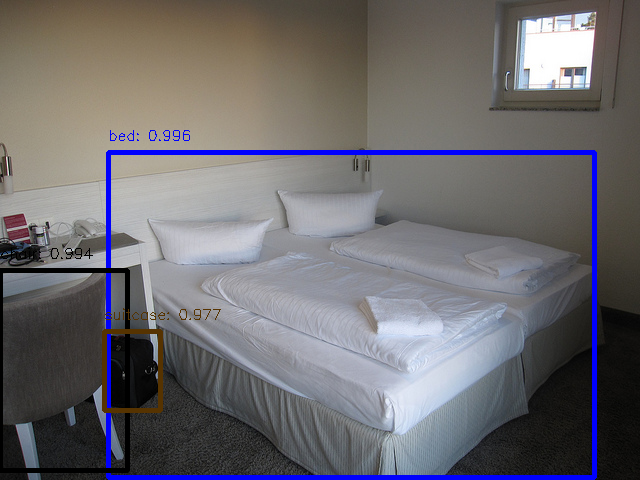}
\end{subfigure}
\hfill
\begin{subfigure}[t]{.24\textwidth}
  \centering
  \includegraphics[width=\linewidth]{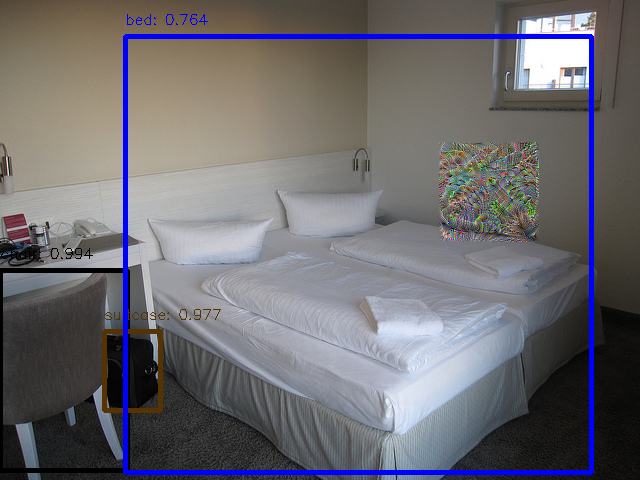}
\end{subfigure}
\hfill
\begin{subfigure}[t]{.24\textwidth}
  \centering
  \includegraphics[width=\linewidth]{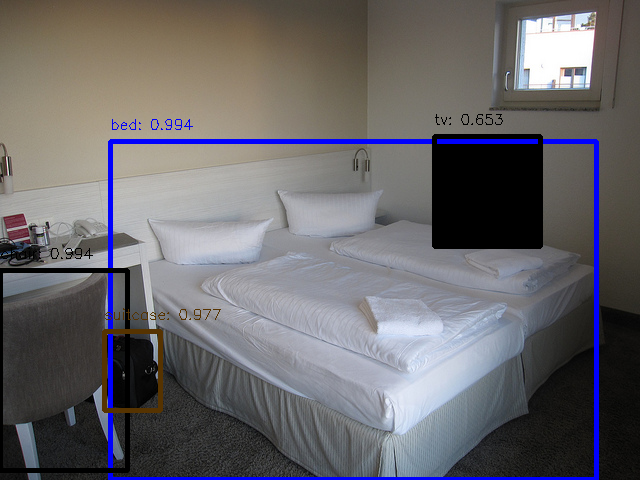}  
\end{subfigure}

\begin{subfigure}[t]{.24\textwidth}
  \centering
  \includegraphics[width=\linewidth]{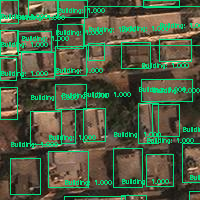}  
  \caption{Ground-truth on clean images.}
\end{subfigure}
\hfill
\begin{subfigure}[t]{.24\textwidth}
  \centering
  \includegraphics[width=\linewidth]{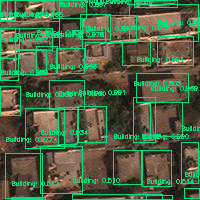}
  \caption{Predictions on clean images}
\end{subfigure}
\hfill
\begin{subfigure}[t]{.24\textwidth}
  \centering
  \includegraphics[width=\linewidth]{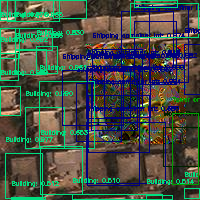}
  \caption{Predictions on adversarial images.}
\end{subfigure}
\hfill
\begin{subfigure}[t]{.24\textwidth}
  \centering
  \includegraphics[width=\linewidth]{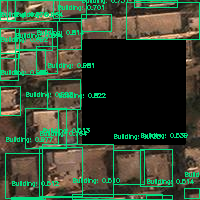}  
  \caption{Predictions on SAC masked images.}
\end{subfigure}

\caption{Examples of failure cases. Row 1: SAC fails to detect and remove the adversarial patch, which happens very rarely. Row 2: the black block from masking out the patch creates a false detection of ``TV". Row 3: the black block from masking out the patch cover foreground objects. See the discussion in Section \ref{sec:failure}.}
\label{fig:failure}
\end{figure*}

There are several failure modes in SAC: 1) SAC completely fails to detect a patch (\eg, ~\cref{fig:failure} row 1), which happens very rarely; 2) SAC successfully detects and removes a patch, but the black blocks from patch removing causes misdetection (\eg, ~\cref{fig:failure} row 2), which happens more often on the COCO dataset since black blocks resemble some object categories in the dataset such as TV, traffic light, and suitcase; 3) SAC successfully detects and removes a patch, but the patch covers foreground objects and thus the object detector fails to detect the objects on the masked image (\eg,~\cref{fig:failure} row 3). We can potentially mitigate the first issue by improving the patch segmenter, such as using more advanced segmentation networks and doing longer self adversarial training. For the second issue, we can avoid it by fine-tuning the base object detetor on images with randomly-placed black blocks. For the third issue, if the attacker is allowed to arbitrarily distort the pixels and destroy all the information within the patch such as in physical patch attacks, there is no chance that we can detect the objects hiding behind the adversarial patches. However, in the case where the patches are less visible, some information may be preserved in the patched area. We can potentially impaint or reconstruct the content within the patches to help detection.

\end{document}